\PassOptionsToPackage{table}{xcolor}
\documentclass{article}

\usepackage[preprint]{neurips_2026}

\usepackage[utf8]{inputenc}
\usepackage[T1]{fontenc}
\usepackage{hyperref}
\usepackage{url}
\usepackage{xspace}
\usepackage{booktabs}
\usepackage{amsfonts}
\usepackage{amsmath}
\usepackage{nicefrac}
\usepackage{microtype}
\usepackage[table]{xcolor}
\usepackage{enumitem}
\usepackage{graphicx}
\usepackage{listings}
\usepackage{multirow}
\usepackage{colortbl}
\usepackage{caption}
\usepackage{subcaption}
\usepackage{algorithm}
\usepackage{algpseudocode}   
\usepackage{pifont}
\usepackage{tikz}
\usepackage[breakable,skins]{tcolorbox}
\usepackage{float}
\usepackage{arydshln}        % for \cdashline in tables
\usepackage{wrapfig}         % for text-wrapped figures
\setcitestyle{numbers}
\usepackage[normalem]{ulem}
\usepackage{subcaption}
\newcommand{\cmark}{\ding{51}}
\newcommand{\xmark}{\ding{55}}

\newtcolorbox{promptbox}[1][]{%
  colback=gray!5,
  colframe=gray!50,
  fonttitle=\small\bfseries,
  title={#1},
  boxrule=0.4pt,
  arc=1.5pt,
  left=4pt, right=4pt, top=2pt, bottom=2pt,
  fontupper=\small\ttfamily,
  breakable
}

\newcommand{\ours}{\textsc{MoBayes}\xspace}
\newcommand{\discuss}[1]{#1}  % was \textcolor{red} during drafting

\newcommand{\figcaption}[1]{\vspace*{-3mm}\caption{#1}\vspace*{-5mm}}
\newcommand{\moveup}{\vspace*{-2mm}}
\newcommand{\moveups}{\vspace*{-1mm}}
\newcommand{\ignore}[1]{}

\newif\ifshowedits
\showeditsfalse  % flip to \showeditstrue to make revisions visible

\ifshowedits
  
  \newcommand{\del}[1]{{\color{red}\sout{#1}}}
  \newcommand{\wip}[1]{{\color{orange!80!black}\uwave{#1}}}
  \newcommand{\delblock}[1]{%
    \begingroup
    \color{red!75!black}%
    \marginpar{\footnotesize\color{red!75!black}\textbf{[DELETED]}}%
    #1%
    \endgroup}
  
  \newcommand{\wipblock}[1]{%
    \begingroup
    \color{orange!80!black}%
    \marginpar{\footnotesize\color{orange!80!black}\textbf{[WIP]}}%
    #1%
    \endgroup}
\else
  
  \newcommand{\del}[1]{}
  \newcommand{\wip}[1]{#1}
  \newcommand{\delblock}[1]{}
  
  \newcommand{\wipblock}[1]{#1}
\fi
\newcommand{\replace}[2]{\del{#1}\wip{#2}}

\definecolor{dhsgreen}{HTML}{22A559}
\definecolor{dhsred}{HTML}{E05252}
\definecolor{f1high}{HTML}{C6EFCE}
\definecolor{f1mid}{HTML}{FFFFCC}
\definecolor{f1low}{HTML}{FFCCCC}
\newcommand{\highlightgreen}[1]{\colorbox{dhsgreen!40}{\textbf{#1}}}
\newcommand{\highlightred}[1]{\colorbox{dhsred!25}{\textbf{#1}}}

\title{\ours: A Modular Bayesian Framework for Separating Reasoning from Language in Conversational Clinical Decision Support}

\author{%
  Yusuf Kesmen\thanks{Equal contribution. Corresponding author: \texttt{yusuf.kesmen@epfl.ch}} \\
  LiGHT, EPFL \\
  \And
  Fay Elhassan\footnotemark[1] \\
  LiGHT, EPFL \\
  \And
  Jiayi Ma\footnotemark[1] \\
  University of Bern \\
  \AND
  Julien Stalhandske \\
  LiGHT, EPFL \\
  \And
  Yena Chang \\
  LiGHT, EPFL \\
  \And
  David Sasu \\
  LiGHT, EPFL \\
  \And
  Alexandra Kulinkina \\
  LiGHT, EPFL \\
  \AND
  Akhil Arora\thanks{Equal supervision} \\
  CLAN, Aarhus University \\
  \And
  Lars Klein\footnotemark[2] \\
  LiGHT, EPFL \\
  \And
  Mary-Anne Hartley\footnotemark[2] \\
  LiGHT, EPFL \\
}

\begin{document}

\maketitle

%% --- Abstract ------------------------------------------------
\begin{abstract}
%% ============================================================
%% ABSTRACT
%% File: sections/abstract.tex
%% ============================================================

Large language models (LLMs) are increasingly used for conversational clinical decision support, yet they conflate next token prediction with probabilistic decision making. We argue that this conflation reflects an architectural limitation: such systems lack explicit posterior tracking, controllable abstention thresholds, and auditable reasoning chains. We introduce \ours, a \underline{\textbf{Mo}}dular \underline{\textbf{Bayes}}ian dialogue framework that separates reasoning from language. The LLM acts only as a language interface, parsing patient conversation into structured observations, while a Bayesian module performs probabilistic inference over these observations to update posteriors, select follow-up questions via expected-information-gain and determine when to stop or defer through calibrated decision thresholds. This design enables explicit posterior tracking, controllable selective decision-making, and replaceable population-specific statistical backends without retraining the language model. Across empirical and LLM-generated knowledge bases, \ours outperforms standalone frontier LLM doctors, including matched model-family comparisons where inexpensive sensor models paired with \ours exceed larger autonomous models at lower cost. The advantage persists under adversarial patient communication styles and across varying diagnostic scenarios. These results suggest that reliable conversational clinical decision support systems should separate probabilistic reasoning from language generation rather than scaling model size alone. Code is available at~\url{https://anonymous.4open.science/r/MoBayes/}.
% When the GitHub repo (EPFLiGHT/MoBayes) is made public, swap to:
% Code is available at~\url{https://github.com/EPFLiGHT/MoBayes}.
\end{abstract}

%% --- Main sections -------------------------------------------
%% ============================================================
%% 1. INTRODUCTION
%% File: sections/introduction.tex
%%
%% STORYLINE (flowing prose, NeurIPS style):
%%   ¶1  Hook: conversation ≠ diagnosis + "never wrong" pitch
%%   ¶2  Classical stats worked, LLMs solved interface, but lost stats
%%   ¶3  Hybrids still let LLM touch probabilities (light citations)
%%   ¶4  Our approach: harsh separation + placeholder results
%%   Contributions (bulleted)
%% ============================================================

\section{Introduction}
\label{sec:intro}

Conversational clinical decision support (CCDS) is sequential abductive decision-making under uncertainty. Large language models are increasingly used in this role, yet they conflate two fundamentally different capabilities: natural-language communication and probabilistic inference, emulating abductive reasoning through next token prediction and token distributions rather than explicit posterior belief states. The result is systems that seem clinically fluent but lack the statistical grounding to make reliable decisions: no explicit probabilistic state, no principled information acquisition, uncalibrated confidence, and unauditable reasoning chains. The response is not better prompting or larger models but \emph{strict separation}: confine the LLM to language tasks and let a deterministic statistical model handle all probabilistic reasoning, with a tunable commit-or-defer rule.

Classical CDS demonstrated that structured rule based and probabilistic models can produce reliable clinical decisions~\cite{shortliffe1976,dedombal1972,miller1982,barnett1987,heckerman1992} but failed to scale due to rigid non-conversational interfaces; modern LLMs~\cite{singhal2023,yang2024medgemini,tu2025amie} have largely solved the interface problem by enabling natural flexible interaction, but at the cost of abandoning explicit probabilistic rigor (no formal posterior, calibrated abstention mechanisms or auditable reasoning chain). 
Recent work injects probabilistic structure into LLM-driven pipelines (\emph{probabilistic-LLM hybrids}) or wraps the LLM in multi-step orchestration (\emph{agent-based diagnostic systems})~\cite{choudhury2025,liu2024dellma,bird2024,zhu2025app,li2024mediq,meddxagent2025,doccha2025}. However, these approaches still rely on LLMs for probability estimation, making them susceptible to hallucination and pseudo-probabilistic reasoning derived from token distributions rather than explicit disease modelling. Other approaches fine-tune LLMs on clinical data to internalise medical statistics  ~\cite{kim2024finetuning}, raising concerns about robustness, privacy, data governance, and auditability while also limiting sustainability: updates require retraining, which can distort previously learned distributions with no formal guarantee the model faithfully represents the underlying distributions at inference time.

We argue that the right response is not larger LLMs but \emph{modular, statistically grounded reasoning}. We introduce \textbf{MoBayes}, a modular Bayesian framework that separates language from probabilistic reasoning. The LLM acts as a conversational interface, parsing patient interactions into structured observations and verbalizing questions selected by the Bayesian module. The Bayesian module performs inference over these observations, selects follow-up questions via expected information gain (EIG), and determines when to stop or defer using calibrated decision thresholds. These operations are defined over an explicit clinically grounded knowledge-base that can be adapted across populations without retraining the LLM, ensuring that no patient data is embedded in model weights.

Our main contributions are:

\textbf{Framework.} \ours constrains the LLM to a structured--evidence interface and delegates inference, question selection, and abstention to an explicit Bayesian module.

\textbf{Empirical advantage.} Across three CCDS benchmarks, MoBayes outperforms prior baselines and matches or exceeds same-family frontier models at substantially lower per-token cost. The advantage persists under adversarial patient personas and across increasing diagnostic complexity.

\textbf{Architectural attribution and selective control.} Controlled comparisons with shared knowledge sources, \wip{including KBs elicited entirely from the same frontier LLM (\S\ref{sec:results_llmkb}),} show that gains are architectural, not informational\wip{: the LLM is more usefully \emph{elicited once} into a tabular KB and reasoned over by the Bayesian engine than invoked turn-by-turn as the reasoner}. A single threshold $\tau$ further traces a continuous accuracy--coverage trade-off, enabling explicit control absent from standalone LLMs.

\begin{figure}[t]
\centering
\moveup
\begin{minipage}[c]{0.70\textwidth}
\vspace{0pt}
\centering
\includegraphics[width=\textwidth]{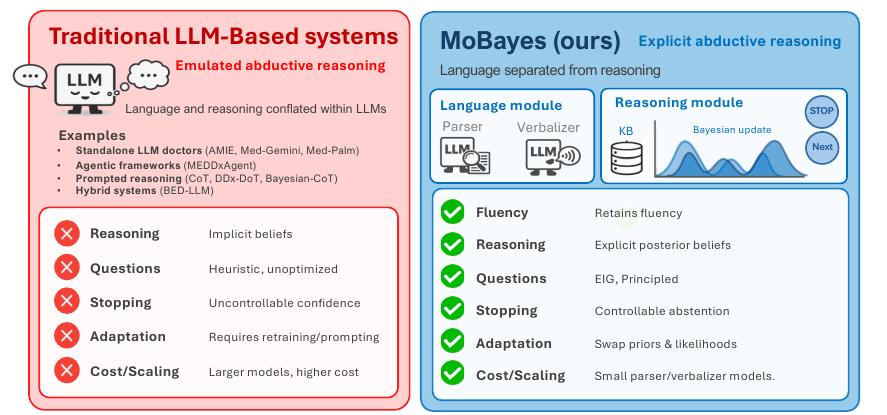}
\end{minipage}\hfill
\begin{minipage}[c]{0.28\textwidth}
\vspace{0pt}
\centering
\includegraphics[width=\textwidth,trim=0 0 0 -1.2cm, clip]{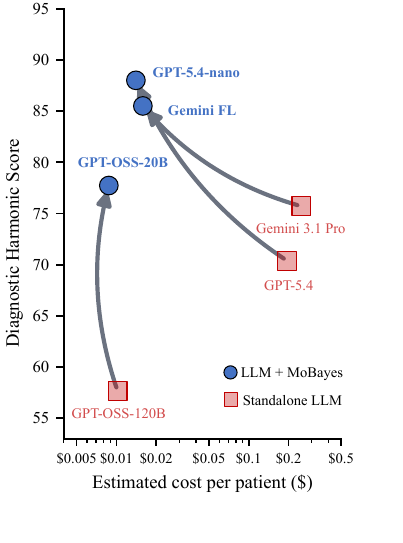}
\end{minipage}\hfill
\moveup
\figcaption{{\textbf{(left)} Traditional LLM systems (Red) compared with \ours (ours, blue). \textbf{(right)} Small sensor LLMs under \ours beat larger same-family standalone models at lower cost. (DDxPlus~50, PatientSim.)}}
\label{fig:paradigm_comparison}
\end{figure}

        % §1  Introduction
%% ============================================================
%% 2. RELATED WORK
%% File: sections/related_work.tex
%% ============================================================

\section{Related Work}
\label{sec:related}

\textbf{Statistical clinical decision support.}
Early CDSS established that structured probabilistic and rule based reasoning produces clinically useful diagnoses, using formalisms ranging from certainty-factor rules and na\"{\i}ve Bayes to belief networks, and causal models~\cite{shortliffe1976,yu1979,dedombal1972,miller1982,barnett1987,heckerman1992,weiss1978}. Despite strong accuracy, adoption stalled because these systems required structured pre-coded input and treated clinicians as data-entry operators, the ``Greek Oracle'' model~\cite{miller1990}. The bottleneck was \emph{interfacing}, not reasoning, motivating designs that preserve auditable probabilistic reasoning while replacing the rigid interface.

\textbf{End-to-end LLM-based diagnosis.}
Large language models substantially reduced the interfacing bottleneck by communicating in natural language. Med-PaLM~2~\cite{singhal2023} and Med-Gemini~\cite{yang2024medgemini} pushed standalone-LLM medical QA to expert level, and AMIE~\cite{tu2025amie} extended this to multi-turn dialogue, with subsequent end-to-end variants pursuing similar designs~\cite{xu2024reasoning,doccha2025}. These systems share a structural limitation: \emph{no formal probabilistic state}, no explicit posterior, no information-theoretic question selection, and no auditable abstention. Conversation, question choice, diagnosis, and stopping are instead entangled into a single generative model. The absence of structured reasoning carries safety costs: clinical LLMs exhibit poor calibration~\cite{savage2025calibration}, produce unstable reasoning across runs~\cite{potamitis2025reasonbench}, are vulnerable to adversarial hallucination~\cite{omar2025hallucination}, and encode geographic and racial biases~\cite{omiye2023race,olatunji2024afrimedqa}.

\textbf{Probability--language hybrids.}
A growing line of work augments LLM-driven pipelines with probabilistic or structured-reasoning components, including Bayesian experimental design, decision theory, active-learning planners, information-pursuit search, abstention-aware questioning, multi-agent decomposition, and structured reasoning~\cite{choudhury2025,liu2024dellma,bird2024,zhu2025app,sun2024planner,hu2024uot,li2024mediq,meddxagent2025,wei2022cot,klein2025foa}. In all of these systems the LLM remains \emph{inside} the diagnostic decision loop, as probability source, candidate generator, planner substrate, or latent reasoning state. An orthogonal line fine-tunes LLMs on clinical data~\cite{chen2024cod,kim2024finetuning}, but learned distributions are not directly auditable and raise data-governance and privacy concerns.

\textbf{Our position: strict separation.}
We propose \emph{strict separation} between language and clinical reasoning: the LLM handles \emph{only} language (parsing utterances into structured evidence and verbalising questions selected by the Bayesian module), while all belief updates, question selection, stopping, and abstention are delegated to an external, inspectable statistical engine. This separates the two components that prior work has historically traded off: classical CDSS-style auditable statistics and LLM-style conversational interface. The diagnostic state remains outside the LLM, patient data is never used for training or fine-tuning, and adapting to a new population requires only updating priors and likelihoods rather than re-aligning the LLM. Figure~\ref{fig:paradigm_comparison}a illustrates this taxonomy; Appendix~\ref{app:related_extended} expands on each of the families above and provides a system-by-system comparison in Table~\ref{tab:rw_comparison}.

        % §2  Related Work
%% ============================================================
%% 3. METHOD
%% File: sections/method.tex
%%
%% Structure follows Lars's 4-subsection proposal:
%%   3.1 Knowledge Base  |  3.2 Bayesian Inference Engine
%%   3.3 Orchestration   |  3.4 Patient-Facing Interface
%% ============================================================

\section{Method}
\label{sec:method}

We formulate diagnostic dialogue as a sequential Bayesian decision process. Let $\mathcal{D} = \{d_1, \dots, d_K\}$ denote the set of $K$ candidate diseases and $\mathcal{F} = \{f_1, \dots, f_N\}$ the set of $N$ observable clinical features, where each feature $f_i$ takes values in a finite set $\mathcal{V}_{f_i}$. At each turn $t$, the system maintains a posterior belief $\mathbf{b}_t \in \Delta^{K-1}$ over diseases, selects the most informative feature to query, updates its belief from the patient's response, and decides whether to terminate. The objective is to commit to a diagnosis when the posterior crosses a threshold $\tau$ and abstain otherwise, within the fixed question budget $T_{\rm max}$.

\ours decomposes clinical dialogue into two modular components: a \emph{language interface} and a \emph{Bayesian reasoning module}. The LLM acts only as a language layer, parsing patient utterances and verbalizing questions selected by the Bayesian module. The Bayesian module maintains posterior beliefs, selects questions, and renders clinical decisions. The two communicate exclusively through structured evidence triples $(f,\,v,\,c)$: feature identifier, schema-validated value, and confidence weight. All probabilistic reasoning is performed by the Baysian module, yielding a fully auditable decision trace. The complete procedure is given as Algorithm~\ref{alg:bmbe} in Appendix~\ref{app:bmbe_algorithm}.
\label{sec:architecture}

%% Figure 1: System architecture (Lars's TikZ)
\begin{figure}[htbp]
\centering
\moveup
\includegraphics[width=\textwidth]{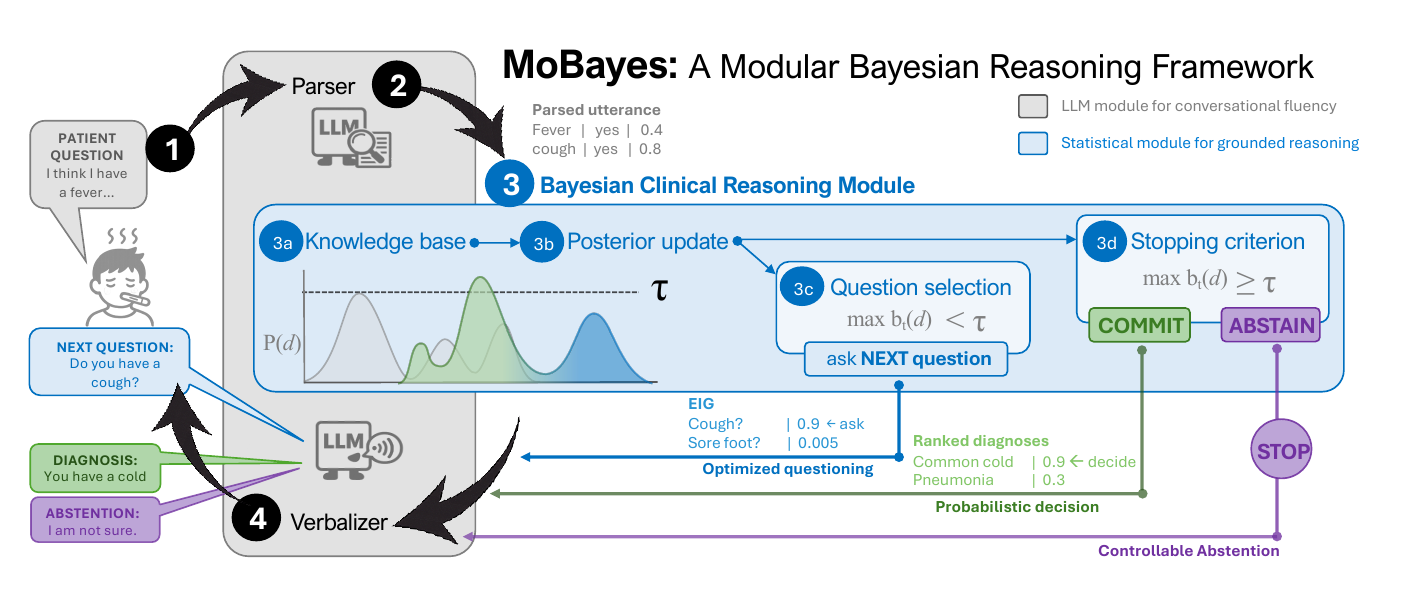}
\moveup
\moveups
\caption{Overview of the \ours architecture. The LLM acts only as a language layer (parses utterances into evidence triples $(f,v,c)$, verbalises questions selected by the Bayesian module); all reasoning, posterior updates, EIG-based question selection, stopping, and controllable abstention reside in the deterministic, auditable Bayesian module.}
\label{fig:architecture}
\end{figure}

%%% ---------------------------------------------------------------
%%% 3.1 KNOWLEDGE BASE
%%% ---------------------------------------------------------------
\subsection{Knowledge base}
\label{sec:kb}

The Bayesian module reasons over a tabular knowledge base of disease priors $\boldsymbol{\pi}$ and per-feature conditional likelihoods $P(X_f{=}v\mid d)$. We use two interchangeable construction paths that populate the same schema.

\textbf{Empirical KB (training data).} On DDxPlus we accumulate co-occurrence counts from the published 1.03M training records~\cite{ddxplus2022} and convert them into Dirichlet--Categorical likelihoods (full formulation in Appendix~\ref{app:kb}, Eq.~\ref{eq:dirichlet_mean}). The 49 pathologies and 314 features are taken directly from the dataset.

\textbf{LLM-elicited KB.} When labelled training data is unavailable, the same KB schema is populated by structured prompting of a frontier LLM in two stages, first feature catalogues per disease, then per-feature conditional distributions, with deduplication and validation. The motivation is modular separation: asking the LLM to externalise its knowledge into an explicit, auditable KB and perform inference separately through Bayesian updating. %This approach outperformsfeeding the same LLM the raw patient utterance directly (Section~\ref{sec:results_llmkb}). 
Two LLM-elicited KBs are used in this paper: (i) for the cross-KB study we draw the disease list from the BMRB Genes/Disease catalogue~\cite{hoch2023bmrb} restricted to the \emph{respiratory diseases} (6) and \emph{acquired diseases} (12) categories, yielding 18 diseases with a deliberate mix of symptomatically overlapping and clearly distinct conditions; (ii) for the AgentClinic-MedQA column we keep AgentClinic's disease list and elicit features and conditional likelihoods over it. Full prompts and validation rules are in Appendix~\ref{app:kb}.

% \textbf{Prior.} The initial belief $b_0(d)=\pi_d$ is the empirical training-set frequency for the empirical KB and uniform for the LLM-elicited KBs. Demographic conditioning is supported but not used in main experiments (Appendix~\ref{app:prior_sensitivity}).

%%% ---------------------------------------------------------------
%%% 3.2 BAYESIAN INFERENCE ENGINE
%%% ---------------------------------------------------------------
\subsection{Bayesian reasoning module}
\label{sec:engine}

Let $\mathcal{C} = \{c_1, \dots, c_5\} \subset (0, 1]$ denote the set of confidence weights and $\phi \colon \mathcal{L} \to \mathcal{C}$ the mapping from linguistic labels $\mathcal{L} = \{\textit{very\_likely}, \textit{likely}, \textit{uncertain}, \textit{unlikely}, \textit{very\_unlikely}\}$ to numerical weights. At each turn $t$, the patient's response about feature $f_t$ is parsed into a tuple $(v_t, c_t) \in \mathcal{V}_{f_t} \times \mathcal{C}$.

\paragraph{Soft evidence and belief update.}
Patient responses often hedge (``I think I had a fever''). Rather than treating every observation as hard evidence, we use a Pearl-style soft-evidence update~\cite{pearl1988probabilistic, chan2005revision}: a parser-supplied confidence $c\in(0,1]$ mixes the standard likelihood with a neutral vector. At $c{=}1$ we recover the standard Bayesian update; at $c{\to}0$ the evidence has vanishing effect. The five linguistic confidence labels and their numerical weights, the log-space normalisation, and the closed-form update equations are deferred to Appendix~\ref{app:jeffrey_theory} (Eqs.~\ref{eq:bayes_update}--\ref{eq:jeffrey_update}). For numeric features a Gaussian-weighted soft match replaces the point likelihood before the soft-evidence update is applied.

\paragraph{Question selection.}
At each turn the Bayesian module queries the unasked feature with the highest expected information gain (EIG), the standard myopic Bayes-optimal policy. We adopt the naive-Bayes factorisation $X_f\perp\mathbf{e}_{1:t}\mid D$; richer factorisations (tree-augmented naive Bayes~\cite{friedman1997tan}, noisy-OR~\cite{pearl1988probabilistic}, latent comorbidity factors~\cite{sucar2015pgm}, sum-product networks~\cite{poon2011spn}) plug in without altering belief representation, EIG, or abstention, at the cost of more complex inference and a quadratic blow-up in the LLM-querying budget for KB construction. Per-turn cost is linear in $|\mathcal{F}|$, $\max_f|\mathcal{V}_f|$, and $K$. Closed forms for EIG, the counterfactual posterior, computational cost, and policy ablations (hierarchical EIG, top-$k$ focused EIG, confidence-activated hybrid) are in Appendix~\ref{app:eig_ablations} (Eqs.~\ref{eq:eig}--\ref{eq:eig_focused}).

%%% ---------------------------------------------------------------
%%% 3.3 ORCHESTRATION
%%% ---------------------------------------------------------------
\subsection{Orchestration: ending the consultation}
\label{sec:stopping}

The system must decide not only \emph{what} to diagnose but \emph{whether} to diagnose at all. After each belief update we evaluate stopping conditions in strict priority: a warm-up period of $T_{\min}$ turns prevents premature commitment while early posteriors remain dominated by the prior; once warm-up elapses, the system commits to $d^* = \arg\max_d b_t(d)$ as soon as $\max_d b_t(d) \geq \tau$; a hard budget $T_{\max}$ guarantees termination, abstaining if no disease has crossed $\tau$ by then (budget values are in Appendix~\ref{app:stopping_params}). The threshold $\tau$ is itself a deployment parameter that gives providers explicit control over the accuracy--coverage trade-off, from high-throughput triage ($\tau$ low, high coverage) to safety-critical referral ($\tau$ high, near-zero error on committed cases), without retraining; autonomous LLM doctors offer no equivalent mechanism, and we characterise this trade-off in Section~\ref{sec:results_operating}.

%%% ---------------------------------------------------------------
%%% 3.4 PATIENT-FACING INTERFACE
%%% ---------------------------------------------------------------
\subsection{Patient-facing interface}
\label{sec:llm_interface}

The LLM serves as a pure language layer with no access to the posterior distribution or diagnostic state (prompt templates in Appendix~\ref{app:prompts}).

\paragraph{Response parsing.}
Given a patient utterance $u_t$ and the queried feature $f_t$ with schema $\mathcal{S}_{f_t} = (\mathcal{V}_{f_t}, \mathcal{C})$, the parser produces $(v_t, c_t) \in \mathcal{V}_{f_t} \times \mathcal{C}$ via a three-tier cascade. First, \emph{deterministic pattern matching} maps clear affirmative, negative, and hedging expressions to $(v, c)$ pairs at zero API cost. When this tier abstains on ambiguous input, an \emph{LLM-based extractor} classifies the response into the schema's value set with a confidence label from $\mathcal{L}$, mapped to a numerical weight via $\phi$. The key design choice is that the LLM performs \emph{structured classification} into $\mathcal{V}_f \times \mathcal{L}$, a small, fixed label set, rather than continuous probability estimation. LLMs exhibit strong ordinal judgement~\cite{liu2024dellma} but poor calibration for continuous probabilities~\cite{kadavath2022}; the discrete-to-numerical mapping $\phi$ shifts the calibration burden from continuous probability estimation to a coarser ordinal task that LLMs handle more reliably. From the Bayesian module's perspective, each $(v, c)$ pair is a noisy observation, and the LLM functions as a discretized probabilistic model rather than a reasoning system.
\paragraph{Question verbalisation.}
Given the EIG-selected feature $f^*$, the LLM receives its name, answer schema $\mathcal{S}_{f^*}$, and a coarse confidence indicator that modulates conversational tone only; the LLM never receives $\mathbf{b}_t$, $H(\mathbf{b}_t)$, or any other internal model state.
              % §3  Method
%% ============================================================
%% 4. EXPERIMENTAL SETUP
%% File: sections/experiments.tex
%% ============================================================

\section{Experimental Setup}
\label{sec:experiments}

% ──────────────────────────────────────────────────────────────
% STORYLINE: Datasets → Patient simulation (personas) → Baselines
%            → Evaluation metrics.
% ──────────────────────────────────────────────────────────────

%% ,  4.1 Datasets --------------------------------------------
\subsection{Datasets}
\label{sec:datasets}

We evaluate on three diagnostic-dialogue benchmarks. \textbf{DDxPlus}~\cite{ddxplus2022} is a large-scale synthetic dataset of ${\sim}1.3$M patient records over 49 diseases and 314 features with a published train/test split; the empirical KB is built from the training split and evaluation uses held-out test patients. To isolate architectural from informational advantages, we additionally evaluate on an \textbf{LLM-elicited KB} whose likelihoods are elicited \emph{entirely} from a frontier LLM (Section~\ref{sec:kb}) in two ecosystems (GPT-5.4 and Gemini~3.1) over a disease list drawn from the BMRB Genes/Disease catalogue~\cite{hoch2023bmrb} restricted to the respiratory and acquired-diseases categories; the same LLM also serves as the standalone doctor in our fair-fight experiments. \textbf{AgentClinic-MedQA}~\cite{schmidgall2024agentclinic} is a USMLE-derived multi-agent dialogue benchmark in which the doctor LLM is evaluated through interaction with the suite's own patient, measurement, and moderator agents. KB construction, case curation, and elicitation protocols are detailed in Appendices~\ref{app:kb}, \ref{app:baselines_protocol}, and~\ref{app:ac_medqa_curation}.

%% ,  4.2 Patient simulation ----------------------------------
\subsection{Patient simulation}
\label{sec:patient_sim}

Evaluating a diagnostic dialogue system requires simulated patients that respond realistically to follow-up questions; several frameworks address this challenge~\cite{medagentsim2025,aihospital2025,patientsim2025,autointeractiveeval2024}. We use three published patient simulators across our experiments, each corresponding to a different benchmark setting.

\textbf{PatientSim.} For the DDxPlus PatientSim column we adopt the PatientSim simulator~\cite{patientsim2025} with minor interface-level adaptations to plug it into our DDxPlus profile schema; the persona taxonomy (personality, language proficiency, recall accuracy, cognitive state) is unchanged. Main experiments use the cooperative \emph{plain} persona; adversarial-persona results are in Appendix~\ref{app:persona_full}.

\textbf{MEDDxAgent history-taking simulator.} For the second DDxPlus column we use MEDDxAgent's published history-taking simulator~\cite{meddxagent2025} on a separate DDxPlus subset, with their orchestrator (DDxDriver) and bench class invoked from their open-source codebase unmodified.

\textbf{AgentClinic patient agent.} As an orthogonal, non-DDxPlus column, we use AgentClinic's published patient and measurement components~\cite{schmidgall2024agentclinic} on AgentClinic-MedQA, with their dialogue protocol and native moderator unchanged.

Each simulator receives the ground-truth profile and never accesses the Bayesian module or KB. Our adaptations are interface-level only and preserve each simulator's dialogue protocol; details in Appendix~\ref{app:patient_sim_adaptations}.

%% ,  4.3 Models and baselines --------------------------------
\subsection{Models and baselines}
\label{sec:baselines}

We compare \discuss{\ours} against three families of prior diagnostic dialogue systems, all run under the same patient simulator and disease universe per benchmark column. Full specifications, prompts, and per-method hyperparameters are in Appendix~\ref{app:baselines_protocol}; Table~\ref{tab:models} lists the LLM backbones.

\textbf{\discuss{\ours} variants.} Each variant pairs the Bayesian engine with a different LLM sensor; since the LLM only parses and verbalises, no medical specialisation is needed. We test six inexpensive backends across providers and architectures (Table~\ref{tab:models}); only the sensor LLM changes between variants.

\textbf{Standalone frontier LLM doctors.} The standalone baseline gives the LLM full control over the clinical dialogue: it decides what to ask, when to stop, and what to diagnose, with no external reasoning aid. We evaluate six frontier models (Table~\ref{tab:models}); the prompt is in Appendix~\ref{app:prompts} and the closed-set normalisation protocol used to score free-text predictions in Appendix~\ref{app:standalone_protocol}.

\textbf{Prior diagnostic-dialogue systems.} We evaluate published systems spanning the four families surveyed in Section~\ref{sec:related}: \emph{prompt-based} (Chain-of-Thought~\cite{wei2022cot}, DDx-CoT and Bayesian-CoT~\cite{savage2024diagreasoning}); \emph{information-pursuit} (UoT~\cite{hu2024uot}); \emph{agent-based} (MediQ~\cite{li2024mediq}, which is MCQ-bound; we split each benchmark into two disjoint subsets to keep its option pool tractable, and for fairness all other baselines are run on the same two-subset partition; MEDDxAgent~\cite{meddxagent2025} in its paper-best configuration; AgentClinic's own doctor agent~\cite{schmidgall2024agentclinic}); and a \emph{fine-tuned} reference (DiagnosisGPT-34B / Chain-of-Diagnosis~\cite{chen2024cod}). We additionally include three configurations of our own to round out the comparison: a \emph{Closed-World} prompt-based baseline, a \emph{strict-abstain} scoring variant of MediQ, and a \emph{closed-world ablation} of MEDDxAgent. Per-method hyperparameters and the rationale for each added configuration are detailed in Appendix~\ref{app:baselines_protocol}. Every non-fine-tuned baseline runs with the same gpt-5.4-nano backbone we use for our headline \discuss{\ours} sensor, so any per-column differences isolate architectural rather than backbone effects.

%% ,  4.4 Evaluation metrics ----------------------------------
\subsection{Evaluation metrics}
\label{sec:metrics}

Following the selective prediction framework~\cite{geifman2017selective}, a selection function $g(x) \in \{0,1\}$ decides whether to commit ($g{=}1$) or abstain ($g{=}0$) based on posterior confidence exceeding~$\tau$. We report two families of metrics: \textbf{Top-$k$ accuracy} ($k \in \{1,3\}$) measures diagnostic correctness over all patients, counting abstentions as errors; and the \emph{commit-or-abstain} \textbf{selective diagnosis} metrics defined below.

\paragraph{Selective diagnosis metrics.}
The \emph{commit-or-abstain} trade-off is summarised by selective accuracy, coverage, and the \textbf{Diagnostic Harmonic Score (DHS)}:
\begin{equation}
\text{Sel.\,Acc} = \frac{\sum_i g(x_i)\,\mathbb{1}[f(x_i){=}y_i]}{\sum_i g(x_i)},
\quad
\text{Cov} = \frac{1}{n}\sum_i g(x_i),
\quad
\text{DHS}_\alpha = \left(\frac{\alpha}{\text{Sel.\,Acc}} + \frac{1-\alpha}{\text{Cov}}\right)^{\!-1}.
\label{eq:metrics}
\end{equation}
At $\alpha{=}0.5$ (used throughout) DHS is the harmonic mean of SelAcc and Coverage; the limits $\alpha{\to}1$ and $\alpha{\to}0$ recover SelAcc and Coverage respectively. $\tau$ is a \emph{deployment parameter}: providers shift along the accuracy--coverage curve to match clinical risk tolerance (Section~\ref{sec:results_operating}).

         % §4  Experimental Setup
%% ============================================================
%% 5. RESULTS
%% File: sections/results.tex
%% ============================================================

\section{Results}
\label{sec:results}

\subsection{Comparison with prior diagnostic-dialogue systems}
\label{sec:results_prior}

Table~\ref{tab:combined_n100} reports the head-to-head comparison against the baselines of Section~\ref{sec:baselines}. \discuss{\ours} achieves the highest DHS on both DDxPlus columns and remains competitive with the best agent-based baseline on AgentClinic-MedQA, while non-\discuss{\ours} methods vary substantially across columns and the fine-tuned reference (DiagnosisGPT-34B) underperforms across all columns.

% \input{tables/tab_baselines}                  % Table 1 — moved to appendix
% \input{tables/tab_combined_n100}              % Table 2 — moved to appendix
%% =============================================================
%% TABLE: Main paper headline — combined n=100, 4 metrics.
%% Auto-generated by scripts/render_paper_tables.py.
%% =============================================================

\begin{table}[t]
\centering
\caption{Combined $n{=}100$ comparison of \discuss{\ours} vs prior diagnostic-dialogue systems. Rows ranked within each category by average DHS. \discuss{\ours} entries marked $^*$ are statistically significant (paired bootstrap, $p{<}0.05$) over the strongest non-\ours\ baseline in the same column. Per-cell 95\% bootstrap CIs in Appendix~Table~\ref{tab:ci_combined}. AgentClinic-MedQA uses a GPT-5.4-nano-elicited KB (Appendix~\ref{app:kb}). DHS coloured per column: the \highlightgreen{greener} the \highlightgreen{better}, the \highlightred{redder} the \highlightred{worse}.}
\label{tab:main_combined}
\footnotesize
\setlength{\tabcolsep}{4pt}
\renewcommand{\arraystretch}{1.05}
\resizebox{\textwidth}{!}{%
\begin{tabular}{@{}l cccc @{\hskip 0.6em} cccc @{\hskip 0.6em} cccc@{}}
\toprule
 & \multicolumn{4}{c}{\textbf{DDxPlus 100 (\ours PatientSim)}} & \multicolumn{4}{c}{\textbf{DDxPlus 100 (HT-sim)}} & \multicolumn{4}{c}{\textbf{AgentClinic-MedQA 100}} \\
\cmidrule(lr){2-5} \cmidrule(lr){6-9} \cmidrule(lr){10-13}
\textbf{Method} & T-1 & SA & DHS$\,\uparrow$ & Tok/case & T-1 & SA & DHS$\,\uparrow$ & Tok/case & T-1 & SA & DHS$\,\uparrow$ & Tok/case \\
\midrule
\multicolumn{13}{l}{\textit{Prompt-based}} \\
\midrule
DDx-CoT                            & 42 & 51 & \cellcolor{dhsred!8}55 & 28.8K & 37 & 78 & \cellcolor{dhsred!22}16 & 18.5K & 40 & 78 & \cellcolor{dhsred!19}62 & 23.7K \\
Closed-World (CW)                  & 40 & 50 & \cellcolor{dhsgreen!8}55 & 32.0K & 45 & 75 & \cellcolor{dhsred!20}21 & 23.9K & 40 & 75 & \cellcolor{dhsred!19}62 & 25.7K \\
Chain-of-Thought (CoT)             & 52 & 55 & \cellcolor{dhsgreen!21}69 & 30.8K & 41 & 63 & \cellcolor{dhsred!17}29 & 15.8K & 71 & \textbf{89} & \cellcolor{dhsgreen!31}84 & 22.5K \\
Bayesian-CoT                       & 60 & 62 & \cellcolor{dhsgreen!25}74 & 46.6K & 41 & 58 & \cellcolor{dhsred!10}45 & 34.1K & 63 & 77 & \cellcolor{dhsgreen!21}79 & 37.8K \\
\midrule
\multicolumn{13}{l}{\textit{Information-pursuit}} \\
\midrule
UoT                                & 28 & 28 & \cellcolor{dhsred!13}44 & 37.7K & 5 & 5 & \cellcolor{dhsred!25}10 & 28.2K & 38 & 40 & \cellcolor{dhsred!25}56 & 34.6K \\
\midrule
\multicolumn{13}{l}{\textit{Agent-based}} \\
\midrule
MediQ-Expert (strict abstain)      & 32 & 51 & \cellcolor{dhsgreen!9}56 & 39.3K & 25 & 57 & \cellcolor{dhsred!8}50 & 41.9K & 37 & 84 & \cellcolor{dhsred!23}58 & 42.6K \\
MediQ-Expert (force-final)         & 43 & 43 & \cellcolor{dhsgreen!13}60 & 39.3K & 34 & 34 & \cellcolor{dhsgreen!9}51 & 41.9K & 56 & 56 & \cellcolor{dhsred!9}72 & 42.6K \\
AgentClinic                        & 27 & 51 & \cellcolor{dhsred!10}52 & 18.1K & 28 & 44 & \cellcolor{dhsgreen!10}52 & 29.7K & 66 & 67 & \cellcolor{dhsgreen!22}80 & 6.2K \\
MEDDxAgent (closed-world abl.)     & 41 & 41 & \cellcolor{dhsgreen!11}58 & 79.9K & 39 & 39 & \cellcolor{dhsgreen!13}56 & 71.9K & \textbf{80} & 80 & \cellcolor{dhsgreen!40}\textbf{89} & 69.0K \\
MEDDxAgent (paper-best)            & 45 & 45 & \cellcolor{dhsgreen!14}62 & 108.3K & 56 & 56 & \cellcolor{dhsgreen!26}72 & 88.1K & 71 & 71 & \cellcolor{dhsgreen!29}83 & 106.8K \\
\midrule
\multicolumn{13}{l}{\textit{Fine-tuned (specialised model)}} \\
\midrule
CoD / DiagnosisGPT-34B             & 11 & 11 & \cellcolor{dhsred!25}20 & 13.4K & 4 & 6 & \cellcolor{dhsred!25}11 & 12.7K & 46 & 78 & \cellcolor{dhsred!13}67 & 31.2K \\
\midrule
\multicolumn{13}{l}{\textit{Ours}} \\
\midrule
\textbf{\ours} (gpt-5.4-nano)      & \textbf{82} & \textbf{85} & \cellcolor{dhsgreen!40}\textbf{90}$^*$ & 16.4K & \textbf{81} & \textbf{87} & \cellcolor{dhsgreen!40}\textbf{90}$^*$ & 18.0K & 70 & 74 & \cellcolor{dhsgreen!29}83 & 18.9K \\
\bottomrule
\end{tabular}%
}
\end{table}
                 % new main-paper headline table

\subsection{Comparison with frontier standalone LLMs}
\label{sec:results_main}

%% ============================================================
%% TABLE 3: \ours sensors vs frontier standalone LLM doctors
%% across all three diagnostic-dialogue benchmarks (n=50 per cell).
%% Reproducible from raw JSONs by `python3 scripts/table21.py`;
%% per-cell 95% CIs in Table~\ref{tab:ci_main}.
%% ============================================================

\begin{table}[t]
\centering
\caption{\discuss{\ours} sensors vs.\ standalone LLM doctors on the three benchmarks of Section~\ref{sec:datasets}; $n{=}50$ per cell. \discuss{\ours} uses DHS-optimal $\tau^*$ per sensor (Appendix~\ref{app:threshold_tuning}). DHS coloured per benchmark column. Per-cell 95\% bootstrap CIs are reported in Appendix Table~\ref{tab:ci_main}; the AgentClinic-MedQA results use a knowledge base elicited from GPT-5.4-nano (construction protocol in Appendix~\ref{app:kb}).}
\label{tab:main}
\footnotesize
\setlength{\tabcolsep}{4pt}
\renewcommand{\arraystretch}{1.05}
\begin{tabular}{@{}l ccc @{\hskip 0.7em} ccc @{\hskip 0.7em} ccc@{}}
\toprule
& \multicolumn{3}{c}{\textbf{DDxPlus 50 (PatientSim)}} & \multicolumn{3}{c}{\textbf{DDxPlus 50 (HT-sim)}} & \multicolumn{3}{c}{\textbf{AgentClinic-MedQA 50}} \\
\cmidrule(lr){2-4} \cmidrule(lr){5-7} \cmidrule(lr){8-10}
\textbf{System}
  & T-1 & SA & DHS$\,\uparrow$
  & T-1 & SA & DHS$\,\uparrow$
  & T-1 & SA & DHS$\,\uparrow$ \\
\midrule
\multicolumn{10}{l}{\textit{Standalone LLM doctors (frontier)}} \\
SA GPT-OSS-120B         & 40 & 40.8 & \cellcolor{dhsred!25}57.6 & 54 & 54.5 & \cellcolor{dhsred!15}59.7 & 48 & 49.0 & \cellcolor{dhsred!25}65.3 \\
SA Kimi K2.5            & 42 & 54.1 & \cellcolor{dhsred!20}62.5 & 42 & 40.7 & \cellcolor{dhsred!25}46.4 & 54 & 57.4 & \cellcolor{dhsred!16}71.3 \\
SA Qwen 3.6 Plus        & 48 & 51.1 & \cellcolor{dhsred!15}66.2 & 60 & 65.0 & \cellcolor{dhsgreen!12}71.7 & 64 & 63.3 & \cellcolor{dhsgreen!8}76.9 \\
SA Llama-4-Maverick     & 56 & 56.5 & \cellcolor{dhsred!11}70.0 & 46 & 56.0 & \cellcolor{dhsred!20}52.8 & 56 & 59.6 & \cellcolor{dhsred!14}72.9 \\
SA GPT-5.4              & 60 & 64.1 & \cellcolor{dhsred!11}70.4 & 58 & 76.0 & \cellcolor{dhsred!14}60.3 & 60 & 62.8 & \cellcolor{dhsred!14}72.6 \\
SA Gemini 3.1 Pro       & 60 & 62.5 & \cellcolor{dhsgreen!14}75.7 & 60 & 73.2 & \cellcolor{dhsgreen!20}77.3 & 66 & 70.2 & \cellcolor{dhsgreen!18}80.4 \\
\midrule
\multicolumn{10}{l}{\bf \textit{\discuss{\ours} + sensor LLM (ours)}} \\
+ Llama-4-Scout       & 56 & 60.9 & \cellcolor{dhsgreen!9}73.3 & 80 & \textbf{87.0} & \cellcolor{dhsgreen!37}89.4 & 68 & 79.1 & \cellcolor{dhsgreen!24}82.4 \\
+ Gemma 4 31B         & 58 & 61.7 & \cellcolor{dhsgreen!12}74.5 & 82 & 83.7 & \cellcolor{dhsgreen!38}90.3 & 64 & 65.3 & \cellcolor{dhsgreen!12}78.4 \\
+ MiniMax M2.5        & 60 & 66.7 & \cellcolor{dhsgreen!16}76.6 & 60 & 75.0 & \cellcolor{dhsgreen!20}77.4 & 50 & 52.1 & \cellcolor{dhsred!22}67.5 \\
+ GPT-OSS-20B         & 60 & 68.2 & \cellcolor{dhsgreen!16}76.8 & 50 & 61.0 & \cellcolor{dhsgreen!9}69.9 & 50 & 58.1 & \cellcolor{dhsred!19}69.4 \\
+ Gemini 3.1 Flash Lite & 74 & 77.1 & \cellcolor{dhsgreen!35}85.5 & \textbf{84} & 85.7 & \cellcolor{dhsgreen!40}\textbf{91.4} & \textbf{78} & \textbf{86.7} & \cellcolor{dhsgreen!40}\textbf{88.3} \\
+ GPT-5.4-nano        & \textbf{78} & \textbf{81.2} & \cellcolor{dhsgreen!40}\textbf{88.0} & 80 & 83.3 & \cellcolor{dhsgreen!37}89.2 & 76 & 79.2 & \cellcolor{dhsgreen!36}86.8 \\
\bottomrule
\end{tabular}
\end{table}

Table~\ref{tab:main} compares six \discuss{\ours} model variants against six frontier standalone LLM doctors across all three diagnostic-dialogue benchmarks. Every \discuss{\ours} variant outperforms every standalone doctor on DHS on at least two of the three benchmarks, and the headline variant (\discuss{\ours}~+~GPT-5.4-nano) leads on all three. The best \ours variant achieves the highest DHS at a fraction of the per-token cost of the best standalone doctor. The advantage is architectural: every \ours variant achieves higher point-estimate DHS than every standalone doctor on DDxPlus, and the headline best-vs-best gap is large.

\subsection{LLM-generated knowledge base}
\label{sec:results_llmkb}

A natural objection is that \discuss{\ours}'s advantage stems from access to data-derived statistics unavailable to the standalone doctor. To control for this, we replace the empirical DDxPlus KB with ones generated entirely by LLMs. In each ecosystem the same frontier LLM populates the KB and, in a separate elicitation, generates evaluation patients (Appendix~\ref{app:synthetic_patients}); both are noisy realisations of the LLM's beliefs. An inexpensive same-family \discuss{\ours} sensor matches the frontier standalone's Sel.\,Acc in both ecosystems at roughly 10x lower per-token cost, and the standalone drops sharply on rare diseases drawn from its own world model (Appendix~\ref{app:llmkb_full}).

\subsection{Cost-effectiveness}
\label{sec:results_cost}

\begin{figure}[h]
\centering
\moveups
\begin{subfigure}[t]{0.36\textwidth}
\centering
\includegraphics[height=5.5cm, trim=0 0.8cm 0 0, clip]{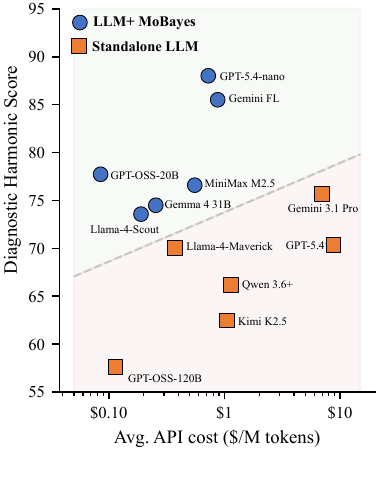} % doi %
\label{fig:cost_combined:a}
\end{subfigure}\hfill
\begin{subfigure}[t]{0.36\textwidth}
\centering
\includegraphics[height=5.5cm, trim=0 0.6cm 0 0, clip]{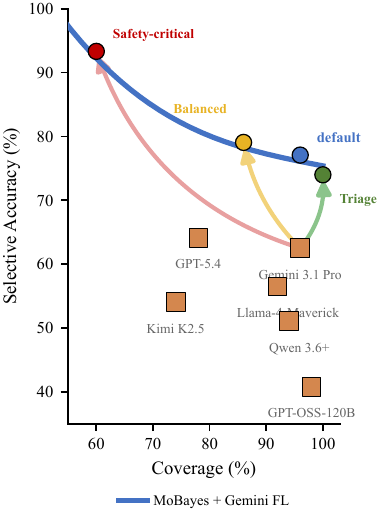}
\label{fig:cost_combined:b}
\end{subfigure}\hfill
\begin{subfigure}[t]{0.27\textwidth}
\centering
\includegraphics[height=5.5cm]{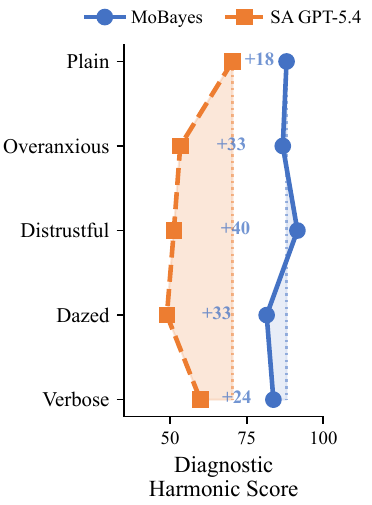}
\label{fig:cost_combined:c}
\end{subfigure}
\vspace{1mm}
\figcaption{\textbf{(a)} \textbf{Cost--accuracy tradeoff.} DHS vs.\ per-token cost: \discuss{\ours} models (circles) outperform standalone doctors (squares) at lower cost (Section~\ref{sec:results_cost}). \textbf{(b)} \textbf{Controllable selective diagnosis.} Sweeping $\tau$ traces the accuracy--coverage frontier for \discuss{\ours}~+~Gemini~FL; arrows mark deployment modes against SA Gemini~3.1~Pro (Section~\ref{sec:results_operating}). \wip{\textbf{(c)} \textbf{Persona robustness.} DHS across five PatientSim personas; shaded bands mark degradation from \emph{Plain} baseline (Appendix~\ref{app:persona_full}).} (DDxPlus~50, PatientSim.)}
\label{fig:cost_combined}
\end{figure}

A key advantage of \ours is that the Bayesian module performs all diagnostic reasoning deterministically at zero API cost; the LLM handles only parsing and verbalisation. Figure~\ref{fig:cost_combined} (left) shows DHS vs.\ per-token API pricing across model scales: \ours models cluster in the top-left quadrant (cheap and accurate) while frontier standalone doctors occupy the bottom-right. The corresponding per-patient cost view, which folds in dialogue length and per-turn token routing, is reported in Figure~\ref{fig:cost_per_patient} of Appendix~\ref{app:cost_analysis}. Figure~\ref{fig:paradigm_comparison}b isolates this effect within three model families. Each arrow shows the result of replacing a frontier standalone doctor with a cheap \ours models from the same family; every arrow points towards lower cost and higher DHS.

\subsection{Controllable selective diagnosis}
\label{sec:results_operating}

A key property of \ours is that deployers can choose their operating point along the accuracy, coverage curve by adjusting $\tau$. Figure~\ref{fig:cost_combined} (right) illustrates the contrast: \ours traces a continuous curve as $\tau$ varies, while standalone doctors (\texttimes) are locked at a single, uncontrollable point. The labelled modes mark representative deployment points. This controllability corresponds to shifting the emphasis parameter $\alpha$ in Eq.~\ref{eq:metrics}: $\alpha$ towards~1 penalises errors (safety-critical referral), while $\alpha$ towards~0 penalises abstention (high-throughput triage). Standalone LLMs offer no equivalent: they either diagnose or abstain based on the model's internal, uncalibrated confidence.

\wipblock{\subsection{Robustness across patient personas}
\label{sec:results_robustness}

A clinically deployed system must remain reliable under noisy, evasive, or confused patient communication. Figure~\ref{fig:cost_combined}c compares DHS across five PatientSim personas: standalone GPT-5.4 collapses sharply under adversarial styles (distrustful, dazed, verbose, overanxious), while \ours stays nearly stable because the Bayesian backbone accumulates evidence independently of the patient's surface communication style (full breakdown in Appendix~\ref{app:persona_full}).}

\subsection{Additional results}

We report further experiments in the appendix: confidence threshold tuning and accuracy, coverage trade-offs (Appendices~\ref{app:threshold_tuning},~\ref{app:acc_coverage}), question-selection policy ablations (Appendix~\ref{app:policy_ablations}), prior distribution sensitivity (Appendix~\ref{app:prior_sensitivity}), and disease scaling analysis examining how performance varies with the number of candidate diseases (Appendix~\ref{app:scaling}).
             % §5  Results
%% ============================================================
%% 7. DISCUSSION
%% File: sections/discussion.tex
%% ============================================================

\section{Discussion}
\label{sec:discussion}

% ──────────────────────────────────────────────────────────────
% STORYLINE: Key findings → Why separation works → Limitations
%            → Broader impact.
% ──────────────────────────────────────────────────────────────

\paragraph{Why strict separation works.}
Diagnosis is fundamentally an abductive probabilistic problem: sequentially updating competing disease hypotheses as evidence accumulates. LLMs lack persistent belief state, cannot compute information-theoretic objectives, and produce uncalibrated confidence through implicit token distributions rather than explicit posterior inference. By delegating these tasks to a deterministic Bayesian engine, \discuss{\ours} gains exact posterior tracking, principled question selection, and controllable abstention. The thesis is supported across multiple experiments: \discuss{\ours} achieves the highest DHS on both DDxPlus columns of Table~\ref{tab:combined_n100}, remains competitive on AgentClinic-MedQA, and matches or exceeds same-family frontier standalone LLMs, including in the fair-fight setting where both draw from the same LLM's beliefs (Appendix~\ref{app:synthetic_patients}). The modular architectural choice also carries into cost-effectiveness: \discuss{\ours} consumes notably few tokens per session, matching or exceeding same-family frontier standalones at a fraction of the per-session cost (Figure~\ref{fig:cost_combined}; breakdown in Appendix~\ref{app:cost_analysis}), with within-family pairings consistently in our favour. Standalone LLMs produce a single take-it-or-leave-it diagnosis. \discuss{\ours} traces a continuous accuracy--coverage frontier via a single threshold $\tau$ (Section~\ref{sec:results_operating}), with no retraining. The Bayesian backbone is also notably robust to adversarial patient communication styles where standalone doctors collapse: under PatientSim's distrustful, dazed, and overanxious personas, \discuss{\ours} maintains stable DHS while the standalone DHS drops sharply (Appendix~\ref{app:persona_full}). The same robustness extends to low-prevalence diseases (Appendix~\ref{app:prevalence_analysis}) and to larger candidate disease spaces (Appendix~\ref{app:scaling}).

\paragraph{Limitations.}
\discuss{\ours} operates under a \emph{closed-world assumption}: it can only diagnose diseases present in the knowledge base, whereas standalone LLMs can draw on open-ended medical knowledge. KB coverage is therefore a critical deployment concern, though the abstention mechanism provides a safety valve and the LLM-elicited KB experiment (Section~\ref{sec:results_llmkb}) shows that constructing a KB does not require labelled clinical data. We also rely on simulated patients rather than real clinical interactions, and the current architecture cannot exploit volunteered evidence that falls outside the asked question, a behaviour standalone LLMs handle naturally by reading the full conversation.

\paragraph{Broader impact.}
\discuss{\ours} is designed for clinical decision \emph{support}, rather than autonomous diagnosis: the abstention mechanism declines uncertain cases rather than forcing a position, and the full decision trace (prior, evidence, posterior per turn) is auditable. The LLM serves only as a language interface and is never trained on patient data, so no patient information is embedded in model weights. Deploying in a new population requires only a local KB, not retraining, making the framework practical for low-resource settings where clinical data exists but compute for LLM fine-tuning does not.
          % §6  Discussion
%% ============================================================
%% 8. CONCLUSION
%% File: sections/conclusion.tex
%% ============================================================

\section{Conclusion}
\label{sec:conclusion}

We presented \discuss{\ours}, a diagnostic dialogue framework built on a strict separation principle: an LLM handles only language, while a deterministic Bayesian module handles all probabilistic reasoning over an explicit knowledge base. The architecture is auditable, architecturally private, and deployable in low-resource settings: swapping the KB adapts the system to a new population with no retraining of the language model. 
Across multiple benchmarks, \discuss{\ours} matches or exceeds diagnostic-dialogue baselines and same-family frontier standalone LLMs at a small fraction of the cost. The architectural origin of the advantage is supported by control setting in which \discuss{\ours} and the standalone doctor draw clinical knowledge from the same frontier LLM, yet \discuss{\ours} still outperforms the standalone system. 
Our broader conclusion is that reliable conversational clinical decision support should not rely on implicit reasoning \del{emerging }from next-token prediction alone\del{. Instead, language and inference should be modularized}\wip{:} LLMs \replace{provide flexible communication, while explicit probabilistic systems perform calibrated reasoning under uncertainty}{for communication, explicit probabilistic systems for calibrated reasoning}.
          % §7  Conclusion

%% --- Acknowledgments -----------------------------------------
\begin{ack}
This work was supported under project ID \#27 as part of the Swiss AI Initiative, through a grant from the ETH Domain and computational resources provided by the Swiss National Supercomputing Centre (CSCS) under the Alps infrastructure. Arora's lab is partly supported by grants from the Novo Nordisk Foundation (NNF24OC0099109), the Pioneer Centre for AI, and EU Horizon 2020 (101168951). We also gratefully acknowledge generous gifts from Microsoft and IT-vest - networking universities.
\end{ack}

%% --- References ----------------------------------------------
\bibliographystyle{unsrtnat}
\newpage
\bibliography{references}

%% --- Appendix ------------------------------------------------
% \newpage
%% ============================================================
%% APPENDIX — Master file
%% File: sections/appendix.tex
%%
%% Structure:
%%   A. Theoretical Foundations  (theory/)
%%   B. System Details           (system/)
%%   C. Extended Experiments     (results/)
%% ============================================================

\appendix

%% ================================================================
%% APPENDIX TABLE OF CONTENTS
%% ================================================================

\section*{Appendix overview}
\label{app:toc}

\noindent
This supplementary material is organized as follows:

\vspace{1em}

\begin{itemize}[leftmargin=2em, itemsep=4pt]
  \item[\textbf{A.}] \textbf{Theoretical foundations} \dotfill \pageref{app:theory}
    \begin{itemize}[leftmargin=1.5em, itemsep=2pt, topsep=2pt]
      \item[A.1] Soft evidence \dotfill \pageref{app:jeffrey_theory}
      \item[A.2] EIG derivation and question-selection ablations \dotfill \pageref{app:eig_ablations}
      \item[A.3] Question budget analysis \dotfill \pageref{app:budget}
      \item[A.4] Confidence threshold tuning \dotfill \pageref{app:threshold_tuning}
    \end{itemize}

  \item[\textbf{B.}] \textbf{System details} \dotfill \pageref{app:system}
    \begin{itemize}[leftmargin=1.5em, itemsep=2pt, topsep=2pt]
      \item[B.1] \ours diagnostic procedure \dotfill \pageref{app:bmbe_algorithm}
      \item[B.2] Prompt templates \dotfill \pageref{app:prompts}
      \item[B.3] Standalone doctor protocol \dotfill \pageref{app:standalone_protocol}
      \item[B.4] Knowledge base construction \dotfill \pageref{app:kb}
      \item[B.5] Patient simulation \dotfill \pageref{app:patient_sim}
      \item[B.6] Implementation details \dotfill \pageref{app:impl_details}
    \end{itemize}

  \item[\textbf{C.}] \textbf{Extended experimental results} \dotfill \pageref{app:extended_results}
    \begin{itemize}[leftmargin=1.5em, itemsep=2pt, topsep=2pt]
      \item[C.1] LLM-elicited knowledge base, full results \dotfill \pageref{app:llmkb_full}
      \item[C.2] Prediction normalisation \dotfill \pageref{app:disease_list_ablation}
      \item[C.3] Confidence threshold experiments \dotfill \pageref{app:threshold_experiments}
      \item[C.4] Question-selection policy ablations \dotfill \pageref{app:policy_ablations}
      \item[C.5] Prior distribution sensitivity \dotfill \pageref{app:prior_sensitivity}
      \item[C.6] Accuracy-coverage analysis \dotfill \pageref{app:acc_coverage}
      \item[C.7] Computational cost analysis \dotfill \pageref{app:cost_analysis}
      \item[C.8] Persona robustness \dotfill \pageref{app:persona_full}
      \item[C.9] Accuracy by disease prevalence \dotfill \pageref{app:prevalence_analysis}
      \item[C.10] Failure mode analysis \dotfill \pageref{app:failure_analysis}
      \item[C.11] Disease scaling analysis \dotfill \pageref{app:scaling}
      \item[C.12] Diagnostic session walkthroughs \dotfill \pageref{app:case_studies}
      \item[C.13] Bootstrap confidence intervals \dotfill \pageref{app:significance}
      \item[C.14] Baseline protocols and adaptations \dotfill \pageref{app:baselines_protocol}
    \end{itemize}

  \item[\textbf{D.}] \textbf{Extended related work} \dotfill \pageref{app:related_extended}
\end{itemize}

\newpage

%% ================================================================
%% A. THEORETICAL FOUNDATIONS
%% ================================================================

\section{Theoretical foundations}
\label{app:theory}

% [REWRITE for NeurIPS: arXiv version framed Eq.~\ref{eq:jeffrey_update} as Jeffrey's
%  conditioning, but the formula is Pearl's virtual evidence; see git history for diff.]
\subsection{Soft evidence}
\label{app:jeffrey_theory}

Standard Bayesian conditioning assumes evidence is observed with certainty. In conversational diagnosis, patient responses carry intrinsic uncertainty: ``I think I had a fever'' is not equivalent to a confirmed temperature reading. We model this with \textbf{Pearl-style virtual evidence}~\cite{pearl1988probabilistic, chan2005revision}, also called \emph{soft evidence}. Concretely, we augment the model with a virtual indicator node $V$ whose likelihood ratio encodes the parser's confidence $c \in (0, 1]$:
\begin{equation}
P(V \mid X_f{=}v) = 1, \qquad P(V \mid X_f{\neq}v) = 1 - c.
\label{eq:virtual_evidence_likelihood}
\end{equation}
For unambiguous responses ($c{=}1$), the standard Bayesian update applies:
\begin{equation}
b_{t+1}(d) \;\propto\; b_t(d)\cdot P(X_f{=}v \mid d).
\label{eq:bayes_update}
\end{equation}
For uncertain responses, the effective likelihood mixes with a neutral vector:
\begin{equation}
L_{\text{eff}}(d) \;=\; c\cdot P(X_f{=}v\mid D{=}d) \;+\; (1-c)\cdot 1, \qquad b_{t+1}(d) \;\propto\; b_t(d)\cdot L_{\text{eff}}(d).
\label{eq:jeffrey_update}
\end{equation} At $c{=}1$, $V$ deterministically asserts the observation and we recover standard Bayes. At $c{\to}0$, the likelihood ratio $P(V \mid X_f{=}v) : P(V \mid X_f{\neq}v) \to 1{:}1$, so $V$ carries no information and the posterior is unchanged. This semantics matches the parser's role: $c$ is the system's confidence in its own extraction, not a posterior probability over the underlying event.

\paragraph{Why ordinal labels?}
The five confidence labels ($\textit{very\_likely}$ through $\textit{very\_unlikely}$) provide a coarse but principled discretisation of patient certainty. This avoids requiring the LLM to output calibrated continuous probabilities. The fixed mapping to numerical weights ($c \in \{1.0, 0.80, 0.50, 0.25, 0.05\}$) is auditable and eliminates a potential source of unchecked model influence on diagnostic outcomes. Table~\ref{tab:confidence_weights} lists the complete mapping.

\begin{table}[H]
\caption{Soft-evidence confidence weights. A three-tier cascade assigns labels: (1)~deterministic pattern matching for clear or hedged language, (2)~LLM-based extraction for ambiguous responses, (3)~heuristic downgrade to \texttt{uncertain} when uncertainty cues are detected.}
\label{tab:confidence_weights}
\centering
\small
\setlength{\tabcolsep}{5pt}
\begin{tabular}{@{}lcl@{}}
\toprule
\textbf{Label} & $\boldsymbol{c}$ & \textbf{When assigned} \\
\midrule
\texttt{very\_likely}   & 1.00 & Clear, direct: ``yes'', ``no'', ``definitely'' \\
\texttt{likely}         & 0.80 & Requires interpretation or inference \\
\texttt{uncertain}      & 0.50 & ``I think so'', ``maybe'', ``not sure'' \\
\texttt{unlikely}       & 0.25 & Tangential or partially related \\
\texttt{very\_unlikely} & 0.05 & Near-zero signal; contradictory \\
\bottomrule
\end{tabular}
\end{table}
       % A.1 Jeffrey's conditioning
\subsection{EIG derivation and question-selection ablations}
\label{app:eig_ablations}
\label{app:eig_proof}

\paragraph{EIG computation details.}
The EIG policy (Eq.~\ref{eq:eig} in main text) selects the unasked feature that maximally reduces expected posterior entropy. The predictive distribution over feature values is:
\begin{equation}
P(X_f{=}v \mid \mathcal{E}_t) \;=\; \sum_{d=1}^{K} P(X_f{=}v \mid d)\; b_t(d)
\label{eq:predictive}
\end{equation}
For each candidate value $v \in \mathcal{V}_f$, a counterfactual posterior is computed via a standard Bayesian update at full confidence ($c{=}1$) without modifying internal state:
\begin{equation}
b_t^{f=v}(d) \;=\; \frac{b_t(d) \cdot P(X_f{=}v \mid d)}{\sum_{d'} b_t(d') \cdot P(X_f{=}v \mid d')}
\label{eq:counterfactual}
\end{equation}
The EIG is the reduction in expected entropy:
\begin{equation}
\operatorname{EIG}(f) \;=\; H(\mathbf{b}_t) \;-\; \sum_{v \in \mathcal{V}_f} P(X_f{=}v \mid \mathcal{E}_t) \cdot H(\mathbf{b}_t^{f=v})
\label{eq:eig}
\end{equation}
where $H(\mathbf{b}) = -\sum_d b(d) \log_2 b(d)$. At each turn the system selects the unasked feature that maximises this quantity:
\begin{equation}
f^* \;=\; \arg\max_{f \in \mathcal{F}\setminus\mathcal{A}_t}\; \operatorname{EIG}(f).
\label{eq:eig_def}
\end{equation}

\paragraph{Computational cost.}
At each turn $t$, the engine evaluates EIG for every unasked feature. Let $N_t = |\mathcal{F} \setminus \mathcal{A}_t|$ denote the remaining features. For each feature $f$, the inner loop iterates over $|\mathcal{V}_f|$ hypothetical values, each requiring a $K$-dimensional posterior update. The per-turn cost is therefore:
\begin{equation}
\mathcal{O}\!\left(N_t \cdot \max_f |\mathcal{V}_f| \cdot K\right)
\label{eq:eig_cost}
\end{equation}
For DDXPlus ($N{=}314$ features, $\max_f |\mathcal{V}_f|{=}10$, $K{=}49$ diseases), this yields ${\sim}154$K operations per turn, computed in ${\sim}50$\,ms on a single CPU core with vectorised NumPy. Since $N_t$ decreases by one each turn, the total session cost is $O(N \cdot T \cdot \max_f |\mathcal{V}_f| \cdot K)$, dominated by the first few turns.

\paragraph{Top-$k$ focused EIG.}
Once the posterior concentrates ($\max_d b_t(d) \geq \tau_{\text{focus}}$), an additional EIG term encourages fine-grained discrimination among the most probable hypotheses:
\begin{equation}
\operatorname{EIG}_{\text{combined}}(f) \;=\; \operatorname{EIG}_{\text{global}}(f) \;+\; \lambda \cdot \operatorname{EIG}_{\text{top-}k}(f)
\label{eq:eig_focused}
\end{equation}
where $\operatorname{EIG}_{\text{top-}k}$ is computed over only the top-$k$ diseases (renormalised). Ablation results (Table~\ref{tab:policy}) show that focused top-$k$ EIG with $k{=}3$, $\lambda{=}0.5$, $\tau_{\text{focus}}{=}0.3$ improves top-1 accuracy by +12\,pp over the global baseline. The main experiments use global EIG ($\lambda{=}0$) to keep the default policy parameter-free; focused EIG is available as an optional enhancement.
            % A.2 EIG derivation & ablations
\subsection{Question budget analysis}
\label{app:budget}

\paragraph{Ideal lower bound.}
Under a uniform prior over $K$ diseases, the initial entropy is $H_0 = \log_2 K$ bits. A perfectly informative binary question removes at most 1 bit, so at least $\lceil \log_2 K \rceil$ questions are required in the noiseless case.

\paragraph{Effect of soft evidence.}

In practice, evidence is soft: the effective likelihood $L_{\text{eff}}(d) = c \cdot P(X_f{=}v \mid d) + (1{-}c)$ (Eq.~\ref{eq:jeffrey_update}) corresponds to Pearl-style virtual evidence with a virtual node $V_c$ defined by $P(V_c \mid X_f{=}v) = 1$ and $P(V_c \mid X_f{\neq}v) = 1-c$ (Appendix~\ref{app:jeffrey_theory}). Since $V_c$ is generated by passing $X_f$ through a $c$-parameterised stochastic channel, $D \to X_f \to V_c$ forms a Markov chain. By the data processing inequality:
\begin{equation}
I(D;\, V_c) \;\leq\; I(D;\, X_f),
\label{eq:mi_dampened}
\end{equation}
i.e., the information about the disease accessible to the engine via the soft observation $V_c$ is upper-bounded by the information available in the clean observation $X_f$. The reduction is monotonic in $c$: at $c{=}1$, $V_c$ deterministically reveals $X_f$ and we recover the full $I(D; X_f)$; at $c{=}0$, $V_c$ is independent of $X_f$ and carries no information. The empirical average confidence $\bar{c} \approx 0.9$ on our benchmarks means the information loss per question is modest.

\paragraph{Budget and stopping-rule parameters.}
\label{app:stopping_params}
\del{We fix the hard budget to $T_{\max}{=}20$ across all benchmarks and sensor backends, inspired by the classic \emph{Twenty Questions} game~\cite{twenty_questions_acl2024}, which has also been adopted as a benchmark for multi-turn planning in LLMs. Real questions are not perfectly bisecting and evidence is soft (Eq.~\ref{eq:mi_dampened}), so the same budget is applied uniformly to the open DDXPlus 49-disease universe and to the closed-set AC-MedQA panel; per-benchmark tuning is confined to the confidence threshold $\tau$ (Appendix~\ref{app:threshold_tuning}).} The warm-up $T_{\min}$ suppresses early-stopping during the first few turns, when posteriors are dominated by the chief complaint and prior spread is largest; we use the same $T_{\min}$ for every backend so that differences in turn budget across rows of Table~\ref{tab:main} reflect the policy and not a hand-tuned schedule. Even under adversarial patient personas (Appendix~\ref{app:persona_full}), where noisy or withheld responses reduce per-turn information, the turn budget suffices for convergence.
         % A.3 Question budget analysis
\subsection{Confidence threshold tuning}
\label{app:threshold_tuning}

The confidence threshold $\tau$ controls the trade-off between coverage (fraction of patients who receive a diagnosis) and selective accuracy (quality among diagnosed patients).

\paragraph{DDXPlus.}
We sweep $\tau \in \{0.00, 0.05, \ldots, 0.95\}$ over the posterior confidence $\max_d b_T(d)$. Patients with $\max_d b_T(d) < \tau$ are abstained. $\tau^*$ is tuned by leave-one-out on a 50-case set. Results are reported in Appendix~\ref{app:threshold_experiments}.

\paragraph{LLM-generated KBs.}
The same procedure applies. Because the candidate set is smaller and the posterior tends to be less peaked, the DHS-optimal $\tau^*$ is lower than on DDXPlus. The full sweep is reported in Appendix~\ref{app:threshold_experiments}.

\paragraph{Global vs.\ per-disease calibration.}
We use a global threshold (one $\tau$ for all diseases) rather than per-disease thresholds $\tau_d$ because the limited set size ($n{=}50$) yields too few positive examples per disease for reliable per-disease calibration. The full threshold sweep is reported in Appendix~\ref{app:threshold_experiments}.
      % A.4 Confidence threshold tuning

%% ================================================================
%% B. SYSTEM DETAILS
%% ================================================================

% \newpage
\section{System details}
\label{app:system}

\subsection{\discuss{\ours} diagnostic procedure}
\label{app:bmbe_algorithm}

Algorithm~\ref{alg:bmbe} formalises the diagnostic loop described in Section~\ref{sec:method}: bulk-extracted chief-complaint evidence is folded into the prior, then each subsequent turn selects the most informative feature via expected information gain (EIG), updates the posterior under Pearl-style soft evidence, and checks the warm-up + threshold stopping rule. Symbols match those introduced in Sections~\ref{sec:kb}--\ref{sec:llm_interface}.

\begin{algorithm}[h]
\caption{\textsc{\ours} Diagnostic Session}\label{alg:bmbe}
\begin{algorithmic}[1]
\Require Knowledge base $\mathcal{K}$ with prior $\boldsymbol{\pi}$, likelihoods $\{P(X_f{=}v \mid d)\}$, schema; threshold $\tau$; warm-up $T_{\min}$; budget $T_{\max}$
\Ensure Diagnosis $d^*$ or \textsc{abstain}

\Statex
\State Initialise $b_0(d) \leftarrow \pi(d)$ for all $d \in \mathcal{D}$
\State LLM bulk-extracts $\{(f_j, v_j, c_j)\}_{j=1}^{J}$ from chief complaint \Comment{\S\ref{sec:llm_interface}}
\For{$j = 1, \ldots, J$}
    \State $\mathbf{b} \leftarrow \textsc{UpdateBelief}(\mathbf{b}, f_j, v_j, c_j)$ \Comment{Eq.~\eqref{eq:jeffrey_update}}
\EndFor

\For{$t = 1, \ldots, T_{\max}$}
    \State $f^* \leftarrow \arg\max_{f \notin \mathcal{A}_t} \operatorname{EIG}(f;\, \mathbf{b}_t)$ \Comment{Eq.~\eqref{eq:eig}}
    \State LLM verbalises $f^*$ as natural-language question
    \State $(v_t,\, c_t) \leftarrow \textsc{Parse}(\text{patient response}, f^*, \mathcal{S}_{f^*})$
    \If{$v_t = \texttt{unknown}$ or $v_t = \texttt{clarification}$}
        \State Re-ask $f^*$ with clarification; parse again $\rightarrow (v_t, c_t)$
    \EndIf
    \State $\mathbf{b}_{t+1} \leftarrow \textsc{UpdateBelief}(\mathbf{b}_t, f^*, v_t, c_t)$
    \If{$t \geq T_{\min}$ \textbf{and} $\max_d b_{t+1}(d) \geq \tau$}
        \State \textbf{break} \Comment{sufficient evidence gathered}
    \EndIf
\EndFor

\Statex
\State \textbf{Decision:}
\State $d^* \leftarrow \arg\max_d\, b_T(d)$
\If{$b_T(d^*) \geq \tau$} \Return $d^*$
\Else{} \Return \textsc{abstain}
\EndIf

\end{algorithmic}
\end{algorithm}

\begin{figure}[h]
\centering
\includegraphics[width=\textwidth]{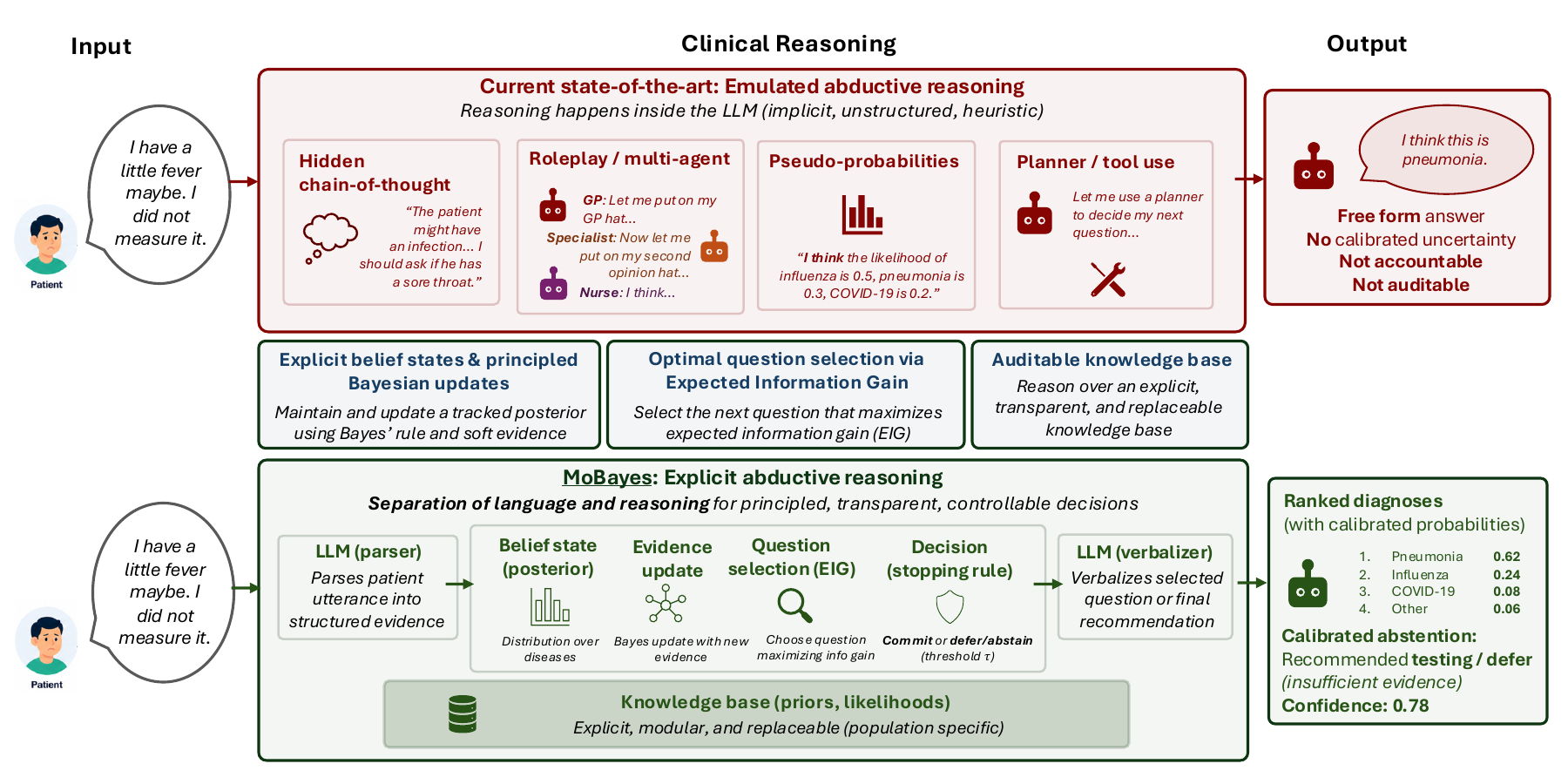}
\caption{Paradigm comparison. \textbf{Top:} current state-of-the-art emulates abductive reasoning inside the LLM (hidden chain-of-thought, role-play, pseudo-probabilities, planner / tool use); the output is a free-form answer with no calibrated uncertainty, accountability, or audit trail. \textbf{Bottom:} \discuss{\ours} separates language and reasoning. The LLM acts only as a parser and verbaliser; an explicit Bayesian module tracks a posterior, applies soft-evidence updates, selects the next question by expected information gain, and decides whether to commit or abstain at threshold $\tau$. Reasoning is conducted over a transparent, modular, replaceable knowledge base, and the output is a ranked differential with calibrated probabilities and a defensible abstain option.}
\label{fig:paradigm_comparison_full}
\end{figure}

\paragraph{Paradigm comparison.}
Figure~\ref{fig:paradigm_comparison_full} contrasts the two paradigms at a glance: existing state-of-the-art systems emulate abductive reasoning inside the LLM (hidden chain-of-thought, multi-agent role-play, pseudo-probability prompts, planner / tool use), producing free-form answers without calibrated uncertainty, accountability, or audit trail. \discuss{\ours} separates language and reasoning: the LLM only parses patient utterances and verbalises selected questions, while a Bayesian module maintains an explicit belief state, updates it with soft evidence, selects the next question by expected information gain, and applies a calibrated stopping rule, all over a transparent and replaceable knowledge base.

  % B.0 \ours diagnostic procedure (moved from Method)
\subsection{Prompt templates}
\label{app:prompts}

All prompts use temperature $0.0$ for deterministic output. The LLM never receives posterior values, entropy, or any internal engine state.

\paragraph{Parsing prompt.}
The parsing prompt receives the patient utterance, the feature under inquiry, and the allowed schema values. A deterministic cascade (pattern matching for binary yes/no, emphatic affirmations, numeric ranges, and direct schema substring matches) is applied \emph{before} the LLM call; the LLM is invoked only as a fallback for ambiguous utterances. The prompt enforces 13 confidence-assignment rules (e.g., prefer \texttt{unknown} over hard negatives when the response is partial or vague) to ensure conservative, faithful extraction.

\begin{promptbox}[Parsing Prompt]
System: You are a medical data extractor.\\
User Response: "\{utterance\}"\\
Feature to Extract: "\{feature\}" (Allowed: \{schema\})\\[4pt]
TASK:\\
1. Classify the user response into one of the Allowed Values.\\
2. Assess confidence: very\_likely, likely, uncertain, unlikely, very\_unlikely\\[4pt]
KEY RULES:\\
-- Direct answer ("Yes", "Not really"): map to closest value.\\
-- Unrelated response: return "unknown|likely".\\
-- Uncertain language ("I think so", "maybe"): use "uncertain".\\
-- Prefer "unknown" over hard negative when partial/vague.\\[4pt]
Return format: "value|confidence\_level"\\
Examples: "high|likely", "yes|uncertain", "no|very\_likely"
\end{promptbox}

\paragraph{Verbaliser prompt.}
The verbaliser receives the feature name, answer schema, and a coarse system-confidence indicator derived from $\max_d b_t(d)$. It modulates conversational tone only; no internal state is exposed.

\begin{promptbox}[Verbaliser Prompt]
System: You are a professional medical diagnostic assistant.\\
Task: Ask the patient a question to check for: \{feature\}.\\[4pt]
CONTEXT:\\
-- Feature Name: \{feature\}\\
-- Expected Details: \{schema\}\\
-- System Confidence: \{low|medium|high\}\\[4pt]
CONSTRAINTS:\\
1. NEVER use technical IDs (e.g., "f\_fever", "d\_flu").\\
2. Speak naturally and empathetically.\\
3. Do NOT mention probabilities or internal values.\\
4. If clarifying a previous confusion, keep it brief.
\end{promptbox}

\paragraph{Bulk intake prompt.}
At session start, a single bulk intake call maps the patient's opening narrative to multiple $(f, v, c)$ triples simultaneously, reducing the number of follow-up questions needed.

\begin{promptbox}[Bulk Intake Prompt]
System: You are an expert medical intake specialist.\\
User Text: "\{narrative\}"\\[4pt]
TASK: Extract any clinical features mentioned in the text.\\[4pt]
ALLOWED FEATURES:\\
-- \{feature\_id\}: \{name\} (Values: \{values\})\\[4pt]
RULES:\\
1. Only extract explicitly mentioned or strongly implied features.\\
2. Extract demographics (age, gender, location) if present.\\
3. Do NOT infer negatives from silence; omit unlisted features.\\
4. Assess confidence for each extracted feature.\\[4pt]
Return JSON: \{"feature\_id": \{"value": "...", "confidence": "likely"\}, "demographics": \{"age": N, ...\}\}
\end{promptbox}

\paragraph{Patient simulator prompt.}
The patient simulator receives the full clinical profile (demographics, chief complaint, symptoms, medical history, observed findings) and persona instructions. Crucially, the patient must answer faithfully according to its observed findings, including clear denials when the finding is negative, but must say ``I'm not sure'' for features not listed in its profile, avoiding fabrication.

\begin{promptbox}[Patient Simulator Prompt]
System: You are a patient in an Emergency Department consultation.\\[4pt]
PATIENT PROFILE:\\
-- Age, Gender, Chief Complaint, Symptoms Present\\
-- Social History, Family History, Medical History, Pain Level\\
-- KNOWN OBSERVED FINDINGS (answer faithfully; do not assume unlisted findings)\\[4pt]
RULES:\\
1. Answer based on KNOWN OBSERVED FINDINGS and patient profile.\\
2. If asked about something listed: answer faithfully (including denials).\\
3. If NOT listed: say ``I'm not sure'' or ``I don't know''. Do not invent symptoms.\\
4. NEVER reveal your diagnosis directly.\\
5. Keep responses concise (1--3 sentences).\\[4pt]
PERSONA:\\
-- Language: \{CEFR level A/B/C\}\\
-- Personality: \{plain $|$ verbose $|$ overanxious $|$ distrustful\}\\
-- Memory: \{high $|$ low recall\}\\
-- Alertness: \{normal $|$ moderate daze $|$ high daze\}
\end{promptbox}

\paragraph{Standalone doctor prompt.}
The standalone LLM doctor receives no external reasoning support. It conducts the full diagnostic interview and outputs a differential (\texttt{[DDX]}) or abstains (\texttt{[ABSTAIN]}).

\begin{promptbox}[Standalone Doctor Prompt ,  Single-Disease (DDXPlus)]
System: You are an emergency department physician conducting a diagnostic interview.\\
Ask focused, one-at-a-time questions to narrow down the diagnosis.\\
After gathering enough information:\\
-- Start with [DDX] followed by your top 5 diagnoses (most to least likely), separated by commas.\\
-- If not confident, start with [ABSTAIN] + short reason, then [POSSIBLE] + your top 3 possible diagnoses.\\[4pt]
Guidelines:\\
-- Ask one question at a time.\\
-- Focus on discriminating features.\\
-- Consider common and serious diagnoses.\\
-- Do not guess when evidence is insufficient.\\
-- Maximum of \{max\_turns\} questions before [DDX]/[ABSTAIN].
\end{promptbox}

\paragraph{Disease matching prompt.}
Free-text doctor outputs from every non-AC baseline are aligned to the active closed disease list \del{($49$ entries on the DDxPlus columns, $31$ on AgentClinic-MedQA) }by a unified LLM-judge pass with the gpt-5.4-nano backbone shared across all baselines. A single call returns the closed-list Top-1, the closed-list Top-3, and an explicit \emph{abstention} flag (true only on a literal refusal-to-commit). Off-list commits are recorded as committed-but-wrong, not as abstentions; this separation is what lets selective accuracy and coverage decouple in Table~\ref{tab:combined_n100}.

\begin{promptbox}[Disease Matching Prompt]
You are a medical-coding assistant. Given the doctor's diagnostic output below, do three things:\\[4pt]
1. \textbf{TOP-1}: pick the disease from the closed list of \{n\} that semantically matches the doctor's TOP candidate (the rank-1 commit). If none match, output \texttt{"Unknown"}.\\[2pt]
2. \textbf{TOP-3}: pick up to three diseases from the closed list that semantically match the doctor's RANKED differential (in order). If a candidate cannot be mapped, output \texttt{"Unknown"} for that slot.\\[2pt]
3. \textbf{ABSTAIN}: set \texttt{"abstained": true} only if the doctor's TOP candidate text is literally \texttt{ABSTAIN}, \texttt{Unknown}, \texttt{I don't know}, empty, or otherwise an explicit refusal. Set \texttt{false} whenever the doctor names any disease, even if that disease is not in the closed list (off-list = committed-but-wrong, not abstain).\\[4pt]
Closed disease list:\\
\{disease\_list\}\\[4pt]
Doctor's diagnostic output:\\
---\\
\{diag\_text\}\\
---\\[4pt]
Semantic-match guidance:\\
-- ``Chronic Obstructive Pulmonary Disease (COPD)'' matches ``Chronic obstructive pulmonary disease''.\\
-- ``CKD'' matches ``Chronic kidney failure''.\\
-- ``Acute acalculous cholecystitis with sepsis'' matches ``Acalculous cholecystitis''.\\
-- Markdown asterisks, parenthetical abbreviations, and minor wording variants are all OK.\\
-- Use \texttt{"Unknown"} only if the candidate clearly does not refer to any listed disease.\\[4pt]
Output JSON exactly in this format and nothing else:\\
\texttt{\{"top1": "<disease|Unknown>", "top3": ["<d1>", "<d2>", "<d3>"], "abstained": true|false\}}
\end{promptbox}

\paragraph{LLM-generated KB prompts.}
The LLM-generated KB is constructed via two sequential prompts. The \emph{feature generation prompt} asks the model to propose clinically plausible features for a given disease (with deduplication against an existing catalog). The \emph{distribution estimation prompt} then elicits $P(\text{feature} \mid \text{disease})$ for all feature--disease pairs: binary features as a single \texttt{prob\_yes}, and continuous features as a discrete distribution over a $0$--$5$ severity scale (where $0$ encodes absence). Features are processed in chunks; outputs are validated (distributions must sum to 1) before being added to the KB.

\begin{promptbox}[Feature Generation Prompt]
You are helping build a synthetic feature catalog for a conversational diagnostic system.\\
Disease: \{disease\_id\}, \{name\}\\
Include only features with P(present | disease) > \{threshold\}.\\
Allowed kinds: symptom\_presence, symptom\_attribute, medical\_history, lifestyle, risk\_factor.\\
Before proposing, check the existing catalog and reuse semantically equivalent features.\\
Return: action=reuse|new, estimated\_probability, justification.
\end{promptbox}

\begin{promptbox}[Distribution Estimation Prompt]
You are generating likelihood parameters for a Bayesian diagnostic model.\\
Disease: \{disease\_id\}, \{name\}\\
For binary features: return prob\_yes.\\
For numerical features (0--5 scale, 0=absent): return probability for each value; must sum to 1.\\
Use medically plausible values; avoid extreme certainty.
\end{promptbox}

\paragraph{AgentClinic-MedQA KB construction prompts.}
The AgentClinic-MedQA KB (Section~\ref{app:ac_medqa_curation}) is built in a multi-stage pipeline rather than a single elicitation pass. Disease-driven and patient-driven feature elicitation prompts follow the same template as the LLM-generated KB feature prompt above (with the closed disease list and patient narratives respectively as context). The three remaining prompts encode the constraints specific to this column: a deduplication pass that preserves both generic and specific feature variants, and the banded continuous likelihood elicitation.

\begin{promptbox}[Preserve-Specificity Dedup Prompt]
You are a clinical knowledge engineer building a diagnostic feature schema. You will receive candidate phrases from two source catalogs:\\
\hspace*{1em}-- ``specific'': disease-driven detailed phrasings (e.g., ``rough hyperkeratotic plaque on sun-exposed area'').\\
\hspace*{1em}-- ``generic'': patient-driven short phrasings (e.g., ``skin lesions'').\\
Many are near-duplicates. Produce a canonical list that PRESERVES BOTH LAYERS.\\
Rules:\\
\hspace*{1em}(1) DROP entirely: negative phrasings, demographic-only items, lab / imaging results, metadata-only items.\\
\hspace*{1em}(2) KEEP both general and specific variants of the same observation; DO NOT merge across granularity.\\
\hspace*{1em}(3) MERGE only true duplicates (same concept, same specificity level).\\
\hspace*{1em}(4) Use clinical phrasing, lowercase, no patient slang.\\
Output a JSON array of canonical feature names, ONLY the array.
\end{promptbox}

\begin{promptbox}[Banded Likelihood Elicitation Prompt]
You are a clinical knowledge engineer assigning $P(\text{YES} \mid \text{disease})$ likelihoods. Disease: \{disease\}.\\
For each relevant feature, assign a probability on a CONTINUOUS scale, using the following bands as a guide (vary values WITHIN bands; do NOT cluster on band midpoints):\\
\hspace*{1em}-- 0.02 default (background; unrelated organ system).\\
\hspace*{1em}-- 0.05--0.10 slight relevance (rare incidental association).\\
\hspace*{1em}-- 0.12--0.25 possible / uncommon presentation.\\
\hspace*{1em}-- 0.40--0.60 common presentation.\\
\hspace*{1em}-- 0.65--0.85 strong / typical / classic presentation.\\
\hspace*{1em}-- 0.90--0.97 pathognomonic / near-defining.\\
For features irrelevant to this disease, OMIT them; they will fall through to the 0.02 default.\\
Output JSON: \texttt{\{"disease": "...", "p\_yes": \{"feature\_name": value, ...\}\}}.
\end{promptbox}
        % B.1 Prompt templates
\subsection{Standalone doctor protocol}
\label{app:standalone_protocol}

Algorithm~\ref{alg:standalone} summarises the standalone doctor inference loop.

\begin{algorithm}[h]
\caption{Standalone LLM Doctor}\label{alg:standalone}
\hrule\vspace{2pt}
\begin{algorithmic}[1]
\Require Patient profile $\mathbf{p}$, LLM doctor $\mathcal{M}$, turn budget $T_{\max}$
\Ensure Diagnosis $\hat{y}$ or \textsc{Abstain}
\State $\mathcal{H} \gets [\,]$ \Comment{dialogue history}
\State $\mathcal{H}.\text{append}(\text{SystemPrompt}(\mathcal{M}, T_{\max}))$
\State $r_{\text{patient}} \gets \text{ChiefComplaint}(\mathbf{p})$
\For{$t = 1, \ldots, T_{\max}$}
    \State $\mathcal{H}.\text{append}(r_{\text{patient}})$
    \State $r_{\text{doctor}} \gets \mathcal{M}(\mathcal{H})$
    \State $\mathcal{H}.\text{append}(r_{\text{doctor}})$
    \If{$r_{\text{doctor}}$ contains \texttt{[DDX]}}
        \State Parse comma-separated list $\to \hat{\mathbf{y}}_{\text{raw}} = [\hat{y}_1, \ldots, \hat{y}_k]$
        \State $\hat{\mathbf{y}} \gets \text{MatchToKB}(\hat{\mathbf{y}}_{\text{raw}}, \mathcal{D}_{\text{KB}})$ \Comment{LLM-based name matching}
        \State \Return $\hat{\mathbf{y}}$
    \ElsIf{$r_{\text{doctor}}$ contains \texttt{[ABSTAIN]}}
        \State \Return \textsc{Abstain} (+ optional \texttt{[POSSIBLE]} list)
    \EndIf
    \State $r_{\text{patient}} \gets \text{PatientSim}(\mathbf{p}, \mathcal{H})$
\EndFor
\State Inject ``provide your final diagnosis now'' into $\mathcal{H}$
\State $r_{\text{doctor}} \gets \mathcal{M}(\mathcal{H})$
\If{$r_{\text{doctor}}$ contains \texttt{[DDX]}}
    \State Parse and $\hat{\mathbf{y}} \gets \text{MatchToKB}(\hat{\mathbf{y}}_{\text{raw}}, \mathcal{D}_{\text{KB}})$
    \State \Return $\hat{\mathbf{y}}$
\Else{} \Return \textsc{Abstain}
\EndIf
\end{algorithmic}
\hrule
\end{algorithm}

The standalone doctor outputs a comma-separated differential (\texttt{[DDX]}) or explicitly abstains (\texttt{[ABSTAIN]}). The prompt requests the top~5 diagnoses, ranked by likelihood. Crucially, the doctor generates free-text disease names \emph{without access to the KB disease list}, mirroring realistic deployment where a physician names conditions from memory. We additionally evaluate a closed-world variant (CW) where the doctor receives the KB disease list at inference time in Section~\ref{sec:results}; the CW baseline directly addresses whether KB-list awareness changes standalone performance.

\paragraph{Post-hoc disease name normalisation.}
Free-text predictions (e.g., ``Acute exacerbation of COPD'') may differ lexically from KB entries (e.g., ``Acute COPD exacerbation / infection''), so the $\text{MatchToKB}$ step of Algorithm~\ref{alg:standalone} aligns each prediction to the canonical disease list using the unified LLM-judge described in Appendix~\ref{app:prompts}. Predictions that the judge cannot match to any closed-list entry are retained as off-list commits and contribute neither to Top-$k$ accuracy nor to abstention, ensuring that naming a disease outside the KB scope incurs a penalty. We manually audited all 250 standalone-doctor commits across the six tested models on DDxPlus50 (sessions where the doctor produced a final \texttt{[DDX]}; abstention sessions do not invoke the matcher). Between 4 and 13 (1.6\% to 5.2\%, 95\% Clopper-Pearson CI [0.4\%, 8.7\%]) involved a matcher error that mapped a clinically distinct entity onto the ground-truth KB entry, generating a false-positive top-1 hit. The range reflects the strictness of clinical equivalence (e.g., ``acute decompensated heart failure'' vs ``acute pulmonary edema''). No false negatives were observed. Matcher errors thus inflate standalone scores; correcting them would strengthen rather than weaken \ours's reported advantage.

The matcher (one GPT-5.4-nano call per session) adds approximately 0.018 cents per standalone session: under 0.1\% of the cost of the most expensive standalone doctors and at most 2\% of the cheapest. It is omitted from Table~\ref{tab:cost} without affecting any reported value at the displayed precision.
      % B.2 Standalone doctor protocol
\subsection{Knowledge base construction}
\label{app:kb}

Both knowledge bases are clinically grounded, each through a different mechanism: DDXPlus derives likelihoods from physician-designed patient records, and the LLM-generated KB from structured elicitation of a frontier model's medical knowledge.
Table~\ref{tab:dataset_summary} provides a summary comparison.

Both KBs share the same Dirichlet--Categorical likelihood formulation. For each disease $d$ and feature $f$ with values $\mathcal{V}_f$, we store empirical counts $n_{f,v,d}$ (co-occurrence in training records, or normalised pseudo-counts derived from LLM elicitations) and define the conditional as a Categorical with additive Laplace smoothing:
\begin{equation}
P(X_f{=}v \mid D{=}d) \;=\; \frac{\alpha_{f,v,d}}{\sum_{v' \in \mathcal{V}_f} \alpha_{f,v',d}}, \qquad \alpha_{f,v,d} = n_{f,v,d} + 1.
\label{eq:dirichlet_mean}
\end{equation}
The Laplace prior ensures $P(X_f{=}v\mid d)>0$ for every triple, preventing zero-likelihood pathologies under sequential inference; counts are normalised to sum to $100$ per (disease, feature) pair before smoothing.

\begin{table}[t]
\centering
\caption{Summary statistics of the knowledge bases used in evaluation.}
\label{tab:dataset_summary}
\small
\begin{tabular}{@{}lccc@{}}
\toprule
                             & \textbf{DDXPlus} & \textbf{LLM-KB (GPT)} & \textbf{LLM-KB (Gemini)} \\
\midrule
Source                       & Empirical        & GPT-5.4               & Gemini 3.1               \\
Diseases                     & 49               & 18                    & 18                       \\
Features                     & 314              & 160                   & 89                       \\
\quad Binary                 & 306              & 154                   & 85                       \\
\quad Categorical / ordinal  & 8                & 6                     & 4                        \\
KB source records            & ${\sim}$1.03M    & --                    & --                       \\
Evaluation patients          & 50               & 50                    & 50                       \\
\bottomrule
\end{tabular}
\end{table}

\paragraph{DDXPlus-derived KB.}
DDXPlus~\cite{ddxplus2022} contains ${\sim}1.03$M training records, each annotated with a ground-truth pathology and a set of observed evidences. We map the 49 pathologies to canonical disease IDs with ICD-10 codes and organise them into 9 chapter-level categories. The 223 original evidences comprise three types: binary, categorical (V-code resolved to readable labels), and multi-choice. Multi-choice evidences are expanded into individual binary sub-features (top-20 values by frequency), yielding 314 features in total.

Co-occurrence counts are accumulated per (disease, feature, value) triple across all training patients. Binary evidences absent from a patient's record are treated as negative, except for a small set of negated features where this default would invert the clinical meaning. Priors are set proportional to empirical disease frequency in the training split.

\begin{figure}[htbp]
\centering
\includegraphics[width=0.48\textwidth]{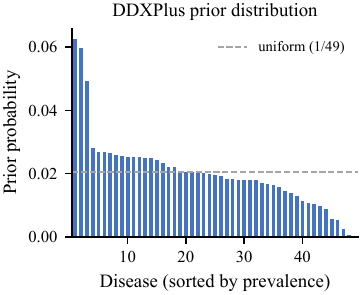}%
\hfill
\includegraphics[width=0.48\textwidth]{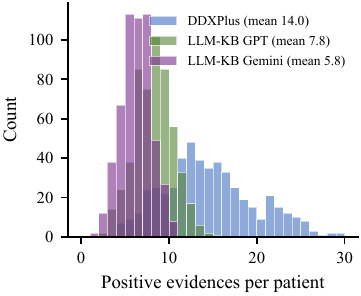}
\caption{\textbf{Left:} DDXPlus prior distribution sorted by prevalence. The long tail ($\max/\min \approx 200\times$) reflects real-world disease frequency; the dashed line shows the uniform baseline. \textbf{Right:} Distribution of positive evidence count per evaluation patient across KBs.}
\label{fig:prior_and_density}
\end{figure}

\paragraph{LLM-generated KB.}
\label{app:kb_generation}
To control for informational advantages (Section~\ref{sec:datasets}), we construct LLM-generated KBs whose likelihoods are elicited entirely from frontier language models. We generate two variants: one using GPT-5.4 and one using Gemini~3.1, enabling a cross-model comparison of zero-shot medical knowledge.

Construction proceeds in two stages (prompts in Appendix~\ref{app:prompts}). In the \emph{feature proposal} stage, the model generates clinically relevant features for each disease. Proposals reference a growing shared catalog to encourage feature reuse across diseases; a deduplication step removes near-duplicate entries. Features with estimated relevance below a threshold $\theta$ are discarded. In the \emph{distribution estimation} stage, the model provides $P(\text{feature} \mid \text{disease})$ for all retained features: a single probability for binary features, and a discrete distribution over a severity scale for ordinal features. All outputs undergo automated validation (probability constraints, identifier consistency) before inclusion; invalid entries are excluded. The resulting pseudo-counts are normalised identically to the DDXPlus KB.

\paragraph{AgentClinic-MedQA KB construction.}
\label{app:ac_medqa_curation}
AgentClinic-MedQA ships no labelled training split, so the frequency-based likelihood estimation used for DDXPlus is unavailable, and a single-pass elicitation from a fixed verbal scale induces a mode-collapse artefact in which most likelihoods cluster on a few discrete values, leaving the engine unable to discriminate symptomatically similar diseases. We therefore assemble the KB in two stages, keeping the benchmark's closed disease universe and its native moderator unchanged and setting priors to uniform.

The feature schema is the union of two complementary symptom catalogs. A \emph{patient-driven} pass extracts findings from the validation-set patient narratives shipped with the benchmark (e.g.\ \emph{skin lesions}), grounding the schema in the language patients actually use. A \emph{disease-driven} pass elicits clinically discriminative findings from the authoring language model one disease at a time (e.g.\ \emph{rough hyperkeratotic plaque on sun-exposed skin}), supplying the specific terminology a clinician would seek. An LLM deduplication pass then merges the two catalogs under a preserve-specificity constraint, so coarse and fine-grained variants of the same observation are retained side by side. The conditional likelihoods $P(\text{yes}\mid d)$ are subsequently authored by a language model under a continuous banded prompt that spreads probability mass across the full $[0,1]$ interval, avoiding the bin-clustering artefact of the simpler verbal-scale elicitation. The benchmark's doctor, patient, measurement, and moderator agents are not modified; per-case adaptations are protocol-level only (Section~\ref{app:baselines_protocol}).

\paragraph{Synthetic patient generation.}
\label{app:synthetic_patients}
Since no real patient data exists for LLM-generated KBs, evaluation patients are generated through a parallel per-disease elicitation: for each disease $d$, the LLM is queried for the conditional probability $P(X_f \mid D{=}d)$ of each relevant clinical feature, and a feature vector is then drawn by an independent coin flip per feature,
\begin{equation}
x_f^{(i)} \sim P(X_f \mid D{=}d) \quad \text{for each feature } f \in \mathcal{F}_d,
\label{eq:patient_sampling}
\end{equation}
a Bernoulli with parameter $P(X_f{=}\texttt{yes} \mid d)$ for binary features and a categorical draw over the $0$--$5$ severity scale for ordinal ones, where $\mathcal{F}_d \subseteq \mathcal{F}$ is the set of features relevant to disease $d$ (those with non-uniform likelihood). The resulting patients are independent samples from the LLM's own per-disease symptom distributions, not handcrafted vignettes. The chief complaint is constructed from the first three positive symptoms; the full feature map is stored as the patient's observed findings. Demographics (age, sex) are sampled uniformly at random. This yields $K \times n_{\text{per}}$ patients (e.g., $18 \times 30 = 540$), from which a stratified subset of $n{=}50$ is selected for benchmarking (at least one patient per disease). The sampling procedure ensures a ``fair fight'': both \discuss{\ours} and standalone doctors face patients whose symptom profiles are consistent with the LLM's own clinical beliefs, neither system has an informational advantage.

\paragraph{Cross-KB transfer analysis.}
\label{app:cross_kb}
To test KB generalisation, we evaluate each ecosystem's \discuss{\ours} on patients generated from the \emph{other} KB. Let $\mathcal{F}^{\text{GPT}}$ and $\mathcal{F}^{\text{Gem}}$ denote the feature sets of the two KBs. Features are matched by canonical name: $\mathcal{F}^{\cap} = \{f : \text{name}(f) \in \text{names}(\mathcal{F}^{\text{GPT}}) \cap \text{names}(\mathcal{F}^{\text{Gem}})\}$. Of 160 GPT and 89 Gemini features, $|\mathcal{F}^{\cap}| = 45$ share identical names and types (28\% of GPT, 51\% of Gemini). When a cross-KB patient presents a feature $f \notin \mathcal{F}_{\text{KB}}$, the engine receives no matching schema entry and skips the update (equivalent to $L_{\text{eff}} = \mathbf{1}$). On average, GPT patients retain 43\% of their features when evaluated against the Gemini KB, and Gemini patients retain 59\% against the GPT KB. Results are reported in Table~\ref{tab:cross_kb}.

\begin{table}[h]
\caption{Cross-KB transfer: each \discuss{\ours} variant evaluated on patients from the other ecosystem. ``Native'' rows repeat the in-ecosystem results for reference. Feature coverage indicates the fraction of patient features recognised by the cross-KB.}
\label{tab:cross_kb}
\centering
\small
\setlength{\tabcolsep}{4pt}
\begin{tabular}{@{}llcccccc@{}}
\toprule
\textbf{System} & \textbf{Patients} & \textbf{Feat.\,Cov.} & \textbf{Top-1} & \textbf{Top-3} & \textbf{Sel.\,Acc} & \textbf{Cov.} & \textbf{DHS} \\
\midrule
\ours + GPT-nano, KB: GPT     & GPT (native)   & 100\% & 70 & 84 & \cellcolor{f1mid}78 & 90 & \cellcolor{f1high}83 \\
\ours + GPT-nano, KB: GPT     & Gemini (cross) & 43\%  & 52 & 78 & \cellcolor{f1low}54 & 96 & \cellcolor{f1low!50}69 \\
\midrule
\ours + Gemini FL, KB: Gemini  & Gemini (native) & 100\% & 64 & 84 & \cellcolor{f1mid!50}66 & 94 & \cellcolor{f1mid}78 \\
\ours + Gemini FL, KB: Gemini  & GPT (cross)     & 59\%  & 42 & 66 & \cellcolor{f1low}48 & 84 & \cellcolor{f1low}61 \\
\bottomrule
\end{tabular}
\end{table}

\begin{figure}[htbp]
\centering
\includegraphics[width=0.48\textwidth]{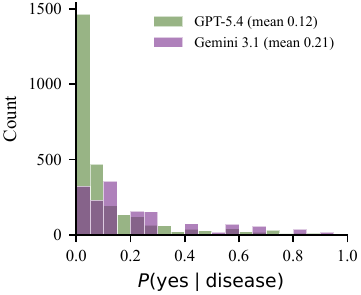}%
\hfill
\includegraphics[width=0.48\textwidth]{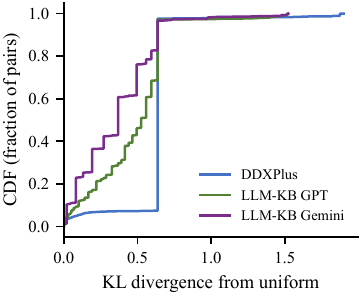}
\caption{\textbf{Left:} Distribution of LLM-elicited binary likelihoods $P(\text{yes}\mid d)$ for both GPT and Gemini KBs; the strong left skew indicates that most disease--feature associations are weak. \textbf{Right:} CDF of per-pair KL divergence from uniform across all three KBs; DDXPlus (empirical) has the highest informativeness, while both LLM-KBs are comparable despite being synthetically generated. \wip{The informativeness gap between the empirical and LLM-elicited KBs is likely prompt-addressable: contrastive or pathognomonic-emphasis elicitation prompts that explicitly ask the model to discriminate symptomatically adjacent diseases should pull the LLM-KB CDF closer to the empirical curve.}}
\label{fig:llmkb_stats}
\end{figure}

\begin{figure}[htbp]
    \centering
    \includegraphics[width=0.48\textwidth]{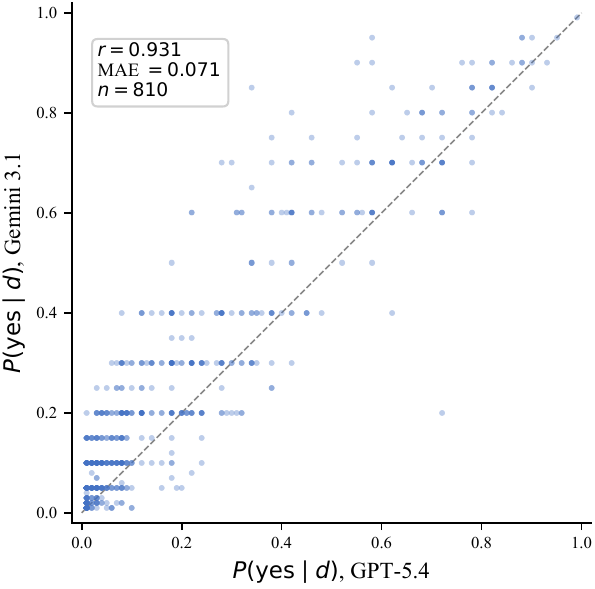}%
    \hfill
    \includegraphics[width=0.48\textwidth]{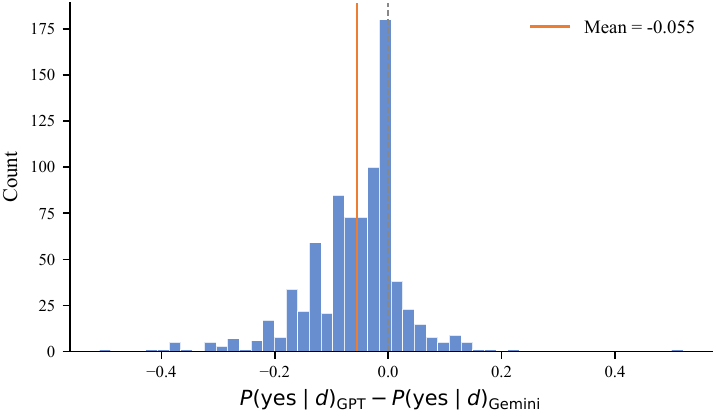}
    \caption{\textbf{Left:} Scatter plot of $P(\text{yes} \mid d)$ for 45 shared features across 18 diseases ($n{=}810$ pairs); dashed line is perfect agreement. \textbf{Right:} Distribution of pairwise likelihood differences; the left-skewed distribution (mean $= -0.055$) confirms Gemini's systematically higher assignments.}
    \label{fig:llmkb_scatter}
    \label{fig:llmkb_diff}
\end{figure}

\begin{figure}[htbp]
    \centering
    \includegraphics[width=0.95\textwidth]{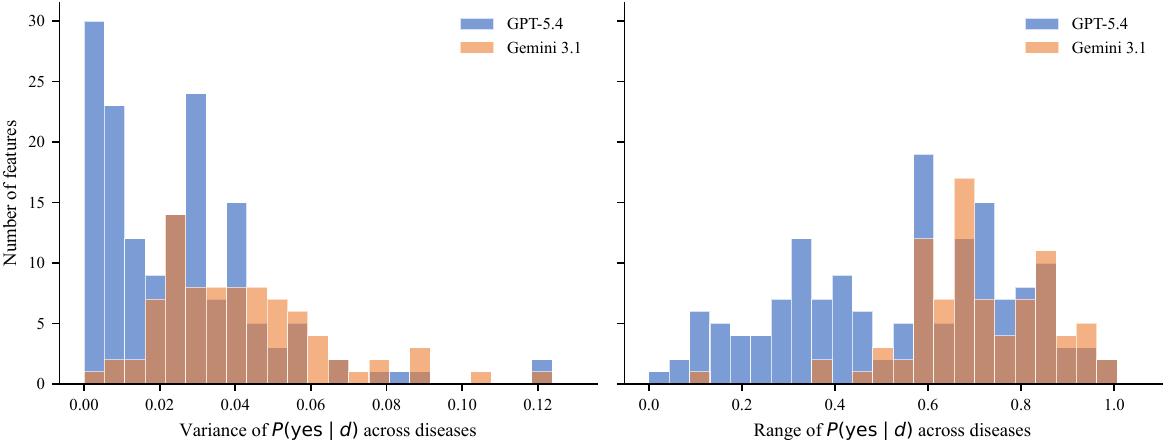}
    \caption{Feature discriminativeness (left: cross-disease variance; right: cross-disease range) for GPT-5.4 (154 features) and Gemini~3.1 (85 features). Gemini's features are more discriminative on average, reflecting both targeted feature selection and more extreme likelihood assignments.}
    \label{fig:llmkb_disc}
\end{figure}

\paragraph{Inter-model agreement on LLM-generated KBs.}
\label{app:llmkb_agreement}
We compare the two LLM-generated KBs (GPT-5.4 and Gemini~3.1, same 18 diseases, same generation pipeline) to quantify inter-model agreement on medical knowledge. GPT generated 160 binary features; Gemini produced 89. Only 45 share identical names (Jaccard $J{=}0.22$), yet downstream agreement on likelihoods is high: on the 45 shared features across 18 diseases (810 pairs; Figure~\ref{fig:llmkb_scatter}), Pearson $r{=}0.931$ with MAE${=}0.071$. A systematic bias exists: Gemini assigns likelihoods 0.055 higher on average (Figure~\ref{fig:llmkb_diff}). Gemini's features are $1.7{\times}$ more discriminative (cross-disease variance 0.041 vs 0.024; Figure~\ref{fig:llmkb_disc}), consistent with a smaller but more targeted feature set. Despite no shared weights, the two KBs converge on what matters, suggesting that frontier LLMs encode broadly consistent medical knowledge when prompted with structured elicitation.

 % B.3 Knowledge base construction
\subsection{Patient simulation}
\label{app:patient_sim}

We simulate patients using an LLM-based conversational agent that receives a ground-truth clinical profile (demographics, chief complaint, symptom findings) and responds to doctor questions in character. The simulator is inspired by the persona taxonomy of PatientSim~\cite{patientsim2025}, which parameterises patient behaviour along four axes: \emph{personality archetype} (cooperative, anxious, distrustful, etc.), \emph{language proficiency}, \emph{recall accuracy} (high/low), and \emph{cognitive confusion} (none/moderate/high).

We select five personas by varying these axes (full definitions in Table~\ref{tab:personas}): \textit{Plain} (cooperative baseline), \textit{Overanxious} (high anxiety, over-reports), \textit{Distrustful} (withholds information), \textit{Dazed} (low recall, moderate confusion), and \textit{Verbose} (tangential, low recall).

The simulator LLM (GPT-5.4-nano) is shared across all experiments to ensure consistent patient behaviour. It sees only the clinical profile and persona instructions; it never accesses the diagnostic engine's internal state, posterior, or knowledge base. The simulator is instructed to answer faithfully for features listed in its profile and to express uncertainty for unlisted features, though we observe that LLMs occasionally infer plausible but ungrounded responses from profile context (see Appendix~\ref{app:failure_analysis} for the impact on diagnostic accuracy).

\paragraph{Patient-simulator adaptations across the three columns.}
\label{app:patient_sim_adaptations}
Each external simulator is integrated through a thin compatibility shim that leaves its diagnostic behaviour, system prompt, and per-turn protocol unchanged; the changes are interface-level only.

\textbf{(1) MEDDxAgent HT-sim.} MEDDxAgent's history-taking simulator is plugged in as released; the patient system prompt, including the upstream instruction to default to ``I don't know'' for any feature outside the profile, is unchanged.

\textbf{(2) AgentClinic native patient agent.} AC's PatientAgent is used as released, with its system prompt and turn-by-turn conversation accumulator preserved verbatim. The lenient parsing prompt of Appendix~\ref{app:prompts} is swapped in on the \emph{engine} parser to handle hedged AC patient utterances; the patient simulator itself is untouched. Patient- and doctor-side token counts are tracked separately on this column.

\textbf{(3) PatientSim on DDxPlus.} We adopt the four-axis persona taxonomy of PatientSim~\cite{patientsim2025} (personality, language proficiency, recall accuracy, cognitive confusion) along with the upstream persona category names. The reference implementation is built around a MIMIC-derived profile schema; we re-wire the same persona taxonomy to drive the DDxPlus profile schema used in our other columns, leaving the persona axes themselves unchanged. Main experiments use the cooperative \emph{Plain} persona; the four adversarial personas (\emph{Overanxious}, \emph{Distrustful}, \emph{Dazed}, \emph{Verbose}) are evaluated in the persona ablation of Appendix~\ref{app:persona_full}.

Per-method doctor-side shims (\ours engine patches, baseline reproductions, closed-set normalisation) are documented separately in Appendix~\ref{app:baselines_protocol}.
      % B.4 Patient simulation
\subsection{Implementation details}
\label{app:impl_details}
\label{app:details}

%% ============================================================
%% TABLE: Model specifications
%% ============================================================

\begin{table}[h]
\caption{Models used in experiments. All models accessed via API (OpenAI direct or OpenRouter). The patient simulator (GPT-5.4-nano) is shared across all experiments.}
\label{tab:models}
\centering
\small
\begin{tabular}{@{}ll@{}}
\toprule
\textbf{\ours sensors} & \textbf{Standalone doctors} \\
\midrule
GPT-5.4-nano          & GPT-5.4          \\
Gemini 3.1 Flash Lite & Gemini 3.1 Pro   \\
Llama-4-Scout         & Llama-4-Maverick \\
GPT-OSS-20B           & GPT-OSS-120B     \\
Gemma 4 31B           & Qwen 3.6 Plus    \\
MiniMax M2.5          & Kimi K2.5        \\
\bottomrule
\end{tabular}
\end{table}

All models are accessed via API. OpenAI models (GPT-5.4, GPT-5.4-nano) are called directly; all other models are routed through OpenRouter. \discuss{\ours} components use temperature~$0$ for deterministic parsing and question generation.\del{ The dialogue budget is capped at $T_{\max} = 20$ turns for both \discuss{\ours} and standalone configurations.}
  % B.5 Implementation details

%% ================================================================
%% C. EXTENDED EXPERIMENTAL RESULTS
%% ================================================================

% \newpage
\section{Extended experimental results}

Unless stated otherwise, the ablations of Appendices use the DDxPlus 50-case set with the PatientSim wrapper.

\subsection{LLM-elicited knowledge base, full results}
\label{app:llmkb_full}

This appendix expands on the LLM-generated knowledge base experiment reported in Section~\ref{sec:results_llmkb}. In each ecosystem (GPT-5.4 and Gemini~3.1), the same frontier LLM populates the closed-disease KB and (separately) generates evaluation patients (Eq.~\ref{eq:patient_sampling}). Both are noisy realisations of the LLM's beliefs; the remaining gap reflects how the LLM's own knowledge is deployed.

\paragraph{Headline finding.}
Across both ecosystems, an inexpensive \discuss{\ours} sensor matches or approaches the frontier standalone doctor on top-1 accuracy at an order-of-magnitude lower per-token cost. A striking result is that the standalone doctor still drops sharply on \emph{rare} diseases even though the patients are simulated from its own world model, while the Bayesian engine leverages the KB uniformly across prevalence. This is direct evidence that the gap is not informational: the LLM \emph{has} the relevant probabilistic content (it generated the KB), but cannot deploy it consistently when forced to operate from natural language alone.

\paragraph{KB oracle row.}
For each ecosystem we report a \emph{KB oracle} row that uses the generating LLM's own posterior over the KB as the diagnostic system. This is the upper bound implied by the KB itself; \discuss{\ours} approaches it without the standalone doctor's inflation from training-data leakage. Both \ours and the standalone are evaluated at $n{=}50$ on the Plain persona.

%% ============================================================
%% TABLE: Combined LLM-KB Ecosystems (GPT + Gemini)
%% Layout: top tabular = system rows x 5 metric cols (matches Table 2.a);
%%         bottom tabular = 2 prevalence-stratified DHS rows transposed
%%         (systems as columns, common / rare as rows).
%% ============================================================

\begin{table}[h]
\centering
\caption{LLM-generated KB experiments, full results. Each ecosystem uses the generating model's own clinical knowledge; patients sampled from the KB's likelihood model. $n{=}50$, Plain persona. Bottom block reports DHS stratified by LLM-estimated ED prevalence; the standalone doctor's drop on \emph{rare} diseases (despite the patients being simulated from its own world model) is the architectural finding highlighted in Section~\ref{sec:results_llmkb}.}
\label{tab:llmkb_full}
\setlength{\tabcolsep}{3pt}

\begin{tabular}{@{}l cc c ccc@{}}
\toprule
& \multicolumn{2}{c}{\textbf{Accuracy}} & & \multicolumn{3}{c}{\textbf{Selective Diagnosis}} \\
\cmidrule(lr){2-3} \cmidrule(lr){5-7}
\textbf{System} & \textbf{Top-1} & \textbf{Top-3} & & \textbf{Sel.\,Acc} & \textbf{Cov.} & \textbf{DHS} \\
\midrule
\multicolumn{7}{l}{\textit{GPT-5.4 ecosystem}} \\[2pt]
\rowcolor{gray!8}
\quad \textit{KB oracle ($c{=}1$)} & \color{gray}\textit{76} & \color{gray}\textit{88} & & \color{gray}--- & \color{gray}--- & \color{gray}--- \\[2pt]
\quad \ours + GPT-5.4-nano     & 70 & 84 & & 78 & 90 & \cellcolor{dhsgreen!30}83 \\
\cdashline{1-7}[0.5pt/2pt]
\quad SA GPT-5.4              & 72 & 80 & & 81 & 86 & \cellcolor{dhsgreen!30}84 \\
\quad SA GPT-5.4-nano         & 52 & 70 & & 72 & 36 & \cellcolor{dhsred!30}48 \\
\midrule
\multicolumn{7}{l}{\textit{Gemini 3.1 ecosystem}} \\[2pt]
\rowcolor{gray!8}
\quad \textit{KB oracle ($c{=}1$)} & \color{gray}\textit{74} & \color{gray}\textit{88} & & \color{gray}--- & \color{gray}--- & \color{gray}--- \\[2pt]
\quad \ours + Gemini FL         & 64 & 84 & & 66 & 94 & \cellcolor{dhsgreen!25}78 \\
\cdashline{1-7}[0.5pt/2pt]
\quad SA Gemini 3.1 Pro        & 70 & 78 & & 73 & 96 & \cellcolor{dhsgreen!30}83 \\
\quad SA Gemini 3.1 Flash Lite & 62 & 76 & & 63 & 92 & \cellcolor{dhsred!5}75 \\
\bottomrule
\end{tabular}

\vspace{4pt}

\setlength{\tabcolsep}{4pt}
\begin{tabular}{@{}l ccc ccc@{}}
\toprule
& \multicolumn{3}{c}{\textbf{GPT-5.4 ecosystem}} & \multicolumn{3}{c}{\textbf{Gemini 3.1 ecosystem}} \\
\cmidrule(lr){2-4} \cmidrule(lr){5-7}
\textbf{Prevalence}
  & \textbf{\ours} & \textbf{SA Full} & \textbf{SA nano}
  & \textbf{\ours} & \textbf{SA Full} & \textbf{SA light} \\
\midrule
Common
  & \cellcolor{dhsgreen!30}83
  & \cellcolor{dhsgreen!40}\textbf{95}
  & \cellcolor{dhsred!10}70
  & \cellcolor{dhsred!10}69
  & \cellcolor{dhsgreen!35}\textbf{88}
  & \cellcolor{dhsgreen!30}82 \\
Rare
  & \cellcolor{dhsgreen!35}\textbf{84}
  & \cellcolor{dhsgreen!15}75
  & \cellcolor{dhsred!35}29
  & \cellcolor{dhsgreen!30}\textbf{82}
  & \cellcolor{dhsgreen!25}79
  & \cellcolor{dhsred!10}71 \\
\bottomrule
\end{tabular}
\end{table}

               % C.0 LLM-elicited KB full results (Table 2.b moved here)
\subsection{Prediction normalisation}
\label{app:disease_list_ablation}

All standalone doctor predictions undergo post-hoc normalisation to KB disease names via the $\text{MatchToKB}$ procedure (Section~\ref{app:standalone_protocol}). The doctor never sees the KB disease list during the dialogue; normalisation is applied only after the doctor commits to a diagnosis. This ensures a fair comparison: the standalone doctor relies solely on its internal medical knowledge, while \discuss{\ours} relies on the structured KB. \wip{The complementary direction, surfacing the KB disease list to the standalone doctor inside its system prompt, is reported separately as the Closed-World (CW) baseline in Table~\ref{tab:combined_n100} (full protocol in Appendix~\ref{app:baselines_protocol}); mere KB-list awareness does not recover the architectural gap.}
\label{app:extended_results}
      % C.1 Prediction normalisation
\subsection{Confidence threshold experiments}
\label{app:threshold_experiments}

Per-sensor $\tau^*$ is tuned by leave-one-out DHS-maximisation on the set used for that run; the chosen value is then applied to score the same set. Table~\ref{tab:threshold} reports the DDXPlus confidence threshold sweep across all six \discuss{\ours} sensor backends.

%% ============================================================
%% TABLE: Confidence threshold sweep (DDXPlus, all \ours sensors)
%% ============================================================

\begin{table}[h]
\caption{Confidence threshold $\tau$ sweep on DDXPlus ($n{=}50$, Plain persona) for all \discuss{\ours} sensor backends. Sel.\,Acc.\ = selective Top-1 accuracy among committed cases; Cov.\ = fraction committed; DHS = harmonic mean of Sel.\,Acc.\ and Cov.\ (Eq.~\ref{eq:metrics}). Bold: DHS-optimal $\tau^*$ per sensor, used in Table~\ref{tab:main}.}
\label{tab:threshold}
\centering
\scriptsize
\setlength{\tabcolsep}{2pt}
\begin{tabular}{c ccc ccc ccc ccc ccc ccc}
\toprule
& \multicolumn{3}{c}{\textbf{GPT-5.4-n}}
& \multicolumn{3}{c}{\textbf{Gemini FL}}
& \multicolumn{3}{c}{\textbf{Llama-4-Sc}}
& \multicolumn{3}{c}{\textbf{GPT-OSS-20B}}
& \multicolumn{3}{c}{\textbf{Gemma 4}}
& \multicolumn{3}{c}{\textbf{MiniMax}} \\
\cmidrule(lr){2-4} \cmidrule(lr){5-7} \cmidrule(lr){8-10} \cmidrule(lr){11-13} \cmidrule(lr){14-16} \cmidrule(lr){17-19}
$\boldsymbol{\tau}$
  & \textbf{SA} & \textbf{C} & \textbf{DHS}
  & \textbf{SA} & \textbf{C} & \textbf{DHS}
  & \textbf{SA} & \textbf{C} & \textbf{DHS}
  & \textbf{SA} & \textbf{C} & \textbf{DHS}
  & \textbf{SA} & \textbf{C} & \textbf{DHS}
  & \textbf{SA} & \textbf{C} & \textbf{DHS} \\
\midrule
0.00 & 78 & 100 & 88 & 74 & 100 & 85 & 56 & 100 & 72 & 61 & 100 & 76 & 64 & 100 & 78 & 60 & 100 & 75 \\
0.10 & 78 & 100 & 88 & 74 & 100 & 85 & 56 & 100 & 72 & 61 & 100 & 76 & 64 & 100 & 78 & 60 & 100 & 75 \\
\textbf{0.20} & \textbf{81} & \textbf{96} & \cellcolor{f1high}\textbf{88} & \textbf{77} & \textbf{96} & \cellcolor{f1high}\textbf{86} & 57 & 98 & 72 & 61 & 100 & 76 & 64 & 100 & 78 & 60 & 100 & 75 \\
\textbf{0.30} & 81 & 94 & 87 & 77 & 94 & 84 & 58 & 96 & 73 & \textbf{65} & \textbf{94} & \cellcolor{f1high}\textbf{77} & \textbf{69} & \textbf{93} & \cellcolor{f1high}\textbf{79} & \textbf{64} & \textbf{94} & \cellcolor{f1high}\textbf{76} \\
\textbf{0.40} & 83 & 92 & 87 & 78 & 92 & 85 & \textbf{61} & \textbf{92} & \cellcolor{f1high}\textbf{73} & 71 & 84 & 77 & 71 & 86 & 78 & 69 & 78 & 73 \\
0.50 & 82 & 90 & 86 & 79 & 86 & 82 & 62 & 84 & 71 & 70 & 76 & 73 & 77 & 79 & 78 & 71 & 76 & 73 \\
0.60 & 81 & 86 & 84 & 81 & 74 & 77 & 64 & 78 & 70 & 74 & 71 & 73 & 81 & 75 & 78 & 77 & 68 & 72 \\
0.70 & 80 & 80 & 80 & 88 & 66 & 75 & 73 & 66 & 69 & 79 & 67 & 73 & 80 & 71 & 76 & 79 & 66 & 72 \\
0.80 & 86 & 72 & 78 & 88 & 64 & 74 & 74 & 62 & 68 & 87 & 61 & 72 & 84 & 68 & 75 & 83 & 58 & 68 \\
0.90 & 85 & 68 & 76 & 93 & 60 & 73 & 85 & 54 & 66 & 90 & 59 & 71 & 83 & 64 & 73 & 89 & 52 & 66 \\
0.95 & 88 & 64 & 74 & 97 & 58 & 73 & 85 & 52 & 65 & 90 & 59 & 71 & 83 & 64 & 73 & 88 & 50 & 64 \\
\bottomrule
\end{tabular}
\end{table}

\paragraph{LLM-generated KB threshold sweep.}
Table~\ref{tab:llmkb_threshold} reports the confidence threshold sweep for both LLM-generated KB ecosystems, including native and cross-KB transfer configurations. The optimal $\tau$ is selected by maximising DHS.

\begin{table}[h]
\caption{Confidence threshold sweep for LLM-generated KB experiments ($n{=}50$, Plain persona). Bold: DHS-optimal operating point. Cross-KB rows evaluate on patients generated from the other ecosystem's KB.}
\label{tab:llmkb_threshold}
\centering
\small
\setlength{\tabcolsep}{3pt}
\scriptsize
\setlength{\tabcolsep}{2pt}
\begin{tabular}{@{}l ccc ccc ccc ccc@{}}
\toprule
& \multicolumn{3}{c}{\textbf{GPT KB (native)}}
& \multicolumn{3}{c}{\textbf{GPT KB $\to$ Gem.\ pat.}}
& \multicolumn{3}{c}{\textbf{Gem.\ KB (native)}}
& \multicolumn{3}{c}{\textbf{Gem.\ KB $\to$ GPT pat.}} \\
\cmidrule(lr){2-4} \cmidrule(lr){5-7} \cmidrule(lr){8-10} \cmidrule(lr){11-13}
$\tau$ & SA & Cov & DHS & SA & Cov & DHS & SA & Cov & DHS & SA & Cov & DHS \\
\midrule
0.00 & 70 & 100 & 82 & 52 & 100 & 68 & 64 & 100 & 78 & 42 & 100 & 59 \\
0.10 & 70 & 100 & 82 & 52 & 100 & 68 & 64 & 100 & 78 & 42 & 100 & 59 \\
\textbf{0.20} & 73 & 96 & 83 & \textbf{54} & \textbf{96} & \cellcolor{f1high}\textbf{69} & 64 & 100 & \cellcolor{f1high}\textbf{78} & 42 & 100 & 59 \\
\textbf{0.30} & \textbf{78} & \textbf{90} & \cellcolor{f1high}\textbf{83} & 61 & 76 & 67 & 66 & 94 & \cellcolor{f1high}78 & \textbf{48} & \textbf{84} & \cellcolor{f1high}\textbf{61} \\
0.40 & 83 & 72 & 77 & 70 & 60 & 65 & 79 & 76 & 77 & 57 & 56 & 57 \\
0.50 & 89 & 52 & 66 & 68 & 56 & 61 & 84 & 62 & 71 & 67 & 42 & 52 \\
0.60 & 96 & 44 & 60 & 68 & 50 & 58 & 88 & 50 & 64 & 73 & 30 & 43 \\
0.70 & 100 & 40 & 57 & 68 & 44 & 54 & 90 & 40 & 55 & 75 & 24 & 36 \\
0.80 & 100 & 38 & 55 & 77 & 34 & 47 & 88 & 32 & 47 & 80 & 20 & 32 \\
0.90 & 100 & 32 & 49 & 80 & 30 & 44 & 100 & 16 & 28 & 78 & 18 & 29 \\
\bottomrule
\end{tabular}
\end{table}
     % C.3 Confidence threshold experiments
\subsection{Question-selection policy ablations}
\label{app:policy_ablations}

Table~\ref{tab:policy} compares question-selection policies on DDXPlus (policy descriptions in Appendix~\ref{app:eig_ablations}).

%% ============================================================
%% TABLE: Question-selection policy ablation (DDXPlus)
%% ============================================================

\begin{table}[h]
\caption{Question-selection policy ablation on DDXPlus (\ours + GPT-5.4-nano, Plain persona). \emph{Global EIG} selects the most informative question across all diseases; \emph{Focused top-$k$ EIG} adds a discrimination term over the $k$ most probable diseases, weighted by $\lambda$, activated once $\max_d b_t(d) \geq \theta$. DHS uses DHS-optimal $\tau^*$. The main experiments (Table~\ref{tab:main}) use Global EIG; focused EIG can improve DHS by up to +7\,pp. Colour intensity: \textcolor{dhsgreen}{green}~=~higher, \textcolor{dhsred}{red}~=~lower.}
\label{tab:policy}
\centering
\small
\setlength{\tabcolsep}{4pt}
\begin{tabular}{@{}ll cc c ccc@{}}
\toprule
& & \multicolumn{2}{c}{\textbf{Accuracy}} & & \multicolumn{3}{c}{\textbf{Selective Diagnosis}} \\
\cmidrule(lr){3-4} \cmidrule(lr){6-8}
\textbf{Policy} & \textbf{Params}
  & \textbf{Top-1} & \textbf{Top-3}
  & & \textbf{Sel.\,Acc} & \textbf{Cov.} & \textbf{DHS} \\
\midrule
Global EIG (baseline)       & $\lambda{=}0$                   & 76 & 88 & & 83 & 92 & \cellcolor{dhsgreen!15}87 \\
\midrule
Focused top-3 EIG           & $\lambda{=}0.3,\;\theta{=}0.3$  & 80 & 88 & & 87 & 92 & \cellcolor{dhsgreen!20}89 \\
Focused top-3 EIG           & $\lambda{=}0.5,\;\theta{=}0.3$  & \textbf{88} & \textbf{92} & & \textbf{96} & 92 & \cellcolor{dhsgreen!40}\textbf{94} \\
Focused top-3 EIG           & $\lambda{=}0.3,\;\theta{=}0.5$  & 84 & 88 & & 88 & 96 & \cellcolor{dhsgreen!30}92 \\
Focused top-3 EIG           & $\lambda{=}1.0,\;\theta{=}0.3$  & 72 & 80 & & 86 & 84 & \cellcolor{dhsred!10}85 \\
Focused top-5 EIG           & $\lambda{=}0.3,\;\theta{=}0.3$  & 84 & \textbf{92} & & 88 & 96 & \cellcolor{dhsgreen!30}92 \\
\bottomrule
\end{tabular}
\end{table}

Focused top-$k$ EIG improves accuracy over the global baseline by concentrating discriminative questions on the most probable diseases once the posterior begins to concentrate. The optimal configuration uses moderate focus strength applied early; too-strong focus degrades performance, likely because the model commits prematurely and stops exploring alternative hypotheses. Wider focus ($k{=}5$) is slightly worse than $k{=}3$, suggesting that a tighter candidate set produces more targeted questions. The main experiments (Table~\ref{tab:main}) use global EIG ($\lambda{=}0$) to keep the default policy simple and parameter-free; focused EIG is available as an optional enhancement.
    % C.4 Question-selection policy ablations
\subsection{Prior distribution sensitivity}
\label{app:prior_sensitivity}

We compare three prior initialisation strategies $b_0(d)$ on DDXPlus ($n{=}50$, \ours + GPT-5.4-nano):

\begin{itemize}[leftmargin=1.5em, itemsep=2pt]
    \item \textbf{Empirical.} $b_0(d) = \hat{p}(d)$, the maximum-likelihood prevalence estimated from the DDXPlus training split.
    \item \textbf{Uniform.} $b_0(d) = 1/K$, assigning equal mass to all $K$ diseases.
    \item \textbf{Conditional.} $b_0(d \mid a, s) \propto \hat{p}(d, a, s)$, prevalence stratified by age group $a$ and sex $s$ (six demographic bins), estimated from training counts.
\end{itemize}

\noindent Table~\ref{tab:prior_ablation} reports results at each prior's DHS-optimal threshold.

%% ============================================================
%% TABLE: Prior distribution ablation (DDXPlus)
%% ============================================================

\begin{table}[h]
\caption{Prior distribution ablation on DDXPlus ($n{=}50$, \ours + GPT-5.4-nano). DHS-optimal $\tau^*$ selected per prior. Colour intensity: \textcolor{dhsgreen}{green}~=~higher, \textcolor{dhsred}{red}~=~lower.}
\label{tab:prior_ablation}
\centering
\small
\setlength{\tabcolsep}{4pt}
\begin{tabular}{@{}l cc c ccc@{}}
\toprule
& \multicolumn{2}{c}{\textbf{Accuracy}} & & \multicolumn{3}{c}{\textbf{Selective Diagnosis}} \\
\cmidrule(lr){2-3} \cmidrule(lr){5-7}
\textbf{Prior}
  & \textbf{Top-1} & \textbf{Top-3}
  & & \textbf{Sel.\,Acc} & \textbf{Cov.} & \textbf{DHS} \\
\midrule
Empirical    & 70 & 88 & & 71 & 98 & \cellcolor{dhsgreen!15}83 \\
Uniform      & 70 & 92 & & 71 & 98 & \cellcolor{dhsgreen!15}83 \\
Conditional  & \textbf{74} & 90 & & \textbf{79} & 94 & \cellcolor{dhsgreen!30}\textbf{86} \\
\bottomrule
\end{tabular}
\end{table}

Empirical and uniform priors achieve identical DHS despite assigning very different initial mass to diseases. This insensitivity is expected given the minimum-question threshold $T_{\min}{=}12$: the engine collects enough evidence to wash out the prior before making a decision. The conditional prior achieves the highest DHS by starting closer to the true posterior for each demographic group, reducing the evidence burden.

As the disease space scales and symptom overlap increases, however, prior choice is likely to matter more: with many competing hypotheses sharing similar features, a well-calibrated prior can break ties that evidence alone cannot resolve within the question budget.

\paragraph{Per-prevalence breakdown.}
Table~\ref{tab:prior_prevalence} stratifies the empirical vs.\ uniform comparison by prevalence tercile (Appendix~\ref{app:prevalence_analysis}).

%% ============================================================
%% TABLE: Prior ablation per prevalence tercile (DDXPlus)
%% ============================================================

\begin{table}[h]
\caption{Effect of prior choice on Top-1 accuracy by disease prevalence ($n{=}50$, \ours + GPT-5.4-nano). Diseases ranked by training-set prevalence and split into terciles. $\Delta$: uniform $-$ empirical.}
\label{tab:prior_prevalence}
\centering
\small
\setlength{\tabcolsep}{5pt}
\begin{tabular}{@{}lc cc c@{}}
\toprule
\textbf{Prevalence group} & $\boldsymbol{n}$
  & \textbf{Empirical} & \textbf{Uniform} & $\boldsymbol{\Delta}$ \\
\midrule
Common (top 16)     & 20 & 70 & 70 & \cellcolor{dhsgreen!0}\phantom{+0}$\pm$0 \\
Medium (mid 17)     & 17 & \textbf{76} & 65 & \cellcolor{dhsred!15}$-$11 \\
Rare (bottom 16)    & 13 & 62 & \textbf{77} & \cellcolor{dhsgreen!20}$+$15 \\
\midrule
All                 & 50 & 70 & 70 & \phantom{+0}$\pm$0 \\
\bottomrule
\end{tabular}
\end{table}

The uniform prior improves accuracy on rare diseases at the cost of medium-prevalence diseases. This trade-off arises because the empirical prior assigns disproportionately high mass to common diseases, which helps conditions sharing features with frequent diagnoses but hinders rare diseases that must overcome a strong prior disadvantage. Under the uniform prior, the posterior is driven entirely by evidence, benefiting rare diseases whose distinctive symptom profiles are otherwise masked by prior dominance.
   % C.6 Prior distribution sensitivity
\subsection{Accuracy-coverage analysis}
\label{app:acc_coverage}

\begin{figure}[h]
\centering
\includegraphics[width=0.85\textwidth]{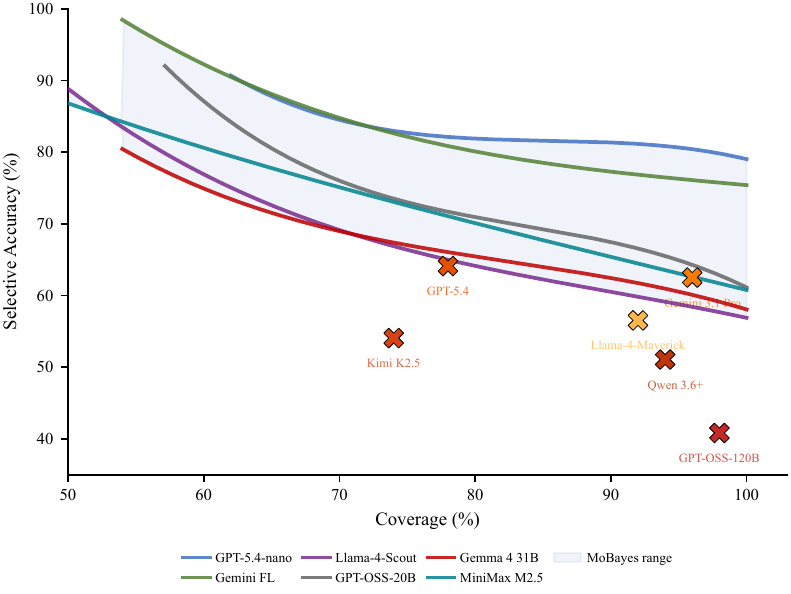}
\caption{DDXPlus selective accuracy vs.\ coverage ($n{=}50$). Each session runs to completion; the curve is traced by sweeping the commit threshold $\tau$ on the final posterior (smoothed via spline interpolation). Standalone doctors (\texttimes) operate at a single, abstention-determined point. \ours dominates at every matched coverage level.}
\label{fig:acc_cov_ddxplus}
\end{figure}

A selective classifier's quality is best assessed not at a single threshold but across the full accuracy--coverage frontier. Figure~\ref{fig:acc_cov_ddxplus} plots selective accuracy as a function of coverage for all six \ours sensors on DDXPlus, with standalone doctors shown as fixed points.

At matched coverage, every \ours sensor dominates every standalone doctor: the \ours envelope lies entirely above the cluster of standalone points. This gap widens as coverage decreases, because \ours's explicit posterior allows it to selectively abstain on genuinely uncertain cases, whereas standalone doctors lack a reliable confidence signal for threshold-based abstention. The result is that \ours offers strictly better accuracy at every operating point a deployer might choose.
        % C.7 Accuracy-coverage analysis
\subsection{Computational cost analysis}
\label{app:cost_analysis}

Table~\ref{tab:cost} reports per-model API pricing and estimated per-patient cost on DDXPlus. The per-token view of the cost--accuracy tradeoff is reported in Figure~\ref{fig:cost_combined} (main text); the per-patient view, which folds in dialogue length and per-turn token routing, is shown in Figure~\ref{fig:cost_per_patient} below.

All API calls in our experiments are routed through CacheSaver~\cite{potamitis2025cachesaver}, a namespace-aware list-valued inference cache that deduplicates repeated queries while preserving i.i.d.\ statistical integrity within a namespace, reducing the marginal compute and carbon cost of repeated evaluations and stabilising reproducibility across runs.

%% ============================================================
%% TABLE: Computational cost comparison (DDXPlus)
%% ============================================================

\begin{table}[h]
\caption{API pricing and per-patient cost on DDXPlus. Listed prices as of April~2026 from each model's OpenRouter page (OpenAI models via direct API). Actual cost may vary by provider and routing. Cost/pat is the mean estimated spend per diagnostic session. Colour intensity: \textcolor{dhsgreen}{green}~=~better, \textcolor{dhsred}{red}~=~worse.}
\label{tab:cost}
\centering
\small
\setlength{\tabcolsep}{3pt}
\begin{tabular}{@{}l cc c cc c c@{}}
\toprule
& \multicolumn{2}{c}{\textbf{Pricing (\$/M)}} & & \multicolumn{2}{c}{\textbf{Performance}} & & \\
\cmidrule(lr){2-3} \cmidrule(lr){5-6}
\textbf{System} & \textbf{In} & \textbf{Out} & & \textbf{Sel.\,Acc} & \textbf{DHS} & & \textbf{Cost/pat} \\
\midrule
\multicolumn{8}{l}{\textit{\ours + sensor LLM (ours)}} \\[2pt]
\quad + GPT-5.4-nano       & 0.20  & 1.25  & & \textbf{81} & \cellcolor{dhsgreen!40}\textbf{88} & & \cellcolor{dhsgreen!25}1.4 \\
\quad + Gemini 3.1 FL      & 0.25  & 1.50  & & 77 & \cellcolor{dhsgreen!35}86 & & \cellcolor{dhsgreen!25}1.6 \\
\quad + GPT-OSS-20B        & 0.03  & 0.14  & & 70 & \cellcolor{dhsgreen!20}78 & & \cellcolor{dhsgreen!35}0.9 \\
\quad + MiniMax M2.5       & 0.12  & 0.99  & & 64 & \cellcolor{dhsgreen!15}76 & & \cellcolor{dhsgreen!20}2.0 \\
\quad + Gemma 4 31B        & 0.13  & 0.38  & & 60 & \cellcolor{dhsgreen!10}74 & & \cellcolor{dhsgreen!35}0.9 \\
\quad + Llama-4-Scout      & 0.08  & 0.30  & & 61 & \cellcolor{dhsgreen!10}73 & & \cellcolor{dhsgreen!30}1.0 \\
\midrule
\multicolumn{8}{l}{\textit{Standalone LLM doctors}} \\[2pt]
\quad SA GPT-5.4           & 2.50  & 15.00 & & 64 & \cellcolor{dhsred!10}70 & & \cellcolor{dhsred!35}19.6 \\
\quad SA Gemini 3.1 Pro    & 2.00  & 12.00 & & 63 & \cellcolor{dhsgreen!15}76 & & \cellcolor{dhsred!35}25.0 \\
\quad SA Qwen 3.6+         & 0.33  & 1.95  & & 51 & \cellcolor{dhsred!20}66 & & \cellcolor{dhsred!10}4.2 \\
\quad SA Kimi K2.5         & 0.38  & 1.72  & & 54 & \cellcolor{dhsred!25}63 & & \cellcolor{dhsred!5}2.7 \\
\quad SA GPT-OSS-120B      & 0.04  & 0.19  & & 41 & \cellcolor{dhsred!30}58 & & \cellcolor{dhsgreen!30}1.0 \\
\quad SA Llama-4-Maverick  & 0.15  & 0.60  & & 57 & \cellcolor{dhsred!10}70 & & \cellcolor{dhsgreen!35}0.9 \\
\bottomrule
\end{tabular}

\vspace{2pt}
{\scriptsize Cost/pat in US cents per diagnostic session.}
\end{table}

\wipblock{\begin{figure}[h]
\centering
\includegraphics[width=0.55\textwidth]{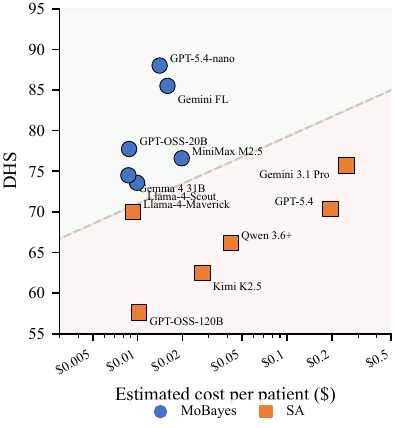}
\caption{DHS vs.\ estimated per-patient API cost on DDXPlus ($n{=}50$). \ours sensors (circles) achieve higher DHS than frontier standalone doctors (squares) at $10$--$25\times$ lower per-patient cost. Cost is the mean over $50$ sessions of doctor-side and patient-simulator token spend, summing input and output tokens at each model's listed rate (Table~\ref{tab:cost}); patient simulator pricing is fixed at GPT-5.4-nano (\$0.20/M~in, \$1.25/M~out).}
\label{fig:cost_per_patient}
\end{figure}}
                % C.8 Computational cost analysis
\subsection{Persona robustness}
\label{app:persona_full}
\label{app:persona_robustness}

A clinically useful system must handle diverse patient communication styles. We evaluate robustness across five patient personas that vary personality, language proficiency, recall accuracy, and cognitive state (Table~\ref{tab:personas}).

%% ============================================================
%% TABLE: Patient Personas
%% ============================================================

\begin{table}[h]
\caption{Patient personas used for robustness evaluation. Each persona varies the simulated patient along four axes from PatientSim~\cite{patientsim2025}.}
\label{tab:personas}
\centering
\small
\setlength{\tabcolsep}{3.5pt}
\begin{tabular}{@{}l cccc p{4.2cm}@{}}
\toprule
\textbf{Persona} & \textbf{Proficiency} & \textbf{Recall} & \textbf{Confusion} & \textbf{Personality} & \textbf{Effect on dialogue} \\
\midrule
Plain       & B & High & None     & Cooperative  & Baseline. Answers directly. \\[2pt]
Overanxious & B & High & None     & Anxious      & Over-reports symptoms, adds false positives. \\[2pt]
Distrustful & B & High & None     & Skeptical    & Withholds information, terse answers. \\[2pt]
Dazed       & A & Low  & Moderate & Confused     & Inconsistent responses across turns. \\[2pt]
Verbose     & B & Low  & None     & Talkative    & Buries relevant info in tangential detail. \\
\bottomrule
\end{tabular}
\end{table}

We test \ours and standalone GPT-5.4-nano across four adversarial personas in addition to the cooperative baseline. Table~\ref{tab:persona_full} reports the complete breakdown.

%% ============================================================
%% TABLE: Full Per-Persona Breakdown (DDXPlus, GPT-5.4-nano)
%% ============================================================

\begin{table}[h]
\caption{Per-persona results on DDXPlus (GPT-5.4-nano as sensor/doctor). \ours uses DHS-optimal $\tau^*$; SA uses self-determined abstention. $\Delta$: \ours $-$ SA. Colour intensity: \textcolor{dhsgreen}{green}~=~higher, \textcolor{dhsred}{red}~=~lower.}
\label{tab:persona_full}
\centering
\small
\setlength{\tabcolsep}{3pt}
\begin{tabular}{@{}l ccc c ccc c c@{}}
\toprule
& \multicolumn{3}{c}{\textbf{\ours}} & & \multicolumn{3}{c}{\textbf{Standalone}} & & \\
\cmidrule(lr){2-4} \cmidrule(lr){6-8}
\textbf{Persona}
  & \textbf{Sel.\,Acc} & \textbf{Cov.} & \textbf{DHS}
  & & \textbf{Sel.\,Acc} & \textbf{Cov.} & \textbf{DHS}
  & & $\boldsymbol{\Delta}$\textbf{DHS} \\
\midrule
Plain             & 81 & 96 & \cellcolor{dhsgreen!40}88  & & 64 & 78 & \cellcolor{dhsred!10}70  & & \cellcolor{dhsgreen!20}+18 \\
Overanxious       & 79 & 96 & \cellcolor{dhsgreen!35}87  & & 80 & 40 & \cellcolor{dhsred!30}53  & & \cellcolor{dhsgreen!35}+34 \\
Distrustful       & 88 & 96 & \cellcolor{dhsgreen!45}92  & & 89 & 36 & \cellcolor{dhsred!30}51  & & \cellcolor{dhsgreen!40}+41 \\
Dazed             & 71 & 96 & \cellcolor{dhsgreen!25}82  & & 50 & 48 & \cellcolor{dhsred!30}49  & & \cellcolor{dhsgreen!35}+33 \\
Verbose           & 72 & 100 & \cellcolor{dhsgreen!30}84  & & 56 & 64 & \cellcolor{dhsred!20}60  & & \cellcolor{dhsgreen!25}+24 \\
\midrule
\textit{Mean (adv.)} & \textit{78} & \textit{97} & \cellcolor{dhsgreen!30}\textit{86}  & & \textit{69} & \textit{47} & \cellcolor{dhsred!30}\textit{53}  & & \cellcolor{dhsgreen!35}\textit{+33} \\
\bottomrule
\end{tabular}
\end{table}

\ours maintains high coverage and stable DHS across all adversarial personas, degrading only moderately from the plain baseline. In contrast, the standalone doctor's coverage collapses under adversarial conditions: when patients are evasive or confused, the model abstains on the majority of cases, producing low DHS despite sometimes achieving high selective accuracy on the few cases it does commit to. The \ours advantage widens under adversarial conditions because the Bayesian backbone accumulates evidence independently of the patient's communication style, even noisy or incomplete responses contribute partial information through Pearl-style soft evidence.

The distrustful persona yields the highest \ours DHS because both systems achieve comparable selective accuracy on committed cases, the difference is coverage: the standalone doctor becomes too uncertain to commit on most patients, while \ours still accumulates sufficient posterior confidence through Bayesian updating even from partial evidence.

\begin{figure}[h]
\centering
\includegraphics[width=0.85\textwidth]{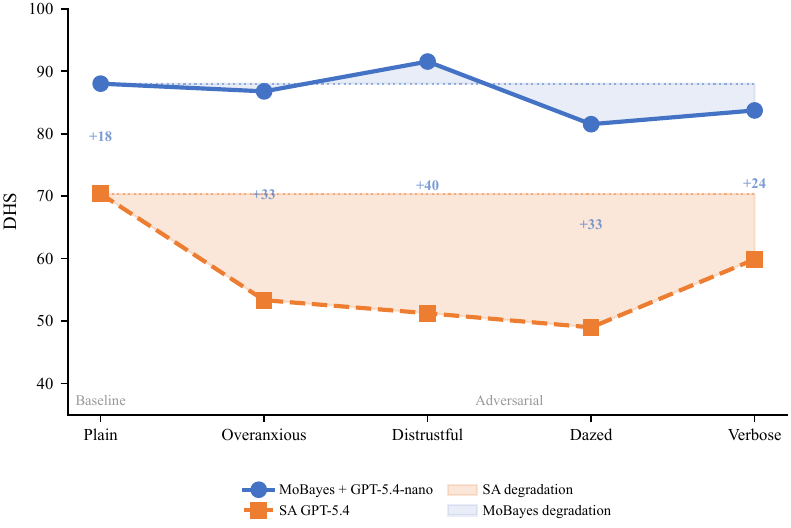}
\caption{DHS across patient personas. Shaded areas show degradation from the plain baseline (dotted lines). \ours (blue) remains near its baseline under all adversarial conditions; the standalone doctor (orange) collapses. The contrast between the two shaded areas illustrates the robustness advantage of structured Bayesian reasoning over end-to-end LLM inference.}
\label{fig:persona_radar}
\end{figure}

Figure~\ref{fig:persona_radar} visualises the consistent gap across all personas.

             % C.9 Persona robustness
\subsection{Accuracy by disease prevalence}
\label{app:prevalence_analysis}

A clinically important question is whether the system maintains accuracy across the full prevalence spectrum. We stratify the 49 DDXPlus diseases into three groups by rank-ordering their population prevalence and splitting into equal-sized terciles: \emph{common} (top~16), \emph{medium} (middle~17), and \emph{rare} (bottom~16).

Table~\ref{tab:prevalence_analysis} reports Top-1 accuracy, selective accuracy, coverage, and DHS for the strongest \ours configuration (GPT-5.4-nano) and the strongest standalone doctor (SA~GPT-5.4).

%% ============================================================
%% TABLE: Accuracy by disease prevalence (DDXPlus, rank tercile)
%% ============================================================

\begin{table}[h]
\caption{Diagnostic performance stratified by disease prevalence tercile on DDXPlus ($n{=}50$). \ours uses DHS-optimal $\tau^*$; SA uses self-determined abstention. Diseases rank-ordered by population prevalence and split into equal-sized groups.}
\label{tab:prevalence_analysis}
\centering
\small
\setlength{\tabcolsep}{4pt}
\begin{tabular}{@{}ll ccccc@{}}
\toprule
& & & & \multicolumn{3}{c}{\textbf{Selective Diagnosis}} \\
\cmidrule(lr){5-7}
\textbf{Prevalence} & \textbf{Model} & \textbf{Top-1} & \textbf{Top-3}
  & \textbf{Sel.\,Acc} & \textbf{Cov.} & \textbf{DHS} \\
\midrule
\multirow{2}{*}{Common ($n{=}20$)}
  & \ours + nano  & \textbf{90} & \textbf{100} & \textbf{90} & \textbf{100} & \cellcolor{dhsgreen!40}\textbf{95} \\
  & SA GPT-5.4   & 55 & 85 & 60 & 75 & \cellcolor{dhsred!10}67 \\
\midrule
\multirow{2}{*}{Medium ($n{=}17$)}
  & \ours + nano  & 71 & 88 & 75 & \textbf{94} & \cellcolor{dhsgreen!25}\textbf{84} \\
  & SA GPT-5.4   & 71 & 88 & \textbf{79} & 82 & \cellcolor{dhsgreen!20}80 \\
\midrule
\multirow{2}{*}{Rare ($n{=}13$)}
  & \ours + nano  & \textbf{69} & 77 & \textbf{75} & \textbf{92} & \cellcolor{dhsgreen!25}\textbf{83} \\
  & SA GPT-5.4   & 54 & 77 & 50 & 77 & \cellcolor{dhsred!15}61 \\
\midrule
\multirow{2}{*}{All ($n{=}50$)}
  & \ours + nano  & \textbf{78} & \textbf{90} & \textbf{81} & \textbf{96} & \cellcolor{dhsgreen!30}\textbf{88} \\
  & SA GPT-5.4   & 60 & 84 & 64 & 78 & \cellcolor{dhsred!5}70 \\
\bottomrule
\end{tabular}
\end{table}

\ours maintains a nearly flat DHS profile across the prevalence spectrum, whereas the standalone doctor degrades sharply on rare diseases. This gap arises from two complementary effects. First, the Bayesian engine conditions on a knowledge base whose likelihood tables are prevalence-agnostic: the Dirichlet counts encode symptom--disease associations regardless of how common a disease is. Second, the standalone doctor inherits the frequency bias of its LLM training corpus, diseases that appear more often in medical text receive disproportionate probability mass, systematically disadvantaging rare conditions.

The coverage column reinforces this distinction. \ours rarely abstains regardless of prevalence, while the standalone doctor abstains more aggressively on rare cases yet still achieves lower accuracy among the cases it does diagnose. In clinical terms, the Bayesian architecture provides both higher reliability and higher willingness to commit on the long tail of the disease distribution.
          % C.10 Accuracy by disease prevalence
\subsection{Failure mode analysis}
\label{app:failure_analysis}

On DDxPlus 50 cases (PatientSim), we classify each committed misdiagnosis (confidence $\geq \tau^*$, top-1 incorrect) along three independent, non-exclusive axes.

\begin{enumerate}[leftmargin=*,itemsep=3pt]

\item \textbf{KB Failure.}
We run an \emph{oracle} test: all ground-truth features are supplied at confidence $c{=}1.0$. We then measure the posterior gap between the oracle's top-1 and top-2 diseases. If this gap falls below a threshold $\gamma$ ($\gamma{=}0.80$), the KB cannot reliably discriminate the disease pair.

\item \textbf{LLM Failure.}
The LLM pipeline (verbaliser + patient simulator + parser) injected incorrect evidence into the engine. Two subtypes:
\begin{itemize}[nosep,leftmargin=1.5em]
    \item \emph{False Positive (FP)}: the engine asks about a feature \emph{absent} from the patient's ground-truth profile; the pipeline returns \texttt{yes}. If more than~2 such turns occur in a session, the case is flagged. The threshold reflects the empirical separation between successful sessions (mean~1.3 FP) and failure sessions (mean~3.5 FP).
    \item \emph{Wrong Evidence (WE)}: the engine asks about a feature \emph{present} in the ground-truth profile, but the extracted value contradicts the ground truth. Any occurrence triggers the flag.
\end{itemize}
This taxonomy is \emph{conservative}: it captures only detectable errors. Cases where the parser maps an uncertain response to \texttt{no} for an absent feature are not flagged, nor are confidence miscalibrations.

\item \textbf{Inference Failure.}
The KB is adequate and the evidence pipeline introduced no detectable errors, yet the engine converged to the wrong diagnosis. Two subtypes:
\begin{itemize}[nosep,leftmargin=1.5em]
    \item \emph{Close}: the ground truth remains in the top-3 posterior at session end, but the question budget or EIG policy did not resolve the differential.
    \item \emph{Diverged}: the ground truth is not in the top-3. The engine committed confidently to a wrong hypothesis.
\end{itemize}

\end{enumerate}

\paragraph{Results.}
Table~\ref{tab:failure_v2} reports the analysis for the six \ours sensors (committed failures only, $\tau^*$ per sensor).

\begin{table}[h]
\caption{Failure mode analysis on DDXPlus (committed misdiagnoses only). Axes are non-exclusive: a case may trigger multiple failure modes. Colour intensity: \textcolor{dhsred}{red}~=~more failures.}
\label{tab:failure_v2}
\centering
\small
\setlength{\tabcolsep}{3pt}
\begin{tabular}{@{}l c c cc c cc@{}}
\toprule
& & & \multicolumn{2}{c}{\textbf{LLM}} & & \multicolumn{2}{c}{\textbf{Inference}} \\
\cmidrule(lr){4-5} \cmidrule(lr){7-8}
\textbf{Sensor} & \textbf{Fail} & \textbf{KB}
  & \textbf{FP} & \textbf{WE}
  & & \textbf{Close} & \textbf{Div.} \\
\midrule
\cellcolor{dhsgreen!15}GPT-5.4-nano       & \cellcolor{dhsgreen!30}9   & 1 & \cellcolor{dhsgreen!25}6  & 0 & & 2 & \cellcolor{dhsgreen!30}0 \\
\cellcolor{dhsgreen!15}Gemini FL          & \cellcolor{dhsgreen!25}11  & 2 & \cellcolor{dhsgreen!20}4  & 1 & & 2 & 2 \\
\hdashline\noalign{\vskip 3pt}
Llama-4-Scout      & \cellcolor{dhsred!15}17  & 2 & \cellcolor{dhsred!5}6  & 2 & & 3 & \cellcolor{dhsred!10}4 \\
GPT-OSS-20B        & \cellcolor{dhsred!5}13  & 4 & \cellcolor{dhsred!5}5  & 0 & & 2 & 2 \\
Gemma 4 31B        & \cellcolor{dhsred!20}18  & 4 & \cellcolor{dhsred!20}11 & 0 & & 1 & 2 \\
MiniMax M2.5       & \cellcolor{dhsred!15}15  & 3 & \cellcolor{dhsred!15}9  & 1 & & 2 & \cellcolor{dhsgreen!20}0 \\
\midrule
\textbf{Total}     & \textbf{83} & \textbf{16} & \textbf{41} & \textbf{4} & & \textbf{12} & \textbf{10} \\
\textbf{\%}        &             & 19\%        & 49\%        & 5\%        & & 14\%        & 12\% \\
\bottomrule
\end{tabular}
\end{table}

\paragraph{KB failures are patient-specific, not sensor-specific.}
The KB failure count varies across sensors (1--4) not because the KB changes, but because stronger sensors resolve cases that weaker ones cannot. The same patients trigger oracle failure regardless of sensor. However, a capable sensor may still overcome a marginal KB gap through better evidence collection, while a weaker sensor fails on those cases and gets counted.

\paragraph{LLM failure dominates and scales with sensor quality.}
The LLM pipeline accounts for the majority of committed errors, with false positives as the dominant subtype. Critically, this failure mode scales with sensor capability: frontier sensors produce substantially fewer false positives than smaller models (Table~\ref{tab:failure_v2}). This suggests that scaling the sensor LLM, without changing the Bayesian engine, directly reduces the dominant failure mode.

\paragraph{Inference failures confirm the architectural thesis.}
The strongest sensors have zero diverged inference failures: every committed error is either a KB limitation or an evidence collection problem. When the LLM pipeline delivers accurate evidence, the Bayesian engine reliably converges to the correct diagnosis. Weaker sensors produce noisier evidence that steers the EIG policy toward unproductive questions, causing the engine to commit to wrong hypotheses even when the KB contains sufficient information.
             % C.11 Failure mode analysis
\subsection{Disease scaling analysis}
\label{app:scaling}

We evaluate how diagnostic performance scales with the size of the candidate disease space. Random subsets of $K \in \{10, 20, 30, 40\}$ diseases are sampled from DDXPlus, and one patient per disease is evaluated for both \ours and the standalone doctor.

\begin{figure}[h]
\centering
% Center the left side vertically
\begin{minipage}[c]{0.48\textwidth}
    \centering
    \includegraphics[width=\textwidth]{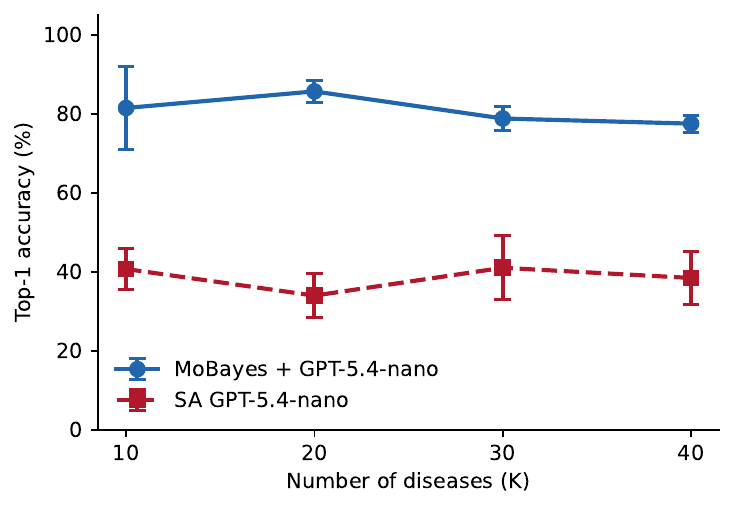}
    \captionof{figure}{Top-1 accuracy vs.\ KB size $K$. \ours remains stable across a $4\times$ increase in disease space; the standalone doctor is flat regardless of $K$.}
    \label{fig:scaling}
\end{minipage}
\hfill
% Center the right side vertically
\begin{minipage}[c]{0.48\textwidth}
    \centering
    \captionof{table}{Disease scaling: Top-1 accuracy on DDXPlus random subsets (3 seeds, $N=K$ patients). $\Delta$: \ours $-$ SA.}
\label{tab:scaling}
\vspace{4pt} % Fine-tune the gap between caption and table
\small
\begin{tabular}{c cc c} % Removed @{} to give the cells some breathing room
\toprule
$\boldsymbol{K}$ & \textbf{\ours} & \textbf{SA} & $\boldsymbol{\Delta}$ \\
\midrule
10 & \cellcolor{dhsgreen!15}82 & \cellcolor{dhsred!10}41 & \cellcolor{dhsgreen!20}+41 \\
20 & \cellcolor{dhsgreen!20}86 & \cellcolor{dhsred!15}34 & \cellcolor{dhsgreen!30}+52 \\
30 & \cellcolor{dhsgreen!15}79 & \cellcolor{dhsred!10}41 & \cellcolor{dhsgreen!20}+38 \\
40 & \cellcolor{dhsgreen!15}78 & \cellcolor{dhsred!10}39 & \cellcolor{dhsgreen!20}+39 \\
\bottomrule
\end{tabular}
\end{minipage}
\end{figure}

\ours maintains stable accuracy as the disease space quadruples, because its inference complexity scales with the KB structure rather than with the LLM's reasoning capacity. The standalone doctor, by contrast, operates over its own implicit disease space regardless of $K$---it does not know how many candidates the KB contains---and its performance remains flat. Providing the candidate list explicitly would likely degrade the standalone doctor further at higher $K$, as a longer list introduces confusion without a structured mechanism to leverage it.
             % C.12 Disease scaling analysis
\subsection{Diagnostic session walkthroughs}
\label{app:case_studies}

\begin{figure}[h]
\centering
\includegraphics[width=0.55\textwidth]{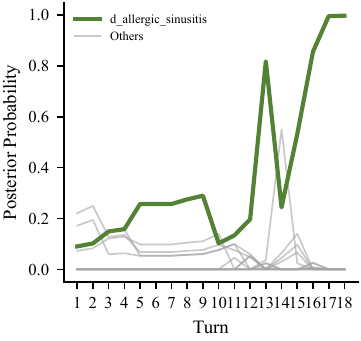}
\caption{Posterior trajectory for a single session (Allergic sinusitis, GPT-5.4-nano). The ground-truth disease (green) starts at 9\%, competes with alternatives, and converges to near-certainty.}
\label{fig:posterior_entropy}
\end{figure}

Figure~\ref{fig:posterior_entropy} illustrates the engine's belief dynamics on a representative case: competing hypotheses rise and fall as evidence accumulates, and the posterior of the ground-truth disease converges to near-certainty.

We present an example session below. 

\newcommand{\pbar}[2]{%
  \tikz[baseline=-0.3em]{%
    \fill[#1] (0,0) rectangle ({#2*1.4cm},0.16cm);%
    \draw[gray!40] (0,0) rectangle (1.4cm,0.16cm);%
  }%
}

\clearpage
%% ─── CASE 1: Spontaneous pneumothorax (success, gradual) ───
\begin{tcolorbox}[breakable, title={\textbf{Case 1: Spontaneous pneumothorax} \hfill \textcolor{green!50!black}{Correct, confidence $= 0.999$}}, colback=white, colframe=blue!50!black, fonttitle=\bfseries\small, left=3pt, right=3pt, top=3pt, bottom=3pt]

\begin{minipage}[t]{0.64\textwidth}
\raggedright\footnotesize

\noindent\textit{``I have chest pain even at rest, upper chest pain, and pleuritic chest pain''}

\medskip
\textcolor{blue!70!black}{\textbf{T1 Dr:}} Does your thoracic pain spread or radiate to another area, such as your chest, ribs, shoulder, or around to the front of your body?\\
\textcolor{orange!70!black}{\textbf{Pt:}} \textit{``No. My pain is in my upper chest and both breasts, and it doesn't spread or radiate to other areas.''}

\medskip
\textcolor{blue!70!black}{\textbf{T2 Dr:}} Do you have chronic obstructive pulmonary disease (COPD), such as chronic bronchitis or emphysema?\\
\textcolor{orange!70!black}{\textbf{Pt:}} \textit{``No. I don't have COPD or chronic bronchitis/emphysema that I know of.''}

\medskip
\textcolor{blue!70!black}{\textbf{T3 Dr:}} What color is the rash? For example, is it red, pink, dark, pale, or yellow, or is there no rash?\\
\textcolor{orange!70!black}{\textbf{Pt:}} \textit{``There is no rash.''}

\medskip
\textcolor{blue!70!black}{\textbf{T4 Dr:}} Do you feel any pain in the upper middle part of your abdomen (the area just below your breastbone)?\\
\textcolor{orange!70!black}{\textbf{Pt:}} \textit{``No. My pain is in my upper chest area, not in the upper middle abdomen below the breastbone.''}

\medskip
\textcolor{blue!70!black}{\textbf{T5 Dr:}} Have you had any surgery or a procedure that required anesthesia within the last month?\\
\textcolor{orange!70!black}{\textbf{Pt:}} \textit{``No. I haven't had any surgery or anesthesia within the last month.''}

\medskip
\textcolor{blue!70!black}{\textbf{T6 Dr:}} Do you feel your heart beating fast, irregularly, or having palpitations right now?\\
\textcolor{orange!70!black}{\textbf{Pt:}} \textit{``No. I don't feel my heart racing or having palpitations right now.''}

\medskip
\textcolor{blue!70!black}{\textbf{T7 Dr:}} Do you currently smoke cigarettes?\\
\textcolor{orange!70!black}{\textbf{Pt:}} \textit{``Yes. I currently smoke cigarettes.''}

\medskip
\textcolor{blue!70!black}{\textbf{T8 Dr:}} Do you consider yourself significantly overweight compared to other people of the same height?\\
\textcolor{orange!70!black}{\textbf{Pt:}} \textit{``No. I don't consider myself significantly overweight.''}

\medskip
\textcolor{blue!70!black}{\textbf{T9 Dr:}} Do you constantly feel fatigued, or do you have sleep that doesn't leave you feeling rested?\\
\textcolor{orange!70!black}{\textbf{Pt:}} \textit{``No. I'm not constantly fatigued, and I don't have sleep that never leaves me rested.''}

\medskip
\textcolor{blue!70!black}{\textbf{T10 Dr:}} Do you drink alcohol excessively, or do you feel you have an addiction to alcohol?\\
\textcolor{orange!70!black}{\textbf{Pt:}} \textit{``No. I don't drink excessively and I don't feel addicted to alcohol.''}

\medskip
\textcolor{blue!70!black}{\textbf{T11 Dr:}} Have you ever smoked cigarettes in the past, but you don't smoke now (a former smoker)?\\
\textcolor{orange!70!black}{\textbf{Pt:}} \textit{``No. I'm still smoking cigarettes now, so I'm not a former smoker.''}

\medskip
\textcolor{blue!70!black}{\textbf{T12 Dr:}} When you feel the pain, does it feel like a tugging or pulling sensation?\\
\textcolor{orange!70!black}{\textbf{Pt:}} \textit{``No. It feels more like a knife-stroke, heartbreaking pain, not tugging or pulling.''}

\medskip
\textcolor{blue!70!black}{\textbf{T13 Dr:}} Does your pain feel exhausting, like it wears you out or drains your energy?\\
\textcolor{orange!70!black}{\textbf{Pt:}} \textit{``No. It's very severe, but it doesn't feel exhausting or like it drains my energy.''}

\medskip
\textcolor{blue!70!black}{\textbf{T14 Dr:}} Have you ever had a spontaneous pneumothorax (a collapsed lung that happened without a clear injury or accident)?\\
\textcolor{orange!70!black}{\textbf{Pt:}} \textit{``Yes. I've had a spontaneous pneumothorax before.''}

\medskip
\textcolor{blue!70!black}{\textbf{T15 Dr:}} Is your pain heartbreaking or emotionally overwhelming in a way that feels hard to bear?\\
\textcolor{orange!70!black}{\textbf{Pt:}} \textit{``Yes. It's heartbreaking and hard to bear.''}

\end{minipage}%
\hfill
\begin{minipage}[t]{0.34\textwidth}
\raggedright\tiny
\textbf{\scriptsize Posterior (Sp.\ pneumo)}\\[3pt]
\textbf{T1:} \pbar{blue!30}{0.219} .219\\[1pt]
\textbf{T2:} \pbar{blue!18}{0.118} .118\\[1pt]
\textbf{T3:} \pbar{blue!20}{0.137} .137\\[1pt]
\textbf{T4:} \pbar{blue!22}{0.162} .162\\[1pt]
\textbf{T5:} \pbar{blue!25}{0.186} .186\\[1pt]
\textbf{T6:} \pbar{blue!27}{0.205} .205\\[1pt]
\textbf{T7:} \pbar{blue!45}{0.404} \textbf{.404}\\[1pt]
\textbf{T8:} \pbar{blue!47}{0.442} .442\\[1pt]
\textbf{T9:} \pbar{blue!48}{0.462} .462\\[1pt]
\textbf{T10:} \pbar{blue!49}{0.472} .472\\[1pt]
\textbf{T11:} \pbar{blue!50}{0.480} .480\\[1pt]
\textbf{T12:} \pbar{blue!50}{0.485} .485\\[1pt]
\textbf{T13:} \pbar{blue!50}{0.488} .488\\[1pt]
\textbf{T14:} \pbar{blue!70}{0.976} \textbf{.976}\\[1pt]
\textbf{T15:} \pbar{blue!70}{0.999} \textbf{.999}\\
\end{minipage}

\end{tcolorbox}

\clearpage
\subsection{Bootstrap confidence intervals}
\label{app:significance}

To quantify the uncertainty in the point estimates, we compute bootstrap CIs ($B{=}2000$, percentile method) for every cell of Tables~\ref{tab:main} and~\ref{tab:combined_n100}. Cell entries are point\,$_{[\text{2.5\%, 97.5\%}]}$. For \ours, $\tau^*$ is locked at the LOOCV value once on the original data (per benchmark, per half for the combined $n{=}100$ table) and held fixed across replicates. Tables~\ref{tab:ci_main} and~\ref{tab:ci_combined} report the full CIs.

%% ============================================================
%% Bootstrap 95% CIs for paper Tables 1, 2, 3, 22.
%% Auto-generated by scripts/ci_tables.py --appendix.
%% B=2000, seed=42.  Cells show T-1, DHS (point [lo, hi]) plus token count.
%% ============================================================

\begin{table}[t]
\centering
\caption{Bootstrap 95\% CIs for Table~\ref{tab:main} (frontier standalone vs.\ \discuss{\ours}, $n{=}50$ per cell). $B=2000$; cells show point [2.5\%, 97.5\%]. DHS coloured per benchmark column.}
\label{tab:ci_main}
\footnotesize
\setlength{\tabcolsep}{4pt}
\renewcommand{\arraystretch}{1.05}
\resizebox{\textwidth}{!}{%
\begin{tabular}{@{}l ccc ccc ccc@{}}
\toprule
 & \multicolumn{3}{c}{\textbf{DDxPlus PatientSim}} & \multicolumn{3}{c}{\textbf{DDxPlus HT-sim}} & \multicolumn{3}{c}{\textbf{AgentClinic-MedQA}} \\
 \cmidrule(lr){2-4} \cmidrule(lr){5-7} \cmidrule(lr){8-10}
\textbf{Method} & T-1 & SA & DHS$\,\uparrow$ & T-1 & SA & DHS$\,\uparrow$ & T-1 & SA & DHS$\,\uparrow$ \\
\midrule
\multicolumn{10}{l}{\textit{Standalone LLM doctors (frontier)}} \\
\midrule
SA GPT-5.4 & 60.0\,\textsubscript{[46.0,72.0]} & 64.1\,\textsubscript{[50.0,78.9]} & \cellcolor{dhsred!11}70.4\,\textsubscript{[59.3,79.8]} & 58.0\,\textsubscript{[44.0,72.0]} & 76.0\,\textsubscript{[57.1,92.0]} & \cellcolor{dhsred!15}60.3\,\textsubscript{[48.0,70.9]} & 60.0\,\textsubscript{[46.0,74.0]} & 62.8\,\textsubscript{[47.7,77.3]} & \cellcolor{dhsred!14}72.6\,\textsubscript{[61.5,82.1]} \\
SA Gemini 3.1 Pro & 60.0\,\textsubscript{[46.0,74.0]} & 62.5\,\textsubscript{[49.0,75.5]} & \cellcolor{dhsgreen!14}75.7\,\textsubscript{[64.8,85.3]} & 60.0\,\textsubscript{[46.0,74.0]} & 73.2\,\textsubscript{[59.0,85.7]} & \cellcolor{dhsgreen!20}77.3\,\textsubscript{[66.9,85.9]} & 66.0\,\textsubscript{[54.0,80.0]} & 68.8\,\textsubscript{[55.1,81.6]} & \cellcolor{dhsgreen!17}80.1\,\textsubscript{[70.1,88.9]} \\
SA Llama-4-Maverick & 56.0\,\textsubscript{[42.0,70.0]} & 56.5\,\textsubscript{[41.7,70.5]} & \cellcolor{dhsred!11}70.0\,\textsubscript{[57.5,80.4]} & 46.0\,\textsubscript{[32.0,60.0]} & 56.0\,\textsubscript{[36.4,75.0]} & \cellcolor{dhsred!20}52.8\,\textsubscript{[39.8,63.2]} & 56.0\,\textsubscript{[42.0,70.0]} & 58.3\,\textsubscript{[44.7,72.9]} & \cellcolor{dhsred!14}72.6\,\textsubscript{[60.6,82.9]} \\
SA Qwen 3.6 Plus & 48.0\,\textsubscript{[34.0,62.0]} & 51.1\,\textsubscript{[36.7,64.6]} & \cellcolor{dhsred!15}66.2\,\textsubscript{[52.2,76.9]} & 60.0\,\textsubscript{[46.0,74.0]} & 65.0\,\textsubscript{[50.0,79.0]} & \cellcolor{dhsgreen!12}71.7\,\textsubscript{[60.3,80.9]} & 64.0\,\textsubscript{[50.0,76.0]} & 63.3\,\textsubscript{[50.0,76.0]} & \cellcolor{dhsgreen!8}76.9\,\textsubscript{[65.8,85.5]} \\
SA Kimi K2.5 & 42.0\,\textsubscript{[28.0,56.0]} & 54.1\,\textsubscript{[37.5,71.0]} & \cellcolor{dhsred!20}62.5\,\textsubscript{[49.2,73.2]} & 42.0\,\textsubscript{[28.0,56.0]} & 40.7\,\textsubscript{[23.1,60.7]} & \cellcolor{dhsred!25}46.4\,\textsubscript{[31.6,58.3]} & 54.0\,\textsubscript{[40.0,68.0]} & 57.4\,\textsubscript{[43.5,71.4]} & \cellcolor{dhsred!16}71.3\,\textsubscript{[59.1,81.5]} \\
SA GPT-OSS-120B & 40.0\,\textsubscript{[26.0,54.0]} & 40.8\,\textsubscript{[27.1,55.1]} & \cellcolor{dhsred!25}57.6\,\textsubscript{[42.2,70.5]} & 54.0\,\textsubscript{[40.0,68.0]} & 54.5\,\textsubscript{[37.5,71.0]} & \cellcolor{dhsred!15}59.7\,\textsubscript{[46.9,70.2]} & 48.0\,\textsubscript{[34.0,62.0]} & 49.0\,\textsubscript{[34.7,63.0]} & \cellcolor{dhsred!25}65.3\,\textsubscript{[51.2,76.5]} \\
\midrule
\multicolumn{10}{l}{\textit{\discuss{\ours} + sensor LLM}} \\
\midrule
+ GPT-5.4-nano & 78.0\,\textsubscript{[66.0,90.0]} & 81.2\,\textsubscript{[68.9,91.7]} & \cellcolor{dhsgreen!40}88.0\,\textsubscript{[79.8,94.7]} & 80.0\,\textsubscript{[68.0,90.0]} & 83.3\,\textsubscript{[72.0,93.6]} & \cellcolor{dhsgreen!37}89.2\,\textsubscript{[82.0,94.8]} & 76.0\,\textsubscript{[64.0,88.0]} & 79.2\,\textsubscript{[67.3,89.8]} & \cellcolor{dhsgreen!36}86.8\,\textsubscript{[79.0,93.6]} \\
+ Gemini 3.1 Flash Lite & 74.0\,\textsubscript{[62.0,86.0]} & 77.1\,\textsubscript{[64.6,88.0]} & \cellcolor{dhsgreen!35}85.5\,\textsubscript{[77.2,92.6]} & 84.0\,\textsubscript{[74.0,94.0]} & 85.7\,\textsubscript{[75.5,94.0]} & \cellcolor{dhsgreen!40}91.4\,\textsubscript{[85.3,96.9]} & 78.0\,\textsubscript{[66.0,88.0]} & 86.7\,\textsubscript{[76.6,95.7]} & \cellcolor{dhsgreen!40}88.3\,\textsubscript{[81.1,93.8]} \\
+ GPT-OSS-20B & 60.0\,\textsubscript{[46.0,74.0]} & 68.2\,\textsubscript{[54.5,81.4]} & \cellcolor{dhsgreen!16}76.8\,\textsubscript{[66.6,85.7]} & 50.0\,\textsubscript{[36.0,64.0]} & 61.0\,\textsubscript{[46.3,75.6]} & \cellcolor{dhsgreen!9}69.9\,\textsubscript{[58.0,79.5]} & 50.0\,\textsubscript{[36.0,64.0]} & 58.1\,\textsubscript{[43.2,73.2]} & \cellcolor{dhsred!19}69.4\,\textsubscript{[57.2,79.2]} \\
+ MiniMax M2.5 & 60.0\,\textsubscript{[46.0,74.0]} & 66.7\,\textsubscript{[53.3,80.0]} & \cellcolor{dhsgreen!16}76.6\,\textsubscript{[66.3,85.3]} & 60.0\,\textsubscript{[46.0,74.0]} & 75.0\,\textsubscript{[61.5,88.1]} & \cellcolor{dhsgreen!20}77.4\,\textsubscript{[67.8,85.9]} & 50.0\,\textsubscript{[36.0,64.0]} & 52.1\,\textsubscript{[37.5,67.3]} & \cellcolor{dhsred!22}67.5\,\textsubscript{[53.9,79.0]} \\
+ Gemma 4 31B & 58.0\,\textsubscript{[44.0,72.0]} & 61.7\,\textsubscript{[47.8,75.5]} & \cellcolor{dhsgreen!12}74.5\,\textsubscript{[62.9,84.0]} & 82.0\,\textsubscript{[70.0,92.0]} & 83.7\,\textsubscript{[72.9,93.8]} & \cellcolor{dhsgreen!38}90.3\,\textsubscript{[82.9,95.8]} & 64.0\,\textsubscript{[50.0,78.0]} & 65.3\,\textsubscript{[52.0,79.2]} & \cellcolor{dhsgreen!12}78.4\,\textsubscript{[67.5,87.6]} \\
+ Llama-4-Scout & 56.0\,\textsubscript{[42.0,70.0]} & 60.9\,\textsubscript{[46.7,75.0]} & \cellcolor{dhsgreen!9}73.3\,\textsubscript{[61.5,83.1]} & 80.0\,\textsubscript{[68.0,90.0]} & 87.0\,\textsubscript{[76.6,95.7]} & \cellcolor{dhsgreen!37}89.4\,\textsubscript{[82.1,94.9]} & 68.0\,\textsubscript{[56.0,80.0]} & 79.1\,\textsubscript{[66.7,90.9]} & \cellcolor{dhsgreen!24}82.4\,\textsubscript{[73.6,89.4]} \\
\bottomrule
\end{tabular}%
}
\end{table}

\begin{table}[t]
\centering
\caption{Bootstrap 95\% CIs and token totals for Table~\ref{tab:combined_n100} (combined $n{=}100$). $B=2000$; cells show point [2.5\%, 97.5\%]. Tok/case is the average doctor + patient prompt+completion tokens per diagnostic conversation.}
\label{tab:ci_combined}
\footnotesize
\setlength{\tabcolsep}{4pt}
\renewcommand{\arraystretch}{1.05}
\resizebox{\textwidth}{!}{%
\begin{tabular}{@{}l ccc ccc ccc@{}}
\toprule
 & \multicolumn{3}{c}{\textbf{DDxPlus PatientSim}} & \multicolumn{3}{c}{\textbf{DDxPlus HT-sim}} & \multicolumn{3}{c}{\textbf{AgentClinic-MedQA}} \\
 \cmidrule(lr){2-4} \cmidrule(lr){5-7} \cmidrule(lr){8-10}
\textbf{Method} & T-1 & DHS$\,\uparrow$ & Tok/case & T-1 & DHS$\,\uparrow$ & Tok/case & T-1 & DHS$\,\uparrow$ & Tok/case \\
\midrule
\multicolumn{10}{l}{\textit{Prompt-based}} \\
\midrule
Bayesian-CoT & 60.0\,\textsubscript{[51.0,69.0]} & \cellcolor{dhsgreen!25}74.0\,\textsubscript{[66.1,80.8]} & 46.6K & 41.0\,\textsubscript{[31.0,51.0]} & \cellcolor{dhsred!10}44.5\,\textsubscript{[34.7,52.8]} & 34.1K & 63.0\,\textsubscript{[53.0,73.0]} & \cellcolor{dhsgreen!21}79.3\,\textsubscript{[72.7,85.4]} & 37.8K \\
Chain-of-Thought (CoT) & 52.0\,\textsubscript{[42.0,62.0]} & \cellcolor{dhsgreen!21}69.0\,\textsubscript{[60.6,76.6]} & 30.8K & 41.0\,\textsubscript{[31.0,51.0]} & \cellcolor{dhsred!17}29.2\,\textsubscript{[19.1,38.4]} & 15.8K & 71.0\,\textsubscript{[62.0,80.0]} & \cellcolor{dhsgreen!31}84.1\,\textsubscript{[78.4,89.3]} & 22.5K \\
Closed-World (CW) & 40.0\,\textsubscript{[31.0,49.0]} & \cellcolor{dhsgreen!8}55.4\,\textsubscript{[45.8,63.2]} & 32.0K & 45.0\,\textsubscript{[35.0,55.0]} & \cellcolor{dhsred!20}20.7\,\textsubscript{[11.0,30.2]} & 23.9K & 40.0\,\textsubscript{[31.0,50.0]} & \cellcolor{dhsred!19}62.3\,\textsubscript{[54.2,69.8]} & 25.7K \\
DDx-CoT & 42.0\,\textsubscript{[32.0,52.0]} & \cellcolor{dhsred!8}54.6\,\textsubscript{[44.9,62.4]} & 28.8K & 37.0\,\textsubscript{[28.0,46.0]} & \cellcolor{dhsred!22}16.1\,\textsubscript{[7.4,25.3]} & 18.5K & 40.0\,\textsubscript{[31.0,50.0]} & \cellcolor{dhsred!19}61.8\,\textsubscript{[53.5,69.6]} & 23.7K \\
\midrule
\multicolumn{10}{l}{\textit{Information-pursuit}} \\
\midrule
UoT & 28.0\,\textsubscript{[19.0,37.0]} & \cellcolor{dhsred!13}43.8\,\textsubscript{[31.9,54.0]} & 37.7K & 5.0\,\textsubscript{[1.0,9.0]} & \cellcolor{dhsred!25}9.5\,\textsubscript{[2.0,16.5]} & 28.2K & 38.0\,\textsubscript{[29.0,47.0]} & \cellcolor{dhsred!25}56.1\,\textsubscript{[45.7,65.3]} & 34.6K \\
\midrule
\multicolumn{10}{l}{\textit{Agent-based}} \\
\midrule
MEDDxAgent (paper-best) & 45.0\,\textsubscript{[35.0,55.0]} & \cellcolor{dhsgreen!14}62.1\,\textsubscript{[51.9,71.0]} & 108.3K & 56.0\,\textsubscript{[46.0,66.0]} & \cellcolor{dhsgreen!26}71.8\,\textsubscript{[63.0,79.5]} & 88.1K & 71.0\,\textsubscript{[62.0,80.0]} & \cellcolor{dhsgreen!29}83.0\,\textsubscript{[76.5,88.9]} & 106.8K \\
MEDDxAgent (closed-world abl.) & 41.0\,\textsubscript{[31.0,51.0]} & \cellcolor{dhsgreen!11}58.2\,\textsubscript{[47.3,67.5]} & 79.9K & 39.0\,\textsubscript{[30.0,49.0]} & \cellcolor{dhsgreen!13}56.1\,\textsubscript{[46.2,65.8]} & 71.9K & 80.0\,\textsubscript{[72.0,88.0]} & \cellcolor{dhsgreen!40}88.9\,\textsubscript{[83.7,93.6]} & 69.0K \\
AgentClinic & 27.0\,\textsubscript{[19.0,36.0]} & \cellcolor{dhsred!10}52.0\,\textsubscript{[42.8,59.9]} & 18.1K & 28.0\,\textsubscript{[19.0,37.0]} & \cellcolor{dhsgreen!10}52.0\,\textsubscript{[41.9,60.2]} & 29.7K & 66.0\,\textsubscript{[56.0,75.0]} & \cellcolor{dhsgreen!22}79.7\,\textsubscript{[72.2,85.8]} & 6.2K \\
MediQ-Expert (force-final) & 43.0\,\textsubscript{[33.0,52.0]} & \cellcolor{dhsgreen!13}60.1\,\textsubscript{[49.6,68.4]} & 39.3K & 34.0\,\textsubscript{[25.0,43.0]} & \cellcolor{dhsgreen!9}50.7\,\textsubscript{[40.0,60.1]} & 41.9K & 56.0\,\textsubscript{[46.0,66.0]} & \cellcolor{dhsred!9}71.8\,\textsubscript{[63.0,79.5]} & 42.6K \\
MediQ-Expert (strict abstain) & 32.0\,\textsubscript{[23.0,41.0]} & \cellcolor{dhsgreen!9}56.2\,\textsubscript{[46.7,64.0]} & 39.3K & 25.0\,\textsubscript{[17.0,34.0]} & \cellcolor{dhsred!8}49.6\,\textsubscript{[40.2,57.4]} & 41.9K & 37.0\,\textsubscript{[27.0,47.0]} & \cellcolor{dhsred!23}57.8\,\textsubscript{[48.5,66.2]} & 42.6K \\
\midrule
\multicolumn{10}{l}{\textit{Fine-tuned (specialised model)}} \\
\midrule
CoD / DiagnosisGPT-34B & 11.0\,\textsubscript{[5.0,18.0]} & \cellcolor{dhsred!25}19.8\,\textsubscript{[9.5,30.5]} & 13.4K & 4.0\,\textsubscript{[1.0,8.0]} & \cellcolor{dhsred!25}10.7\,\textsubscript{[2.6,20.3]} & 12.7K & 46.0\,\textsubscript{[36.0,56.0]} & \cellcolor{dhsred!13}67.2\,\textsubscript{[58.9,74.1]} & 31.2K \\
\midrule
\multicolumn{10}{l}{\textit{Ours}} \\
\midrule
\textbf{\ours} (gpt-5.4-nano) & 82.0\,\textsubscript{[74.0,89.0]} & \cellcolor{dhsgreen!40}90.3\,\textsubscript{[85.6,94.3]} & 16.4K & 81.0\,\textsubscript{[73.0,88.0]} & \cellcolor{dhsgreen!40}90.0\,\textsubscript{[85.3,93.8]} & 18.0K & 70.0\,\textsubscript{[61.0,79.0]} & \cellcolor{dhsgreen!29}83.1\,\textsubscript{[77.1,88.5]} & 18.9K \\
\bottomrule
\end{tabular}%
}
\end{table}

\paragraph{Combined point estimates with category-internal DHS ranking.}
Table~\ref{tab:combined_n100} reports the same combined-$n{=}100$ comparison as Table~\ref{tab:ci_combined} but as point estimates, broken out into all five metrics (T-1, T-3, SA, Cov, DHS), with rows ranked by average DHS within each category.

%% =============================================================
%% TABLE: Combined n=100 baseline comparison.
%% Auto-generated by scripts/render_paper_tables.py.
%% Rows sorted within categories by avg DHS desc.
%% =============================================================

\begin{table}[t]
\centering
\caption{Combined $n{=}100$ baseline comparison: canonical $n{=}50$ plus a disjoint $n{=}50$ replication. Per Appendix~\ref{app:threshold_tuning}, \discuss{\ours}'s threshold $\tau^*$ is LOOCV-tuned on each 50-case set; combined SA, Cov, DHS aggregate committed/correct counts across halves rather than re-tuning on $n{=}100$. T-$k$ is naive (independent of commitment); SA, Cov, DHS are commit-based. DHS coloured per benchmark column; rows ranked within each category by average DHS.}
\label{tab:combined_n100}
\footnotesize
\setlength{\tabcolsep}{3pt}
\renewcommand{\arraystretch}{1.05}
\resizebox{\textwidth}{!}{%
\begin{tabular}{@{}l ccccc @{\hskip 0.6em} ccccc @{\hskip 0.6em} ccccc@{}}
\toprule
 & \multicolumn{5}{c}{\textbf{DDxPlus 100 (\ours PatientSim)}} & \multicolumn{5}{c}{\textbf{DDxPlus 100 (MEDDxAgent HT-sim)}} & \multicolumn{5}{c}{\textbf{AgentClinic-MedQA 100}} \\
\cmidrule(lr){2-6} \cmidrule(lr){7-11} \cmidrule(lr){12-16}
\textbf{Method} & T-1 & T-3 & SA & Cov. & DHS$\,\uparrow$ & T-1 & T-3 & SA & Cov. & DHS$\,\uparrow$ & T-1 & T-3 & SA & Cov. & DHS$\,\uparrow$ \\
\midrule
\multicolumn{16}{l}{\textit{Prompt-based}} \\
\midrule
DDx-CoT                            & 42.0 & 75.0 & 50.8 & 59.0 & \cellcolor{dhsred!8}54.6 & 37.0 & 54.0 & 77.8 & 9.0 & \cellcolor{dhsred!22}16.1 & 40.0 & 75.0 & 78.4 & 51.0 & \cellcolor{dhsred!19}61.8 \\
Closed-World (CW)                  & 40.0 & 64.0 & 50.0 & 62.0 & \cellcolor{dhsgreen!8}55.4 & 45.0 & 59.0 & 75.0 & 12.0 & \cellcolor{dhsred!20}20.7 & 40.0 & 73.0 & 75.5 & 53.0 & \cellcolor{dhsred!19}62.3 \\
Chain-of-Thought (CoT)             & 52.0 & 71.0 & 54.8 & 93.0 & \cellcolor{dhsgreen!21}69.0 & 41.0 & 59.0 & 63.2 & 19.0 & \cellcolor{dhsred!17}29.2 & 71.0 & 83.0 & \textbf{88.8} & 80.0 & \cellcolor{dhsgreen!31}84.1 \\
Bayesian-CoT                       & 60.0 & 75.0 & 62.0 & 92.0 & \cellcolor{dhsgreen!25}74.0 & 41.0 & 55.0 & 58.3 & 36.0 & \cellcolor{dhsred!10}44.5 & 63.0 & 77.0 & 76.8 & 82.0 & \cellcolor{dhsgreen!21}79.3 \\
\midrule
\multicolumn{16}{l}{\textit{Information-pursuit}} \\
\midrule
UoT                                & 28.0 & 41.0 & 28.0 & \textbf{100.0} & \cellcolor{dhsred!13}43.8 & 5.0 & 9.0 & 5.0 & \textbf{100.0} & \cellcolor{dhsred!25}9.5 & 38.0 & 52.0 & 39.6 & 96.0 & \cellcolor{dhsred!25}56.1 \\
\midrule
\multicolumn{16}{l}{\textit{Agent-based}} \\
\midrule
MediQ-Expert (strict abstain)      & 32.0 & -- & 50.8 & 63.0 & \cellcolor{dhsgreen!9}56.2 & 25.0 & -- & 56.8 & 44.0 & \cellcolor{dhsred!8}49.6 & 37.0 & -- & 84.1 & 44.0 & \cellcolor{dhsred!23}57.8 \\
MediQ-Expert (force-final)         & 43.0 & -- & 43.0 & \textbf{100.0} & \cellcolor{dhsgreen!13}60.1 & 34.0 & -- & 34.0 & \textbf{100.0} & \cellcolor{dhsgreen!9}50.7 & 56.0 & -- & 56.0 & \textbf{100.0} & \cellcolor{dhsred!9}71.8 \\
AgentClinic                        & 27.0 & 29.0 & 50.9 & 53.0 & \cellcolor{dhsred!10}52.0 & 28.0 & 29.0 & 43.8 & 64.0 & \cellcolor{dhsgreen!10}52.0 & 66.0 & 68.0 & 66.7 & 99.0 & \cellcolor{dhsgreen!22}79.7 \\
MEDDxAgent (closed-world abl.)     & 41.0 & 68.0 & 41.0 & \textbf{100.0} & \cellcolor{dhsgreen!11}58.2 & 39.0 & 60.0 & 39.0 & \textbf{100.0} & \cellcolor{dhsgreen!13}56.1 & \textbf{80.0} & 85.0 & 80.0 & \textbf{100.0} & \cellcolor{dhsgreen!40}\textbf{88.9} \\
MEDDxAgent (paper-best)            & 45.0 & 73.0 & 45.0 & \textbf{100.0} & \cellcolor{dhsgreen!14}62.1 & 56.0 & 75.0 & 56.0 & \textbf{100.0} & \cellcolor{dhsgreen!26}71.8 & 71.0 & \textbf{88.0} & 71.0 & \textbf{100.0} & \cellcolor{dhsgreen!29}83.0 \\
\midrule
\multicolumn{16}{l}{\textit{Fine-tuned (specialised model)}} \\
\midrule
CoD / DiagnosisGPT-34B             & 11.0 & 17.0 & 11.0 & \textbf{100.0} & \cellcolor{dhsred!25}19.8 & 4.0 & 7.0 & 5.8 & 69.0 & \cellcolor{dhsred!25}10.7 & 46.0 & 46.0 & 78.0 & 59.0 & \cellcolor{dhsred!13}67.2 \\
\midrule
\multicolumn{16}{l}{\textit{Ours}} \\
\midrule
\textbf{\ours} (gpt-5.4-nano)      & \textbf{82.0} & \textbf{92.0} & \textbf{84.5} & 97.0 & \cellcolor{dhsgreen!40}\textbf{90.3} & \textbf{81.0} & \textbf{92.0} & \textbf{87.1} & 93.0 & \cellcolor{dhsgreen!40}\textbf{90.0} & 70.0 & 84.0 & 74.5 & 94.0 & \cellcolor{dhsgreen!29}83.1 \\
\bottomrule
\end{tabular}%
}
\end{table}

        % C.13 Bootstrap CIs (Tables 20 + 21)
\subsection{Baseline protocols and adaptations}
\label{app:baselines_protocol}

\paragraph{Per-method configuration.}
All non-fine-tuned methods use gpt-5.4-nano. Released code is run under each paper's recommended best configuration; only the scientifically substantive parameters are listed.

\begin{itemize}\itemsep2pt
\item \textbf{Closed-World prompt enrichment.} CW is not a separate method but a prompt-level addendum: the standalone doctor's system prompt is augmented with the active closed disease list and an instruction to prefer names from this list at commit time. Including CW measures whether merely surfacing the candidate space is sufficient on top of an unconstrained standalone doctor.

\item \textbf{Chain-of-Thought (CoT)~\cite{wei2022cot}.} Prompt augmented with the standard zero-shot Chain-of-Thought instruction: each turn the model emits a step-by-step rationale that summarises findings, lists top candidates with supporting and refuting evidence, and identifies the next discriminating action.

\item \textbf{DDx-CoT~\cite{savage2024diagreasoning}.} Prompt augmented with the Differential Diagnosis Chain-of-Thought instruction of Savage et al.: each turn the model builds or refines a differential and lists supporting and refuting findings per candidate before choosing the next question. The instruction is taken verbatim; the source's single-turn worked examples are omitted because we evaluate in a multi-turn dialogue setting.

\item \textbf{Bayesian-CoT~\cite{savage2024diagreasoning}.} Prompt augmented with Savage et al.'s Bayesian Inference CoT instruction: each turn the model is asked to state an explicit posterior over top candidates and update it with new findings using prior plus likelihood-ratio reasoning. This mirrors \ours's Bayesian engine in instruction form, so the comparison isolates whether prompt-level Bayesian reasoning is sufficient or whether the deterministic engine is needed.

\item \textbf{MediQ-Expert~\cite{li2024mediq}.} The paper's recommended ScaleExpert variant with rationale generation, self-consistency $n{=}3$, an abstention threshold of at least $4$ on a $1$ to $5$ Likert scale, and a per-case question budget. MediQ requires hand-curated multiple-choice options for each case and a single 100-patient pool is impractical for it at this option granularity, so we evaluate every benchmark column on disjoint samples and report each separately; for fairness, every other baseline is evaluated on the same samples. We report two scoring conventions:
\begin{itemize}\itemsep0pt
  \item \emph{force-final} (paper-faithful): cases that exhaust the budget without crossing the Likert threshold are forced to commit on the highest-confidence intermediate choice; coverage is $100\%$ by construction.
  \item \emph{strict-abstain}: those budget-exhausted cases are recorded as abstentions instead. We report this as a paper-side ablation because force-final inflates coverage to $100\%$ and counts the model's unconfident commits as commits, depressing selective accuracy; strict-abstain isolates what MediQ would do if its own threshold-based abstention rule were respected at scoring time.
\end{itemize}

\item \textbf{MEDDxAgent (paper-best)~\cite{meddxagent2025}.} The recommended best iterative configuration: a single-LLM diagnosis agent with dynamic 5-shot fewshot retrieval (active case excluded), top-2 PubMed retrieval, and a fixed schedule of 3 iterations $\times$ 5 history-taking questions per iteration. MEDDxAgent has no abstention mechanism by design, so its coverage is $100\%$ by construction.

\item \textbf{MEDDxAgent (closed-world variant).} Same orchestration as paper-best but with PubMed retrieval disabled and dynamic fewshot replaced by a static random draw, isolating the contribution of MEDDxAgent's external retrieval scaffolding from the rest of its orchestration.

\item \textbf{AgentClinic~\cite{schmidgall2024agentclinic}.} AgentClinic's bundled LLM doctor: short single-turn questions, optional \emph{request test} actions (resolved as ``I don't know'' on the DDxPlus columns where no measurement agent is available), and commitment within the published question budget.

\item \textbf{CoD / DiagnosisGPT-34B~\cite{chen2024cod}.} The released DiagnosisGPT-34B checkpoint (Yi-34B fine-tune) and its broad-domain disease retriever, run inside the published CoD dialogue loop with the default per-case inquiry budget and the built-in commit confidence threshold. CoD's free-text final diagnoses are mapped to the active closed disease universe through a one-shot LLM-judge prompt that compares the free-form output (and the model's top-confidence candidates) against the closed list and either picks the closest semantic match or returns \texttt{Unknown}; substring matching is brittle for CoD's verbose differentials, so this LLM mapping is necessary to score CoD on the same closed universe as the other baselines.

\item \textbf{UoT~\cite{hu2024uot}.} LLM information-pursuit method that expands a yes/no question tree and selects the question maximising expected information gain estimated from the LLM's own answer probabilities. UoT's tree expansion stalls when the patient cannot answer cleanly: hedged or ``I don't know'' responses are correctly parsed as ``no information'' by our yes/no parser, but the resulting empty turn provides no posterior update so the tree cannot make progress. The effect is most pronounced on the HT-sim column, where the vendor patient prompt explicitly defaults to ``I don't know'' for any feature outside the profile, which substantially deflates UoT's reported numbers there.
\end{itemize}

\paragraph{AgentClinic protocol adaptations.}
The AgentClinic-MedQA column reuses AgentClinic's published patient and moderator agents unchanged. Two evaluation-level adjustments are needed for cross-baseline comparability: (i) the AC-50 closed disease universe is pinned and every doctor must commit within or abstain over this set (patient utterances are unaffected); (ii) every doctor opens with the same fixed greeting so the chief complaint is elicited from the patient simulator on the first turn rather than implicitly assumed.

\section{Extended related work}
\label{app:related_extended}

This appendix expands on Section~\ref{sec:related} with a system-by-system discussion of the four families that motivate \discuss{\ours}'s strict separation, ending with a synthesis of where each family places the LLM relative to the diagnostic decision loop. Table~\ref{tab:rw_comparison} summarises which of the three core diagnostic components (knowledge base, posterior tracking, question selection) each system delegates to the LLM versus to an external module.

\paragraph{Auditable probabilistic CDSS.}
The classical CDSS literature explored multiple probabilistic formalisms for diagnosis. MYCIN~\cite{shortliffe1976} represented diagnostic uncertainty through certainty factors and rule-based inference and was evaluated against infectious-disease specialists in a blinded protocol~\cite{yu1979}. \citet{dedombal1972} showed that even a flat na\"{\i}ve Bayes model could surpass senior surgeons on acute abdominal pain when fed structured signs and symptoms. INTERNIST-I and its successor QMR~\cite{miller1982} scaled heuristic disease-profile scoring to several hundred internal-medicine conditions, and DXplain~\cite{barnett1987} maintained a disease--finding matrix with modified Bayesian updates over thousands of diseases. Later work introduced richer graphical-model representations: Pathfinder~\cite{heckerman1992} placed lymph-node pathology under a full Bayesian belief network, and CASNET~\cite{weiss1978} expressed glaucoma management through a layered causal model linking findings to disease states to therapeutic actions. All of these systems shared the same structural pattern, an explicit probabilistic engine over a curated knowledge base coupled to a structured-input modality that demanded the clinician encode the case in the system's vocabulary; their failure to scale into routine practice, captured by the ``Greek Oracle'' diagnosis~\cite{miller1990}, was an interface failure rather than a reasoning failure.

\paragraph{End-to-end LLM-based diagnostic systems.}
Modern LLM-driven systems span single-turn QA, multi-turn dialogue, and domain-specialised model checkpoints. Med-PaLM~2~\cite{singhal2023} and Med-Gemini~\cite{yang2024medgemini} pushed standalone-LLM medical question-answering to expert level on MedQA-style benchmarks but operate single-turn with no belief state and no information-seeking behaviour. AMIE~\cite{tu2025amie} extended this paradigm to multi-turn dialogue through self-play training between simulated doctor and patient roles, achieving expert-comparable diagnostic performance on a curated history-taking benchmark. \citet{xu2024reasoning} aligned multi-turn medical dialogue with abductive clinical reasoning, while DocCHA~\cite{doccha2025} decomposes the diagnostic process into modular sub-tasks (symptom collection, history acquisition, causal-graph construction) but routes all decisions through the LLM. A parallel line of domain-specialised checkpoints fine-tunes base LLMs on clinical corpora, exposing only an end-to-end interface. Across all of these, no calibrated posterior, principled stopping rule, or deployment-time abstention mechanism is exposed: language and reasoning remain entangled inside a single generative model.

\paragraph{Probability--language hybrids.}
Hybrid systems inject probabilistic structure into LLM-driven pipelines while keeping the LLM in at least one part of the inference loop. BED-LLM~\cite{choudhury2025} formulates each turn as Bayesian Experimental Design: a Rao-Blackwellised estimator extracts a posterior over a target variable from the LLM's own predictive distributions and selects the question that maximises expected information gain, with the probabilistic model itself sourced from the language model. DeLLMa~\cite{liu2024dellma} pairs an LLM with classical decision theory, eliciting a utility function from the model and selecting actions that maximise expected utility. BIRD~\cite{bird2024} aligns a Bayesian network with LLM abductions and then performs a deductive inference step, with the goal of producing controllable probability estimates more reliable than direct LLM scoring. APP~\cite{zhu2025app} combines clinical guidelines with entropy-based active learning and Bayesian-style updates for human-centric medical dialogue. \citet{sun2024planner} couple an LLM parser with a separately RL-trained planner. UoT~\cite{hu2024uot} performs information-pursuit through an LLM-generated binary question tree with Q-value-based selection, originally for the 20-Questions setting. MediQ~\cite{li2024mediq} contributes the abstention mechanism most directly relevant to clinical dialogue: it shows that pairing question-asking with explicit scale-based abstention recovers the accuracy that naive prompting otherwise loses. CoD / DiagnosisGPT~\cite{chen2024cod} fine-tunes a clinical LLM with an explicit confidence head and a $9{,}604$-disease retriever, while~\citet{kim2024finetuning} examine the privacy and safety trade-offs of clinical fine-tuning more broadly. Across the family, the LLM remains responsible for at least one of: producing the probabilistic distribution, generating candidate questions, or scoring outputs, which keeps language and probability entangled.

\paragraph{Agentic diagnostic frameworks.}
A complementary family wraps LLMs in multi-agent orchestration. MEDDxAgent~\cite{meddxagent2025} runs a four-agent loop (DDx-driver, history-taking, knowledge-retrieval, diagnosis) with PubMed or closed-world retrieval and optional dynamic few-shot prompting, where every agent calls the same backbone LLM. AgentClinic~\cite{schmidgall2024agentclinic} simulates a clinical environment through doctor, patient, measurement, and moderator agents, each implemented as an LLM with role-specific prompts. \citet{li2024mediq} package their abstention-aware pipeline as a five-step Expert framework (Assessment, Abstention, Question, Integration, Decision). These systems modularise the workflow but rely on heuristic role-playing rather than information-theoretic objectives for question selection, and the LLM is still the primary inference engine inside each role; we adapt several of them as baselines (Section~\ref{sec:baselines}, Appendix~\ref{app:baselines_protocol}).

\paragraph{Safety, calibration, and bias risks.}
The structural absence of a calibrated probabilistic state in end-to-end LLMs has been linked to several measurable safety risks. \citet{savage2025calibration} report that verbalised confidence from clinical LLMs shows minimal variation between correct and incorrect diagnoses, undermining its use as an abstention signal. \citet{omar2025hallucination} demonstrate that frontier clinical LLMs are highly vulnerable to adversarial hallucination during clinical decision support, with multi-model assurance providing only partial mitigation. The medical knowledge encoded in these models is also skewed: \citet{omiye2023race} document propagation of race-based clinical patterns, and \citet{olatunji2024afrimedqa} report substantial accuracy gaps on African medical questions relative to Western populations. These risks motivate architectures that keep the probabilistic state outside the language model, expose explicit confidence rather than verbalised self-reports, and avoid embedding patient-specific information in model weights at all.

\paragraph{Why \discuss{\ours} differs.}
Across the four families above, no system keeps the LLM strictly out of the diagnostic decision loop. Classical CDSS provided the auditable statistical engine but coupled it to a structured-input interface that excluded clinicians from natural conversation. End-to-end LLMs solved the interface but folded conversation, question selection, diagnosis, and stopping into a single generative model. Hybrid systems and agentic frameworks reintroduce probabilistic components and modular workflows but still rely on the LLM as a probability source, candidate generator, planner substrate, or reasoning state. \discuss{\ours}'s contribution is not a new probabilistic engine and not a new LLM, but a deployment pattern that reuses classical statistical reasoning and modern conversational interfaces without letting either contaminate the other. Table~\ref{tab:rw_comparison} makes this explicit: for each surveyed system, it indicates whether the LLM is responsible for the knowledge base, the posterior, the question-selection criterion, and the abstention rule.

%% ============================================================
%% TABLE: Related Work Comparison
%% File: tables/tab_rw_comparison.tex
%% ============================================================

\begin{table}[t]
\caption{System-by-system comparison of diagnostic dialogue approaches against the architectural axes our position depends on (Section~\ref{sec:related}). \textbf{MT}: supports multi-turn dialogue. \textbf{Belief}: explicit probabilistic posterior over diseases ($\checkmark$), an LLM-derived approximate posterior (Approx), or none ($\times$). \textbf{KB}: clinical knowledge source (Cur.\ = curated; Trained = baked into model weights; RAG = retrieval over external corpus; LLM-el.\ = elicited from a frontier LLM; ``--'' = none). \textbf{Q-sel.}: question-selection criterion (LLM = LLM-internal heuristic; EIG = expected information gain; Tree = lookahead tree search; RL = learned policy; Conf.\ = conformal-set shrinkage; Util.\ = expected-utility maximisation; ``--'' = no question-asking). \textbf{Abst.}: abstention mechanism (Heur.\ = ad hoc; Calib.\ = calibrated; Ctrl.\ = continuously controllable; ``--'' = none). \textbf{LLM-free}: the LLM is not in the diagnostic decision loop. Systems marked with $\ast$ have no public code release or are closed products and are therefore cited but not reproduced as baselines in our experiments.}
\label{tab:rw_comparison}
\centering
\footnotesize
\setlength{\tabcolsep}{4pt}
\renewcommand{\arraystretch}{1.05}
\begin{tabular}{l c c c c c c}
\toprule
\textbf{System} & \textbf{MT} & \textbf{Belief} & \textbf{KB} & \textbf{Q-sel.} & \textbf{Abst.} & \textbf{LLM-free} \\
\midrule
\multicolumn{7}{l}{\textit{Classical CDSS}} \\
\midrule
MYCIN$^{\ast}$~\cite{shortliffe1976}                & \xmark & \cmark & Cur. & --       & --     & \cmark \\
INTERNIST/QMR$^{\ast}$~\cite{miller1982}            & \xmark & \cmark & Cur. & --       & --     & \cmark \\
DXplain$^{\ast}$~\cite{barnett1987}                 & \xmark & \cmark & Cur. & --       & --     & \cmark \\
Pathfinder$^{\ast}$~\cite{heckerman1992}            & \xmark & \cmark & Cur. & --       & --     & \cmark \\
CASNET$^{\ast}$~\cite{weiss1978}                    & \xmark & \cmark & Cur. & --       & --     & \cmark \\
\midrule
\multicolumn{7}{l}{\textit{End-to-end frontier LLMs}} \\
\midrule
Med-PaLM~2$^{\ast}$~\cite{singhal2023}              & \xmark & \xmark & Trained      & --   & --    & \xmark \\
Med-Gemini$^{\ast}$~\cite{yang2024medgemini}        & \xmark & \xmark & Trained      & --   & --    & \xmark \\
AMIE$^{\ast}$~\cite{tu2025amie}                     & \cmark & \xmark & Trained & LLM  & --    & \xmark \\
DocCHA$^{\ast}$~\cite{doccha2025}                   & \cmark & \xmark & Cur.      & LLM  & Heur. & \xmark \\
\midrule
\multicolumn{7}{l}{\textit{Domain-tuned medical LLMs}} \\
\midrule
Meditron~\cite{chen2023meditron}                    & \cmark & \xmark & Trained & --   & --    & \xmark \\
MedGemma~\cite{sellergren2025medgemma}              & \cmark & \xmark & Trained & --   & --    & \xmark \\
HuatuoGPT-o1~\cite{chen2024huatuogpt}               & \cmark & \xmark & Trained & --   & --    & \xmark \\
CoD / DiagnosisGPT~\cite{chen2024cod}               & \cmark & Approx & Trained & Entropy & Heur. & \xmark \\
DoctorAgent-RL~\cite{feng2025doctoragentrl}         & \cmark & \xmark & Trained & RL   & --    & \xmark \\
\midrule
\multicolumn{7}{l}{\textit{Probability--language hybrids}} \\
\midrule
BED-LLM$^{\ast}$~\cite{choudhury2025}               & \cmark & Approx & --      & EIG       & --     & \xmark \\
APP$^{\ast}$~\cite{zhu2025app}                      & \cmark & Approx & RAG     & Entropy   & --     & \xmark \\
Planner-LLM~\cite{sun2024planner}                   & \cmark & \xmark & --      & RL        & --     & \xmark \\
UoT~\cite{hu2024uot}                                & \cmark & \xmark & --      & Tree+EIG  & --     & \xmark  \\
CIP$^{\ast}$~\cite{chan2025cip}                                        & \cmark & \xmark & --      & Conf.     & Calib. & \xmark \\
MedKGI$^{\ast}$~\cite{wang2025medkgi}                                     & \cmark & Approx & Cur.     & EIG       & Heur.  & \xmark \\
MedClarify$^{\ast}$~\cite{wong2026medclarify}                                & \cmark & Approx & --      & EIG       & --     & \xmark \\
\midrule
\multicolumn{7}{l}{\textit{Agentic frameworks}} \\
\midrule
MEDDxAgent~\cite{meddxagent2025}                    & \cmark & \xmark & RAG  & LLM & --    & \xmark \\
AgentClinic~\cite{schmidgall2024agentclinic}        & \cmark & \xmark & --   & LLM & --    & \xmark \\
MediQ~\cite{li2024mediq}                            & \cmark & \xmark & --   & LLM & Heur. & \xmark \\
\midrule
\textbf{\discuss{\ours} (Ours)}                      & \cmark & \cmark & Cur./LLM-el. & EIG & Ctrl. & \cmark \\
\bottomrule
\end{tabular}
\end{table}

        % D Extended related work

%% ================================================================
%% E. OPEN ITEMS (remove before submission)
%% ================================================================

%% \input{sections/appendix/todos}

%% --- NeurIPS Checklist (stripped for arXiv to avoid double-blind footprint) ---
% \newpage
% \input{checklist}

\end{document}